\documentclass[a4paper,11pt]{article}
\usepackage{latexsym,amssymb,enumerate,amsmath,epsfig,amsthm,dsfont,bm}
\usepackage[margin=1in]{geometry}
\usepackage{setspace,color}
\usepackage{tikz}
\usepackage{floatrow}
\usepackage{graphicx}
\usepackage{algorithmic}
\usepackage[colorlinks=true, allcolors=blue]{hyperref}
\usepackage{epstopdf}
\usepackage[multidot]{grffile}
\usepackage{comment}
\usepackage{multirow}
\usepackage{smile}

\DeclareMathOperator*{\argmin}{arg\,min}

\allowdisplaybreaks

\begin{document}

\title{A PCA Based Model for Surface Reconstruction from Incomplete Point Clouds}
\author{
Hao Liu\thanks{Department of Mathematics, Hong Kong Baptist University, Kowloon Tong, Hong Kong
	(Email: {haoliu@hkbu.edu.hk}). Hao Liu was partially supported by the National Natural Science Foundation of China 12201530, HKRGC ECS 22302123, HKRGC GRF 12301925 and Guangdong and Hong Kong Universities ``1+1+1'' Joint Research Collaboration Scheme UICR0800008-24.}
}
\date{}
\maketitle

\begin{abstract}
	Point cloud data represents a crucial category of information for mathematical modeling, and surface reconstruction from such data is an important task across various disciplines. However, during the scanning process, the collected point cloud data may fail to cover the entire surface due to factors such as high light-absorption rate and occlusions, resulting in incomplete datasets. Inferring surface structures in data-missing regions and successfully reconstructing the surface poses a challenge. In this paper, we present a Principal Component Analysis (PCA) based model for surface reconstruction from incomplete point cloud data. Initially, we employ PCA to estimate the normal information of the underlying surface from the available point cloud data. This estimated normal information serves as a regularizer in our model, guiding the reconstruction of the surface, particularly in areas with missing data. Additionally, we introduce an operator-splitting method to effectively solve the proposed model. Through systematic experimentation, we demonstrate that our model successfully infers surface structures in data-missing regions and well reconstructs the underlying surfaces, outperforming existing methodologies.
\end{abstract}

\section{Introduction}
Surface reconstruction serves as a fundamental tool in data visualization, three-dimensional modeling, and advanced data processing. Its applications span various fields, including urban reconstruction \cite{wang2018lidar}, medical imaging \cite{tasdizen2003geometric}, computer graphics \cite{lai2013ridge}, the preservation of cultural heritage \cite{berger2017survey}, and the development of the metaverse \cite{zhao2022metaverse}, all of which have garnered significant attention in recent years. For instance, in urban reconstruction, surface reconstruction algorithms utilize point cloud data obtained from light detection and ranging (LiDAR) systems to create three-dimensional urban models. These models play a crucial role in urban planning, virtual tourism, and real-time emergency response, among other applications. Furthermore, the metaverse, a three-dimensional virtual environment designed to offer immersive experiences, relies on surface reconstruction as an essential component, as it directly influences the quality of three-dimensional representations of real-world objects within the virtual world.

Point cloud data, a widely utilized data type for surface reconstruction, can be readily obtained through LiDAR technology. However, the underlying surfaces often exhibit complex structures, which complicates the reconstruction process. Over the past few decades, significant efforts have been dedicated to developing robust and efficient mathematical models and algorithms to address this challenge \cite{zhao2000implicit,liang2013robust, amenta1998new, edelsbrunner1998shape, dong2011wavelet, marcon2008fast}; for a comprehensive review of this research direction, see \cite{berger2017survey}. Among the various approaches, a substantial subset of methods employs implicit representations of surfaces, such as level set functions, which have demonstrated effectiveness in reconstructing surfaces with intricate topological features.

In the level set method, a surface $\Gamma$ is implicitly represented as the zero level set of a level set function $\phi$. Zhao et al. \cite{zhao2000implicit,zhao2001fast} proposed a reconstruction approach that minimizes the distance between every point on the surface and the point cloud data. Specifically, let $f(\xb)$ denote the distance function from a point $\xb$ to the point cloud. The surface reconstruction is formulated as the following optimization problem:
\begin{align}
	\min_{\Gamma} \left[\int_{\Gamma} f^s(\xb) d \sigma\right]^{1/s} =\min_{\phi} \left[\int_{\Omega} f^s(\xb)|\nabla\phi|\delta(\phi)d\xb\right]^{1/s},
	\label{eq.zhao.dis}
\end{align}
where $d \sigma$ represents the surface area element, $\Omega$ denotes the computational domain, and $\delta$ is the Dirac delta function. In equation (\ref{eq.zhao.dis}), the energy term represents the surface area of $\Gamma$ weighted by the distance function $f(\xb)$. The first formulation involves an integral over the reconstructed surface $\Gamma$, while the second formulation transforms it into an integral over the computational domain $\Omega$. Thus, the energy expressed in (\ref{eq.zhao.dis}) characterizes the surface area of $\Gamma$ in relation to the distance function $f(\xb)$.

Model (\ref{eq.zhao.dis}) has inspired several subsequent works. Building upon this foundation, \cite{he2020curvature} introduced surface curvature as a regularizer, resulting in a model that effectively enhances the recovery of surface edges and corners. In \cite{law2025approximated}, the authors proposed to use the $L_1$ curvature to regularize the reconstructed surface, which well keeps sharp features. Additionally, image segmentation-based models have been proposed in \cite{ye2010fast,liang2013robust}, where the distance function $f(\mathbf{x})$ is utilized within edge indicator functions to accurately locate the surface. In \cite{liu2017level}, model (\ref{eq.zhao.dis}) is integrated with principal component analysis to reconstruct curves on Riemannian manifolds. 
To address the challenges posed by noisy data, \cite{shi2004shape} developed a model that employs the negative log-likelihood of the data as a fidelity term while utilizing surface area as a regularization strategy. When information regarding surface normal directions is available, \cite{liu2008implicit} introduced a regularization term that aligns the normals of the reconstructed surface with the provided normal information. The Poisson surface reconstruction method, proposed in \cite{kazhdan2006poisson}, reconstructs the surface from oriented points by solving a Poisson problem. Furthermore, \cite{cui2023surface} presented a directional G-norm-based model for reconstructing surfaces from noisy point clouds. 
Level set-based methods for surface fairing are explored in \cite{lai2013ridge,brito2013fast}. 
In addition to level set-based approaches, numerous other methods have been proposed for surface processing tasks. Graph-cut-based methods are examined in \cite{hornung2006robust,paris2006surface,wan2012reconstructing,wansurface,shi2012curvature} for surface reconstruction from point clouds. Surface reconstruction from strokes is investigated in \cite{wu2007shapepalettes,hahn2013stroke}, while surface smoothing is addressed in \cite{liu2019triangulated,liu2022operator,elsey2009analogue}.

Surface reconstruction models typically involve solving a nonconvex optimization problem. One approach to address this challenge is through gradient descent, which introduces an artificial time parameter, as demonstrated in works such as \cite{zhao2001fast,shi2004shape,liu2017level}. To ensure the stability of the algorithm, gradient descent generally necessitates a small time step. A two-stage method was proposed in \cite{brito2013fast}, where the first stage processes high-order quantities, such as normal information, while the second stage utilizes this information to reconstruct the surface. However, this method involves an inner iteration, which can increase overall CPU time.

To enhance the efficiency of the algorithm, the alternating direction method of multipliers (ADMM) and operator splitting methods have been introduced in \cite{he2020curvature,cui2023surface,estellers2012efficient}. While ADMM can effectively solve the optimization problem, its performance is sensitive to the choice of hyperparameters, necessitating additional effort to identify optimal parameters. Operator-splitting methods, as employed in \cite{he2018fast,he2020curvature,liu2022operator}, decompose a complex optimization problem into several simpler subproblems by introducing auxiliary variables. Following this decomposition, each subproblem either has a closed-form solution or can be solved efficiently. Notably, operator-splitting methods are less sensitive to hyperparameters. Research has indicated that operator-splitting methods are more efficient than ADMM for image processing tasks \cite{deng2019new,duan2022fast,he2024euler}. An accelerated operator-splitting method was proposed in \cite{wu2024variational}. In \cite{wang2021efficient}, a threshold dynamics method was proposed to solve (\ref{eq.zhao.dis}), which greatly improved the efficiency.

Existing methods for surface reconstruction primarily focus on point cloud data that contain samples (potentially perturbed by noise) distributed across the entire surface. In practice, however, the point cloud data used for surface reconstruction is often incomplete due to factors such as high light-absorption rate and occlusions during the scanning process \cite{tagliasacchi2009curve,mitra2007dynamic}. When point data is missing in certain regions, the challenge of inferring surface information in these areas and reconstructing the complete surface remains underexplored.
In this context, the authors of \cite{schnabel2009completion} addressed the issue of incomplete point cloud data by incorporating normal information. They proposed a reconstruction method that utilizes a predefined set of shape primitives. Additionally, the extraction of curve skeletons from incomplete point clouds has been investigated in \cite{tagliasacchi2009curve}.

In this paper, we propose a novel model for reconstructing surfaces from incomplete point cloud data, which utilizes estimated surface normal information from the point cloud. Our model implicitly represents the surface using level set methods. To infer surface structures in regions with missing data, we employ principal component analysis (PCA) to estimate surface normals. We assume that the surface normals exhibit minimal variation between data-available and data-missing regions. For locations in the data-missing region that are adjacent to data-available regions, the normals estimated by PCA serve as effective approximations of the true surface normals. 
Our proposed model incorporates surface curvature and the estimated normals as regularizers. Given that PCA utilizes neighboring information, it demonstrates robustness against noise. Additionally, employing the estimated normals as regularizers enhances the model's performance in the presence of noisy data. We introduce an operator-splitting method to effectively solve the proposed model. Our contributions are summarized as follows:
\begin{enumerate}
	\item We present a novel PCA based model for surface reconstruction from incomplete point clouds. This model utilizes normals estimated by PCA as regularizers to infer surface structures in regions with missing data.
	\item We propose an operator-splitting method to solve the model. Each subproblem in our algorithm either has a closed-form solution or can be addressed by efficient solvers.
	\item We demonstrate the efficacy of the proposed model through systematic numerical experiments. In comparison to existing methods for incomplete data, our model effectively infers surface structures in data-missing regions and achieves superior surface reconstruction.
\end{enumerate}

The remainder of this paper is organized as follows: In Section \ref{sec.formulation}, we introduce our model and reformulate it as an initial-value problem. We present our operator-splitting method in Section \ref{sec.splitting}, along with its numerical discretization in Section \ref{sec.discretization}. Implementation details are discussed in Section \ref{sec.implementation}. The effectiveness of the proposed model is demonstrated through systematic numerical experiments in Section \ref{sec.experiments}. Finally, we conclude the paper in Section \ref{sec.conclusion}.

\section{Model Formulation}
\label{sec.formulation}
In this section, we propose our PCA based model as an optimization problem and subsequently convert it into an equivalent initial value problem using the level set formulation.

\subsection{Model derivation}
Given a set of points $\cS$ sampled from a $d-1$ dimensional surface $\Gamma^* \in \RR^d$, our objective is to reconstruct the surface from the point cloud $\cS$. We define $f(\xb) = \min_{\zb \in \cS} \|\zb - \xb\|_2$ as the distance from $\xb$ to $\cS$. The function $f(\xb)$ approaches 0 when $\xb$ is close to the point cloud $\cS$. This distance function is utilized by \cite{zhao2000implicit,zhao2001fast} to locate and reconstruct the surface $\Gamma^*$, as described in (\ref{eq.zhao.dis}).

Sometimes, the collected point cloud is incomplete due to obstacles. In regions where point cloud data is missing, controlling the reconstructed surface using model (\ref{eq.zhao.dis}) becomes challenging. If we have access to the normal direction information of $\Gamma^*$ in the data-missing regions, this difficulty can be mitigated by utilizing the normal information: in areas where data is absent, we can constrain the reconstructed surface to be orthogonal to the normal direction. For any $\mathbf{x} \in \Gamma^*$, let $\mathbf{n}^*(\mathbf{x})$ denote the normal direction of $\Gamma^*$ at $\mathbf{x}$. We can use the term $1 - |\mathbf{n}^*(\mathbf{x}) \cdot \mathbf{n}(\mathbf{x})|^2$, where $\mathbf{n}(\mathbf{x})$ represents the normal direction of the reconstructed surface, as a penalty term to regularize the reconstruction. This term encourages $\mathbf{n}(\mathbf{x})$ to align with $\mathbf{n}^*(\mathbf{x})$.

However, this strategy presents two challenges: (i) The normal direction information of $\Gamma^*$ is typically unknown in many situations. (ii) Since the reconstructed surface $\Gamma$ does not exactly coincide with $\Gamma^*$, it is necessary to extend the definition of the normal direction to the entire computational domain, or at least to a subregion encompassing $\Gamma$.

In this paper, we propose utilizing principal component analysis to estimate the normal direction and subsequently reconstruct the surface. We consider the following model:
\begin{align}
	\widehat{\Gamma} = \argmin_{\Gamma} \left[\eta_0 \int_{\Gamma} f^2(\mathbf{x}) d\sigma + \frac{\eta_1}{2} \int_{\Gamma} |\kappa(\mathbf{x})|^{2} d\sigma + \frac{\eta_2}{2} \int_{\Gamma} r(\mathbf{x}) (1 - |\mathbf{p}_d(\mathbf{x}) \cdot \mathbf{n}(\mathbf{x})|^2) d\sigma \right],
	\label{eq.energy.1}
\end{align}
where $\kappa(\mathbf{x})$ denotes the curvature of $\Gamma$ at $\mathbf{x}$, $\mathbf{p}_d(\mathbf{x})$ is the $d$-th principal direction computed from $\mathcal{S}$ at $\mathbf{x}$, $\eta_0, \eta_1, \eta_2 \geq 0$ are weight parameters, and $r(\mathbf{x}) \geq 0$ is a weight function. 

In (\ref{eq.energy.1}), the first term serves as a data fidelity term that encourages the surface to remain close to the point cloud. The second term penalizes the curvature of $\Gamma$, promoting smoothness and assisting in the recovery of surface corners, as discussed in \cite{he2020curvature}.

The third term in (\ref{eq.energy.1}) penalizes the angles between $\mathbf{n}(\mathbf{x})$ and $\mathbf{p}_d(\mathbf{x})$. When $\mathbf{x} \in \Gamma^*$ is close to $\mathcal{S}$, $\mathbf{p}_d(\mathbf{x})$ serves as a reliable estimate of the exact surface normal $\mathbf{n}^*(\mathbf{x})$. Even when $\mathbf{x} \notin \Gamma^*$, one can still compute $\mathbf{p}_d(\mathbf{x})$ using PCA, which effectively extends $\mathbf{n}^*(\mathbf{x})$ from $\Gamma^*$ to a tubular neighborhood enclosing $\mathcal{S}$. 

\begin{figure}[t!]
	\centering
	\begin{tabular}{cccc}
		(a) & (b) & (c) & (d) \\
		\includegraphics[width=0.2\textwidth]{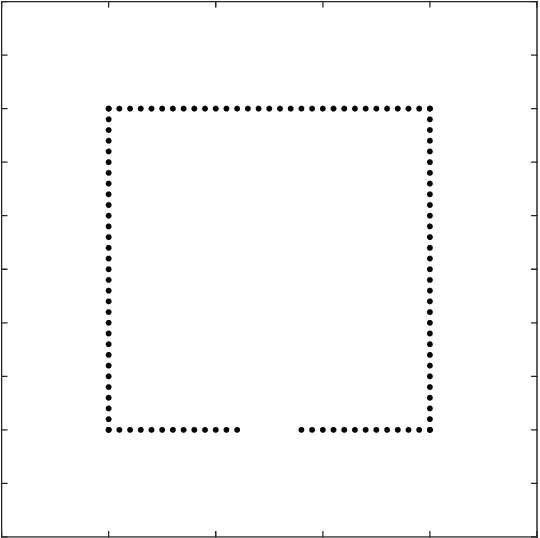} &
		\includegraphics[width=0.2\textwidth]{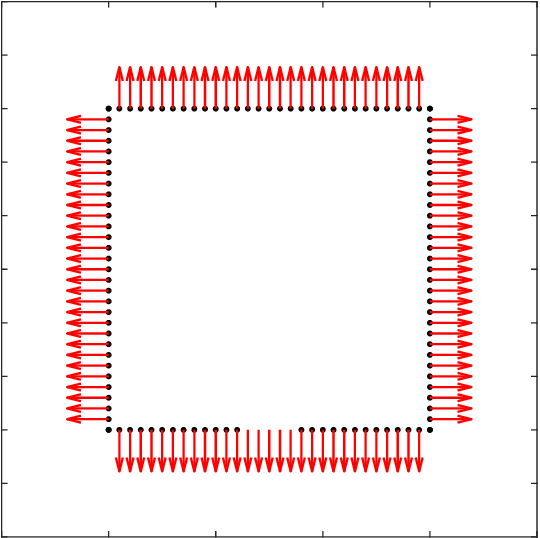} &
		\includegraphics[width=0.2\textwidth]{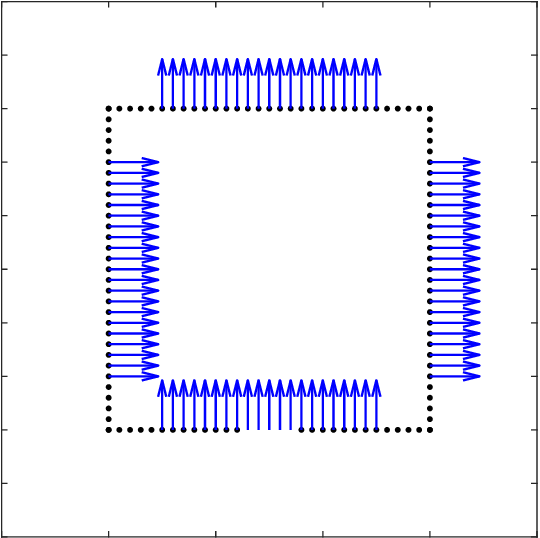} &
		\includegraphics[width=0.2\textwidth]{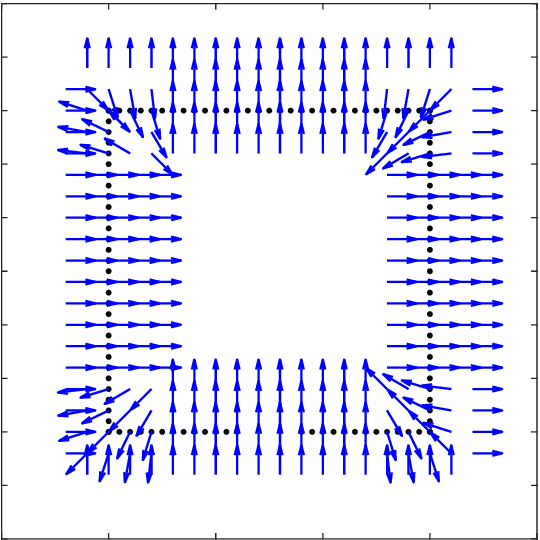} 
	\end{tabular}
	\caption{An example by PCA. (a) The incomplete point cloud sampled from a square. Data in the central region of the bottom edge are missing. (b) Exact normal directions on the square. (c) Estimated normal directions by PCA in the central region of each edge. (d) Estimated normal direction field by PCA.}
	\label{fig.pca.example}
\end{figure}

For instance, consider a point cloud sampled from a square where the data in the center of the bottom edge is missing, as illustrated in Figure \ref{fig.pca.example}(a). The exact surface normals are depicted in (b). By applying PCA, the estimated normals in the central region of each edge are shown in (c). We observe that the estimated normals align with the exact ones up to the sign. In (d), we present the estimated normals around the point cloud. PCA naturally extends the surface normals from the point cloud to form a narrow tube that encompasses it.

The weight function $r(\mathbf{x})$ in the third term is location-dependent, controlling the influence of the normal penalty across different regions. For example, in regions where data is available, one typically prefers the surface to be primarily driven by the distance and curvature terms. Conversely, in data-missing regions, it is desirable for the normal information term to act as the dominant regularizer. Details regarding the choice of $r(\mathbf{x})$ will be discussed in Section \ref{sec.r}. By minimizing this term, $\Gamma$ is compelled to align normally with $\mathbf{p}_d(\mathbf{x})$. A similar term is utilized in \cite{liu2017level} for dimension reduction on Riemannian surfaces.

Introducing the third term in (\ref{eq.energy.1}) offers three key benefits: (i) The vector $\mathbf{p}_d(\mathbf{x})$ serves as an approximate extension of $\mathbf{n}^*(\mathbf{x})$ to a larger region, aiding in the control of the reconstructed surfaces in data-missing regions. (ii) Since the principal direction $\mathbf{p}_d(\mathbf{x})$ is estimated using the neighboring information around $\mathbf{x}$, it varies smoothly, up to the sign. The inclusion of the third term in (\ref{eq.energy.1}) mitigates abrupt changes in $\widehat{\Gamma}$ from data-available regions to data-missing regions, thereby enhancing the smoothness of the reconstructed surface. (iii) Furthermore, as the computation of $\mathbf{p}_d(\mathbf{x})$ relies on neighboring information, it exhibits resilience to noise. This characteristic contributes to the robustness of $\widehat{\Gamma}$ against noise.

\subsection{A level set formulation}

Problem (\ref{eq.energy.1}) involves integrals defined over the surface $\Gamma$, which presents challenges for direct solution. To address this difficulty, we employ level set methods to convert the integration domain to $\Omega$.

Let $\phi$ be a level set function so that
\begin{align}
	\phi(\xb)\begin{cases}
		>0 & \mbox{ when $\xb$ is outside $\Gamma$,}\\
		=0 & \mbox{ when } \xb\in \Gamma,\\
		<0 & \mbox{ when $\xb$ is inside $\Gamma$.}
	\end{cases}
\end{align} 
The surface $\Gamma$ is represented as the zero level set of $\phi$: $\Gamma=\{\xb:\phi(\xb)=0\}$.
Let 
$$
H(a)=\begin{cases}
	1 & \mbox{ for } a>0,\\
	0 & \mbox{ otherwise}
\end{cases}
$$
be the heavyside function, and $\delta(a)=H'(a)$ be the Dirac delta function.  
Problem (\ref{eq.energy.1}) is equivalent to
\begin{align}
	\widehat{\psi}=\argmin_{\phi} \bigg[ \eta_0\int_{\Omega} f^2(\xb)\delta(\phi) |\nabla \phi| d\xb 	&+ \frac{\eta_1}{2} \int_{\Omega}  \left|\nabla \cdot \frac{\nabla \phi}{|\nabla \phi|}\right|^{2} \delta(\phi) |\nabla \phi|d\xb \nonumber\\
	&+ \frac{\eta_2}{2} \int_{\Omega} r(\xb) \left(1-\left|\pb_d(\xb)\cdot \frac{\nabla\phi}{|\nabla\phi|}(\xb)\right|^{2}\right) \delta(\phi) |\nabla \phi|d\xb  \bigg]
	\label{eq.energy.2}
\end{align}
with $\widehat{\Gamma}$ being the zero level set of $\widehat{\psi}$. Rather than defining integrals over the surface $\Gamma$ in (\ref{eq.energy.1}), the energy in (\ref{eq.energy.2}) is expressed as integrals over the computational domain $\Omega$.

\subsection{Reformulation of (\ref{eq.energy.2})}
In (\ref{eq.energy.2}), $f$ and $\pb_d$ can be pre-computed using $\cS$ and are independent of $\phi$. Thus, our focus is solely on $\phi$. To decompose the nonlinearity and simplify the second and third terms, we introduce auxiliary variables $\vb=\frac{\nabla\phi}{|\nabla\phi|}$, and $p=\nabla \cdot \frac{\nabla\phi}{|\nabla\phi|}$. The presence of the Dirac delta function $\delta(\phi)$ complicates the numerical solution of problem (\ref{eq.energy.2}). To address this issue, we consider a smoothed version as proposed in \cite{chan2001active,bae2016augmented}.
\begin{align}
	\delta_{\varepsilon}(\phi)=\frac{\varepsilon}{\pi(\varepsilon^2+\phi^2)}.
	\label{eq.delta.approx}
\end{align}

We approximate problem (\ref{eq.energy.2}) as
\begin{align}
	\begin{cases}
		\min\limits_{\phi} \bigg[\displaystyle\eta_0\int_{\Omega} f^2(\xb)\delta(\phi) |\nabla \phi| d\xb +  \eta_1 \displaystyle\int_{\Omega}  \left|p\right|^{2} \delta_{\varepsilon}(\phi) |\nabla \phi|d\xb  +\eta_2\displaystyle\int_{\Omega} r(\xb) g(\vb) \delta_{\varepsilon}(\phi) |\nabla \phi|d\xb \\
		\hspace{2cm}+ \displaystyle\frac{\alpha_1}{2} \displaystyle\int_{\Omega}\left| \vb- \frac{\nabla \phi}{|\nabla \phi|}\right|^2d\xb 
		+ \frac{\alpha_2}{2} \displaystyle\int_{\Omega}\left| p- \nabla\cdot\frac{\nabla \phi}{|\nabla \phi|}\right|^2d\xb \bigg],\\
		p=\nabla\cdot \vb,\\
		|\vb|=1,
	\end{cases}
	\label{eq.energy.3}
\end{align}
where $g(\vb)=1-\left|\pb_d\cdot \vb\right|^{2}$, and $\alpha_1,\alpha_2>0$ are two weight parameters.
The constraints $p=\nabla\cdot \vb$ and $|\mathbf{v}| = 1$ are explicitly introduced to ensure the stability of the derived algorithm.
Define the set and its indicator functions as
\begin{align*}
	&\Sigma_1=\left\{(\vb,p):p=\nabla\cdot \vb\right\}, \quad \Sigma_2=\{\vb: |\vb|=1\},\\ &I_{\Sigma_1}(\vb,p)=\begin{cases}
		0 &\mbox{ if } (\vb,p)\in \Sigma_1,\\
		+\infty &\mbox{ otherwise},
	\end{cases}, \quad 
	I_{\Sigma_2}(\vb)=\begin{cases}
		0 &\mbox{ if } (\vb)\in \Sigma_2,\\
		+\infty &\mbox{ otherwise}.
	\end{cases}
\end{align*}
We define the functional
\begin{align*}
	&J_1(\phi)=\eta_0\int_{\Omega} f^2(\xb)\delta_{\varepsilon}(\phi) |\nabla \phi| d\xb,\\
	&J_2(\phi,p)=\eta_1 \displaystyle\int_{\Omega}  \left|p\right|^{2} \delta_{\varepsilon}(\phi) |\nabla \phi|d\xb,\\
	&J_3(\phi,\vb)= \eta_2\displaystyle\int_{\Omega}  r(\xb)g(\vb) \delta_{\varepsilon}(\phi) |\nabla \phi|d\xb,\\
	&J_4(\phi,\vb)=\frac{\alpha_1}{2} \int_{\Omega}\left| \vb- \frac{\nabla \phi}{|\nabla \phi|}\right|^2d\xb,\\
	&J_5(\phi,p)=\frac{\alpha_2}{2} \int_{\Omega}\left| p- \nabla\cdot\frac{\nabla \phi}{|\nabla \phi|}\right|^2d\xb.
\end{align*}
Solving (\ref{eq.energy.3}) is equivalent to solving the following unconstrained problem:
\begin{align}
	\min\limits_{\phi,\vb,p} \left[J_1(\phi)+ J_2(\phi,p) + J_3(\phi,\vb)+ J_4(\phi,\vb)+ J_5(\phi,p)+I_{\Sigma_1}(\vb,p)+ I_{\Sigma_2}(\vb)\right].
	\label{eq.energy.4}
\end{align}

Let $(\psi^*,\ub^*,q^*)$ be a minimizer of (\ref{eq.energy.4}), it satisfies the optimality condition
\begin{align}
	\begin{cases}
		\partial_{\phi} J_1(\psi^*) + \partial_{\phi} J_2(\psi^*,q^*) + \partial_{\phi} J_3(\psi^*,\ub^*) + \partial_{\phi} J_4(\psi^*,\ub^*) + \partial_{\phi} J_5(\psi^*,q^*)\ni 0,\\
		\partial_{\vb} J_3(\psi^*,\ub^*)+\partial_{\vb} J_4(\psi^*,\ub^*)+ \partial_{\vb}I_{\Sigma_1}(\ub^*,q^*)+ \partial_{\vb} I_{\Sigma_2}(\ub^*) \ni 0,\\
		\partial_p J_2(\psi^*,q^*) + \partial_p J_5(\psi^*,q^*) + \partial_{p}I_{\Sigma_1}(\ub^*,q^*) \ni 0,
	\end{cases}
\end{align}
where $\partial$ denotes the generalized gradient \cite[Section 1.2]{clarke1990optimization}.

We associate (\ref{eq.energy.4}) with the following initial value problem in the flavor of gradient flow:
\begin{align}
	\begin{cases}
		\frac{\partial \psi}{\partial t}+ \partial_{\phi} J_1(\psi) + \partial_{\phi} J_2(\psi,q) + \partial_{\phi} J_3(\psi,\ub) + \partial_{\phi} J_4(\psi,\ub) + \partial_{\phi} J_5(\psi,q) \ni 0,\\
		\gamma_1\frac{\partial \ub}{\partial t}+\partial_{\vb} J_3(\psi,\ub)+ \partial_{\vb} J_4(\psi,\ub)+ \partial_{\vb}I_{\Sigma_1}(\ub,q)+ \partial_{\vb} I_{\Sigma_2}(\ub) \ni \mathbf{0},\\
		\gamma_2\frac{\partial q}{\partial t}+\partial_p J_2(\psi,q) +\partial_p J_5(\psi,q) +\partial_{p}I_{\Sigma_1}(\ub,q) \ni 0,\\
		(\psi(0),\ub(0),q(0))=(\psi_0,\ub_0,q_0),
	\end{cases}
	\label{eq.ivp}
\end{align}
where $\gamma_1,\gamma_2$ are constants controling the evolution speed of $\ub$ and $q$, and $(\psi_0,\ub_0,q_0)$ is the initial condition.
The task is converted to finding the steady state solution of (\ref{eq.ivp}).
\section{Operator-splitting method}
\label{sec.splitting}
Problem (\ref{eq.ivp}) is well suited to be solved by operator-splitting method. Here we adopt the simple Lie scheme \cite{lie1893theorie}, which splits one time stepping of (\ref{eq.ivp}) into several sequential substeps. A detailed discussion on operator-splitting methods can be found in \cite{glowinski2016some,glowinski2017splitting}. Our main splitting strategy is to distribute operators in (\ref{eq.ivp}) into substeps so that each substep either can be solved explicitly or can be approximately solved efficiently. Given $(\psi^n,\ub^n,q^n)$, we update 
$(\psi^n,\ub^n,q^n)\rightarrow (\psi^{n+1/4},\ub^{n+1/4},q^{n+1/4}) \rightarrow (\psi^{n+2/4},\ub^{n+2/4},q^{n+2/4}) \rightarrow (\psi^{n+3/4},\ub^{n+3/4},q^{n+3/4}) \rightarrow$\\
$(\psi^{n+1},\ub^{n+1},q^{n+1})$ in four substeps\\
Substep 1:\\
Solve 
\begin{align}
	\begin{cases}
		\begin{cases}
			\frac{\partial \psi}{\partial t}+ \partial_{\phi} J_1(\psi)  \ni 0 \\
			\gamma_1\frac{\partial \ub}{\partial t}+\partial_{\vb}J_3(\psi,\ub)=\mathbf{0} \\
			\gamma_2\frac{\partial q}{\partial t}+\partial_{p}J_2(\psi,q)=0 
		\end{cases} \mbox{ in } \Omega\times (t^n,t^{n+1}),\\
		(\psi(t^n),\ub(t^n),q(t^n))=(\psi^{n},\ub^{n},q^{n}),
	\end{cases}
	\label{eq.split.1}
\end{align}
and set $\psi^{n+1/4}=\psi(t^{n+1}),\ub^{n+1/4}=\ub(t^{n+1}),q^{n+1/4}=q(t^{n+1})$.\\
Substep 2:\\
Solve 
\begin{align}
	\begin{cases}
		\begin{cases}
			\frac{\partial \psi}{\partial t}+\partial_{\phi} J_4(\psi,\ub)+ \partial_{\phi} J_5(\psi,q)\ni 0 \\
			\gamma_1\frac{\partial \ub}{\partial t} + \partial_{\vb}J_4(\psi,\ub) +\partial_{\vb}I_{\Sigma_1}(\ub,q)\ni\mathbf{0} \\
			\gamma_2\frac{\partial q}{\partial t}+\partial_{p} J_5(\psi,q) +\partial_{p}I_{\Sigma_1}(\ub,q) \ni 0 
		\end{cases} \mbox{ in } \Omega\times (t^n,t^{n+1}),\\
		(\psi(t^n),\ub(t^n),q(t^n))=(\psi^{n+1/4},\ub^{n+1/4},q^{n+1/4}),
	\end{cases}
	\label{eq.split.2}
\end{align}
and set $\psi^{n+2/4}=\psi(t^{n+1}),\ub^{n+2/4}=\ub(t^{n+1}),q^{n+2/4}=q(t^{n+1})$.\\
Substep 3:\\
Solve 
\begin{align}
	\begin{cases}
		\begin{cases}
			\frac{\partial \psi}{\partial t} =0 \\
			\gamma_1\frac{\partial \ub}{\partial t}+  \partial_{p} I_{\Sigma_2}(\ub)\ni \mathbf{0} \\
			\gamma_2\frac{\partial q}{\partial t}=0 
		\end{cases} \mbox{ in } \Omega\times (t^n,t^{n+1}),\\
		(\psi(t^n),\ub(t^n),q(t^n))=(\psi^{n+2/4},\ub^{n+2/4},q^{n+2/4}),
	\end{cases}
	\label{eq.split.3}
\end{align}
and set $\psi^{n+3/4}=\psi(t^{n+1}),\ub^{n+3/4}=\ub(t^{n+1}),q^{n+3/4}=q(t^{n+1})$.\\
Substep 4:\\
Solve
\begin{align}
	\begin{cases}
		\begin{cases}
			\frac{\partial \psi}{\partial t}  + \partial_{\phi} J_2(\psi,q) + \partial_{\phi} J_3(\psi,\ub)\ni 0 \\
			\gamma_1\frac{\partial \ub}{\partial t} =\mathbf{0}\\
			\gamma_2\frac{\partial q}{\partial t}=0 
		\end{cases} \mbox{ in } \Omega\times (t^n,t^{n+1}),\\
		(\psi(t^n),\ub(t^n),q(t^n))=(\psi^{n+3/4},\ub^{n+3/4},q^{n+3/4}),
	\end{cases}
	\label{eq.split.4}
\end{align}
and set $\psi^{n+1}=\psi(t^{n+1}),\ub^{n+1}=\ub(t^{n+1}),q^{n+1}=q(t^{n+1})$.

Our overall strategy of scheme (\ref{eq.split.1})--(\ref{eq.split.4}) is as follows: We first update $\psi$ using the distance function $f$ in Substep 1, which pushes $\psi$ towards the point cloud, and use the updated $\psi$ to update $\ub$ and $q$. Substep 2 focuses on the penalty terms and constraints between $\ub$ and $q$ in (\ref{eq.ivp}). Substep 3 uses the constraint $\Sigma_2$ to project $\ub$ to a unit vector. The updated $\ub$ and $q$ are used in Substep 4 to regularize $\psi$. There exists many other splitting strategies. For example, one can combine Substep 1 and 4 to get a three-step splitting scheme. In our experiments, scheme (\ref{eq.split.1})--(\ref{eq.split.4}) is more stable and gives better results.

Scheme (\ref{eq.split.1})--(\ref{eq.split.3}) is semi-constructive since one still needs to solve the four subproblems. To time discretize the scheme, we adopt the backward Euler method for each subproblem, leading to a Markchuk--Yanenko \cite{marchuk1990splitting,glowinski2003finite} type scheme:
\begin{align}
	&\begin{cases}
		\frac{\psi^{n+1/4}-\psi^n}{\Delta t} +\partial_{\phi} J_1(\psi^{n+1/4})   \ni 0,\\
		\gamma_1\frac{\ub^{n+1/4}-\ub^{n}}{\Delta t}+ \partial_{\vb} J_3(\psi^{n+1/4},\ub^{n+1/4})=\mathbf{0},\\
		\gamma_2\frac{q^{n+1/4}-q^{n}}{\Delta t}+ \partial_{p} J_2(\psi^{n+1/4},q^{n+1/4})=0,
	\end{cases}
	\label{eq.split.dis.1}\\
	&\begin{cases}
		\frac{\psi^{n+2/4}-\psi^{n+1/4}}{\Delta t}+\partial_{\phi} J_4(\psi^{n+2/4},\ub^{n+2/4})+ \partial_{\phi} J_5(\psi^{n+2/4},q^{n+2/4})  \ni 0,\\
		\gamma_1\frac{\ub^{n+2/4}-\ub^{n+1/4}}{\Delta t}+\partial_{\vb} J_4(\psi^{n+2/4},\ub^{n+2/4}) +\partial_{\vb}I_{\Sigma_1}(\ub^{n+2/4},q^{n+2/4}) \ni\mathbf{0},\\
		\gamma_2\frac{q^{n+2/4}-q^{n+1/4}}{\Delta t} + \partial_{p} J_5(\psi^{n+2/4},q^{n+2/4}) +\partial_{p}I_{\Sigma_1}(\ub^{n+2/4},q^{n+2/4})\ni 0,
	\end{cases}
	\label{eq.split.dis.2}\\
	&\begin{cases}
		\frac{\psi^{n+3/4}-\psi^{n+2/4}}{\Delta t} = 0,\\
		\gamma_1\frac{\ub^{n+3/4}-\ub^{n+2/4}}{\Delta t}+  \partial_{p} I_{\Sigma_4}(\ub^{n+3/4})\ni \mathbf{0},\\
		\gamma_2\frac{q^{n+3/4}-q^{n+2/4}}{\Delta t}=0,
	\end{cases}
	\label{eq.split.dis.3}\\
	&\begin{cases}
		\frac{\psi^{n+1}-\psi^{n+3/4}}{\Delta t}+ \partial_{\phi} J_2(\psi^{n+1},q^{n}) + \partial_{\phi} J_3(\psi^{n+1},\ub^n)\ni 0,\\
		\gamma_1\frac{\ub^{n+1}-\ub^{n+3/4}}{\Delta t}=\mathbf{0},\\
		\gamma_2\frac{q^{n+1}-q^{n+3/4}}{\Delta t}=0.
	\end{cases}
	\label{eq.split.dis.4}
\end{align}
In the rest of this section, we discuss solutions to each subproblem.
\subsection{On the solution to (\ref{eq.split.dis.1}) }
For $\psi^{n+1/4}$, one can derive that
\begin{align*}
	\partial_{\phi}J_1(\phi)= -\eta_0\delta_{\varepsilon}(\phi)\left[\nabla \cdot \left(f^2 \frac{\nabla \phi}{|\nabla \phi|}\right)\right].
\end{align*}

Note that the first equation in (\ref{eq.split.dis.1}) is a backward Euler discretization of 
\begin{align}
	\frac{\partial \psi}{\partial t} =\eta_0\delta_{\varepsilon}(\psi)\left[\nabla \cdot \left(f^2 \frac{\nabla \psi}{|\nabla \psi|}\right)\right].
	\label{eq.psi1.1}
\end{align}

Based on (\ref{eq.psi1.1}), we subtract a Laplacian term on both sides to get
\begin{align}
	\frac{\partial \psi}{\partial t} -\beta_1\nabla^2 \psi = -\beta_1\nabla^2 \psi +\eta_0\delta_{\varepsilon}(\psi)\left[\nabla \cdot \left( f^2 \frac{\nabla \psi}{|\nabla \psi|}\right)\right],
	\label{eq.psi1.2}
\end{align}
where $\beta_1>0$ is a constant. We use a semi--implicit method to time discretize (\ref{eq.psi1.2}): the left--hand side is treated implicitly and the right--hand side is treated explicitly
\begin{align}
	\psi^{n+1/4} -\Delta t\beta_1\nabla^2 \psi^{n+1/4} = \psi^n-\Delta t\beta_1\nabla^2 \psi^n +\Delta t\eta_0\delta_{\varepsilon}(\psi^n)\left[\nabla \cdot \left( f^2\frac{\nabla \psi^n}{|\nabla \psi^n|}\right)\right].
	\label{eq.psi1.3}
\end{align}
This technique is known as the frozen coefficient method and has been applied in image processing \cite{deng2019new,he2020curvature}.

For $\ub^{n+1/4}$, it solves
\begin{align*}
	\min_{\vb} \left[\frac{\gamma_1}{2} \int_{\Omega}|\vb-\ub^n|^2 d\xb - \Delta t\frac{\eta_2}{2} \int_{\Omega} r(\xb) |\pb_d\cdot \vb|^{2} \delta_{\varepsilon}(\psi^{n+1/4})|\nabla \psi^{n+1/4}|d\xb \right]
\end{align*}
and satisfies the optimality condition
\begin{align*}
	\gamma_1 \ub^{n+1/4} -\gamma_1 \ub^n-\Delta t\eta_2r(\xb)\delta_{\varepsilon}(\psi^{n+1/4})|\nabla \psi^{n+1/4}|(\pb_d\cdot \ub^{n+1/4})\pb_d=0,
\end{align*}
which is linear in $\ub^{n+1/4}$.
We have the closed-form solution
\begin{align}
	\ub^{n+1/4}=\left(\gamma_1I-\Delta t\eta_2r(\xb) \delta_{\varepsilon}(\psi^{n+1/4})|\nabla \psi^{n+1/4}|\pb_d\otimes \pb_d\right)^{-1} (\gamma_1 \ub^n),
	\label{eq.u.1}
\end{align}
where $(\pb_d\otimes\pb_d)\vb=(\pb_d\cdot \vb)\pb_d$.

For $q^{n+1/4}$, it solves
\begin{align*}
	\min_{p} \left[\frac{\gamma_2}{2} \int_{\Omega}|p-q^n|^2 d\xb + \Delta t\frac{\eta_1}{2} \int_{\Omega} |p|^{2}\delta_{\varepsilon}(\psi^{n+1/4})|\nabla \psi^{n+1/4}| d\xb\right].
\end{align*}

The optimality condition of $q^{n+1/4}$ is
\begin{align*}
	\gamma_2 q^{n+1/4} -\gamma_2 q^n+\Delta t \eta_1\delta_{\varepsilon}(\psi^{n+1/4})|\nabla \psi^{n+1/4}|q^{n+1/4}=0,
\end{align*}
which can be solved as
\begin{align}
	q^{n+1/4}=\frac{\gamma_2 q^n}{\gamma_2+\Delta t \eta_1\delta_{\varepsilon}(\psi^{n+1/4})|\nabla \psi^{n+1/4}|}.
	\label{eq.q.1}
\end{align}
\subsection{On the solution to (\ref{eq.split.dis.2})}
In (\ref{eq.split.dis.2}), $(\psi^{n+2/4},\ub^{n+2/4},q^{n+2/4})$ 
\begin{align}
	\min_{\phi,(\vb,p)\in \Sigma_1} & \bigg[\frac{1}{2} \int_{\Omega} |\phi-\psi^{n+1/4}|^2d\xb + \frac{\gamma_1}{2} \int_{\Omega} |\vb-\ub^{n+1/4}|^2d\xb + \frac{\gamma_2}{2} \int_{\Omega} |p-q^{n+1/4}|^2d\xb \nonumber\\
	&+ \frac{\Delta t \alpha_1}{2}\int_{\Omega} \left| \vb-\frac{\nabla \phi}{|\nabla \phi|}\right|^2d\xb + \frac{\Delta t\alpha_2}{2} \int_{\Omega} \left|p-\nabla \cdot \frac{\nabla \phi}{|\nabla \phi|}\right|^2d\xb \bigg],
	\label{eq.dis2.1.penal.1}
\end{align}
where $\alpha_1,\alpha_2>0$ are some constants. 
Denote $\nbb_{\Omega}$ as the outward normal of $\Omega$. Solving (\ref{eq.dis2.1.penal.1}) is equivalent to solve the following fourth-order PDE system (see a derivation in Appendix \ref{sec.variation.curvature}):
\begin{align}
	\begin{cases}
		\frac{\psi^{n+2/4}-\psi^{n+1/4}}{\Delta t}=\alpha_1\nabla\cdot \left(\frac{\nabla \psi^{n+2/4} \cdot \ub^{n+2/4}}{|\nabla \psi^{n+2/4}|^3}\nabla\psi^{n+2/4}-\frac{\ub^{n+2/4}}{|\nabla \psi^{n+2/4}|}\right) \\
		\hspace{3cm}- \alpha_2\bigg[\nabla\cdot \Big( \left(\frac{\nabla\psi^{n+2/4}}{|\nabla \psi^{n+2/4}|^3} \cdot \nabla \left(q^{n+2/4}-\nabla\cdot\frac{\nabla \phi}{|\nabla \psi^{n+2/4}|}\right)\right) \nabla\psi^{n+2/4}\\
		\hspace{3cm}- \frac{1}{|\nabla \psi^{n+2/4}|} \nabla \left(\nabla\cdot \ub^{n+2/4}-\nabla\cdot\frac{\nabla \psi^{n+2/4}}{|\nabla \psi^{n+2/4}|}\right)  \Big)\bigg],\\
		\gamma_1\frac{\ub^{n+2/4}-\ub^{n+1/4}}{\Delta t}+\alpha_1 \ub^{n+2/4}-\alpha_2\nabla(\nabla\cdot \ub^{n+2/4})=\alpha_1 \frac{\nabla \psi^{n+2/4}}{|\nabla \psi^{n+2/4}|}-\alpha_2\nabla\left( \nabla \cdot \frac{\nabla \psi^{n+2/4}}{|\nabla \psi^{n+2/4}|}\right)\\
		q^{n+2/4}=\nabla\cdot \ub^{n+2/4},\\
		\nabla \psi^{n+2/4}\cdot \nbb_{\Omega}=0 \mbox{ on } \partial \Omega,\\
		\ub^{n+2/4}\cdot \nbb_{\Omega}=0 \mbox{ on } \partial \Omega,\\
		\nabla\cdot \ub^{n+2/4}-\nabla\cdot\frac{\nabla \psi^{n+2/4}}{|\nabla \psi^{n+2/4}|}=0 \mbox{ on } \partial \Omega,\\
		\nabla \left(\nabla\cdot \ub^{n+2/4}-\nabla\cdot\frac{\nabla \psi^{n+2/4}}{|\nabla \psi^{n+2/4}|}\right)\cdot \nbb_{\Omega}=0 \mbox{ on } \partial \Omega,\\
		(\gamma_2+\Delta t\alpha_2) (\nabla \cdot \ub^{n+2/4})=\gamma_2 q^{n+1/4}+ \Delta t\alpha_2 \nabla \cdot \frac{\nabla \psi^{n+1/4}}{|\nabla \psi^{n+1/4}|} \mbox{ on } \partial \Omega.
	\end{cases}
	\label{eq.opti.fourth-order}
\end{align}
Directly solving (\ref{eq.opti.fourth-order}) is very challenging. In this paper, we consider an approximation of (\ref{eq.dis2.1.penal.1}) which can be solved explicitly. Since $\psi$ is a level set function, $|\nabla \phi|\neq 0$ almost everywhere. If we further assume $\psi$ is close to a signed distance function (this can be done by reinitialization, see Section \ref{sec.reinitial} for details), then $|\nabla \phi|\approx 1$. As a result, the functional in (\ref{eq.dis2.1.penal.1}) is continuous in $\psi$.
When $\Delta t$ is small, we expect $\psi^{n+2/4}$ stays close to $\psi^{n+1/4}$, and $\frac{\nabla \psi^{n+2/4}}{|\nabla \psi^{n+2/4}|}$ and $\nabla \cdot \frac{\nabla \psi^{n+2/4}}{|\nabla \psi^{n+2/4}|}$ stay close to $\frac{\nabla \psi^{n+1/4}}{|\nabla \psi^{n+1/4}|}$ and $\nabla \cdot \frac{\nabla \psi^{n+1/4}}{|\nabla \psi^{n+1/4}|}$, respectively. We approximate (\ref{eq.dis2.1.penal.1}) by
\begin{align}
	\min_{\phi,(\vb,p)\in \Sigma_3} & \Bigg[\frac{1}{2} \int_{\Omega} |\phi-\psi^{n+1/4}|^2d\xb + \frac{\gamma_1}{2} \int_{\Omega} |\vb-\ub^{n+1/4}|^2d\xb + \frac{\gamma_2}{2} \int_{\Omega} |p-q^{n+1/4}|^2d\xb \nonumber\\
	&+ \frac{\Delta t\alpha_1}{2}\int_{\Omega} \left| \vb-\frac{\nabla \psi^{n+1/4}}{|\nabla \psi^{n+1/4}|}\right|^2d\xb + \frac{\Delta t\alpha_2}{2} \int_{\Omega} \left|p-\nabla \cdot \frac{\nabla \psi^{n+1/4}}{|\nabla \psi^{n+1/4}|}\right|^2d\xb \Bigg].
	\label{eq.dis2.1.penal.2}
\end{align}
We thus have
\begin{align}
	\psi^{n+2/4}=&\psi^{n+1/4},\nonumber\\
	(\ub^{n+2/4},q^{n+2/4})=&\argmin_{(\vb,p)\in \Sigma_3} \Bigg[\frac{\gamma_1}{2} \int_{\Omega} |\vb-\ub^{n+1/4}|^2d\xb + \frac{\gamma_2}{2} \int_{\Omega} |p-q^{n+1/4}|^2d\xb  \nonumber\\
	&+ \frac{\Delta t\alpha_1}{2}\int_{\Omega} \left| \vb-\frac{\nabla \psi^{n+1/4}}{|\nabla \psi^{n+1/4}|}\right|^2d\xb + \frac{\Delta t\alpha_2}{2} \int_{\Omega} \left|p-\nabla \cdot \frac{\nabla \psi^{n+1/4}}{|\nabla \psi^{n+1/4}|}\right|^2d\xb \Bigg].
	\label{eq.dis2.2}
\end{align}
Equivalently, we have
\begin{align}
	\begin{cases}
		\ub^{n+2/4}=\argmin\limits_{\vb} \Bigg[\displaystyle\frac{\gamma_1}{2} \int_{\Omega} |\vb-\ub^{n+1/4}|^2d\xb + \frac{\gamma_2}{2} \int_{\Omega} |\nabla \cdot \vb-q^{n+1/4}|^2d\xb \\
		\hspace{2cm} + \displaystyle\frac{\Delta t\alpha_1}{2} \int_{\Omega} \left| \vb-\frac{\nabla \psi^{n+1/4}}{|\nabla \psi^{n+1/4}|}\right|^2d\xb + \frac{\Delta t\alpha_2}{2} \int_{\Omega} \left|\nabla \cdot \vb-\nabla \cdot \frac{\nabla \psi^{n+1/4}}{|\nabla \psi^{n+1/4}|}\right|^2d\xb \Bigg],\\
		q^{n+2/4}=\nabla\cdot \ub^{n+2/4}.
	\end{cases}
	\label{eq.dis2.3}
\end{align}
Note that $\ub^{n+2/4}$ is the weak solution of
\begin{align}
	\begin{cases}
		(\gamma_1+\Delta t\alpha_1)\ub^{n+2/4}- (\gamma_2 +\Delta t\alpha_2)\nabla(\nabla \cdot  \ub^{n+2/4})=  \\
		\hspace{2cm} \left(\gamma_1\ub^{n+1/4}+\Delta t\alpha_1 \frac{\nabla \psi^{n+1/4}}{|\nabla \psi^{n+1/4}|}\right) -\nabla \left(\gamma_2q^{n+1/4}+ \Delta t\alpha_2 \nabla\cdot \frac{\nabla \psi^{n+1/4}}{|\nabla \psi^{n+1/4}|}\right),\\
		(\gamma_2+\Delta t\alpha_2) (\nabla \cdot \ub^{n+2/4})=\gamma_2 q^{n+1/4}+ \Delta t\alpha_2 \nabla \cdot \frac{\nabla \psi^{n+1/4}}{|\nabla \psi^{n+1/4}|} \mbox{ on } \partial \Omega.
	\end{cases}
	\label{eq.step2.u}
\end{align}

After $\ub^{n+2/4}$ is solved, we compute $q^{n+2/4}=\nabla\cdot \ub^{n+2/4}$.

\subsection{On the solution to (\ref{eq.split.dis.3})}
In (\ref{eq.split.dis.3}), $\ub^{n+3/4}$ solves
\begin{align*}
	\min_{\vb\in \Sigma_4} \int_{\Omega} |\vb-\ub^{n+2/4}|^2d\bx,
\end{align*}
which can be solved explicitly as
\begin{align}
	\ub^{n+3/4}=\frac{\ub^{n+2/4}}{|\ub^{n+2/4}|}.
	\label{eq.step3.u}
\end{align}

\subsection{On the solution to (\ref{eq.split.dis.4})}
For $\psi^{n+1}$, one can derive that
\begin{align*}
	\partial_{\phi} J_2(\phi,p) + \partial_{\phi} J_3(\phi,\vb)= -\delta_{\varepsilon}(\phi)\left[\nabla \cdot \left(G \frac{\nabla \phi}{|\nabla \phi|}\right)\right]
\end{align*}
with 
\begin{align}
	G=\eta_1|q^{n+3/4}|^2- \eta_2r(\xb)(1-|\ub^{n+3/4}\cdot \pb_d|^2).
	\label{eq.psi4.G}
\end{align}

Note that (\ref{eq.split.dis.4}) is a backward Euler discretization of 
\begin{align}
	\frac{\partial \psi}{\partial t} =\delta(\psi)\left[\nabla \cdot \left(G \frac{\nabla \psi}{|\nabla \psi|}\right)\right].
	\label{eq.psi4.1}
\end{align}

Based on (\ref{eq.psi4.1}), we use the frozen coefficient method to get
\begin{align}
	\frac{\partial \psi}{\partial t} -\beta_2\nabla^2 \psi = -\beta_2\nabla^2 \psi +\delta(\psi)\left[\nabla \cdot \left( G \frac{\nabla \psi}{|\nabla \psi|}\right)\right],
	\label{eq.psi4.2}
\end{align}
where $\beta_2>0$ is a constant. Problem (\ref{eq.psi4.2}) is time discretized by a semi--implicit method as
\begin{align}
	\psi^{n+1} -\Delta t\beta_2\nabla^2 \psi^{n+1} = \psi^{n+3/4}-\Delta t\beta_2\nabla^2 \psi^{n+3/4} +\Delta t\left[\nabla \cdot \left( G\frac{\nabla \psi^{n+3/4}}{|\nabla \psi^{n+3/4}|}\right)\right].
	\label{eq.psi4.3}
\end{align}
Our method is summarized in Algorithm \ref{alg.1}.
\begin{algorithm}[t!]
	\caption{\label{alg.1}An operator-splitting method for solving problem (\ref{eq.energy.2})}
	\begin{algorithmic}
		\STATE {\bf Input:} Point cloud data $\cS$, weight function $r(\xb)$, PCA window size $\lambda$, model parameters $\eta_0,\eta_1,\eta_2$, time step $\Delta t$, algorithm parameters $\gamma_1,\gamma_2,\alpha_1,\alpha_2$ and frozen coefficients $\beta_1,\beta_2$.
		\STATE {\bf Initialization:} \\
		(i) Compute the distance function $f(\xb)$ by solving (\ref{eq.eikonal}). \\
		(ii) Compute vector field $\pb_d$ according to (\ref{eq.pbd}).\\
		(iii) Initialize $(\psi_0,\ub_0,q_0)$ according to Section \ref{sec.initial}. Set $(\psi^0,\ub^0,q^0)=(\psi_0,\ub_0,q_0)$
		\STATE {\bf Iterating}:
		\WHILE{not converge}
		\STATE 1. Solve (\ref{eq.split.dis.1}) using (\ref{eq.psi1.3}), (\ref{eq.u.1}) and (\ref{eq.q.1}) to obtain $(\psi^{n+1/4}, \ub^{n+1/4},q^{n+1/4})$.
		\STATE 2. Solve (\ref{eq.split.dis.2}) using (\ref{eq.dis2.2})-(\ref{eq.step2.u}) to obtain $(\psi^{n+2/4}, \ub^{n+2/4},q^{n+2/4})$.
		\STATE 3. Solve (\ref{eq.split.dis.3}) using (\ref{eq.step3.u}) to obtain $(\psi^{n+3/4}, \ub^{n+3/4},q^{n+3/4})$.
		\STATE 4. Solve (\ref{eq.split.dis.4}) using (\ref{eq.psi4.3}) to obtain $(\psi^{n+1}, \ub^{n+1},q^{n+1})$.
		\STATE 5. Set $n=n+1$.
		\ENDWHILE
		\STATE {\bf Output:} The converged level set function $\psi^*$.
	\end{algorithmic}
\end{algorithm}

\subsection{On the periodic boundary condition}
Problems (\ref{eq.psi1.3}), (\ref{eq.step2.u}), and (\ref{eq.psi4.3}) involve solving linear second-order PDEs, which can be efficiently addressed using the fast Fourier transform (FFT) when equipped with periodic boundary conditions. In fact, these problems can be readily modified to incorporate periodic boundary conditions, as demonstrated in \cite{deng2019new,liu2021color}. Throughout the remainder of this paper, we will consistently consider periodic boundary conditions and utilize FFT to enhance the efficiency of our algorithm.

\section{Numerical discretization}
\label{sec.discretization}
In this section, we discuss the discrete analog of the solution to each subproblem, considering periodic boundary conditions. We present the settings for two-dimensional space, noting that similar settings can be defined for three-dimensional space. Let $\Omega = [0,M] \times [0,N] \subset \mathbb{R}^2$, where $M$ and $N$ are positive integers. We discretize $\Omega$ with a spatial step of $\Delta x_1 = \Delta x_2 = 1$. For a function $v$ defined on $\Omega$, we denote $v(i,j)$ as $v_{i,j}$ for $i = 1, \ldots, M$ and $j = 1, \ldots, N$. Similarly, for a vector-valued function $\mathbf{u} = (u^1, u^2)^{\top}$, we denote $\mathbf{u}_{i,j} = (u_{i,j}^1, u_{i,j}^2)^{\top}$.

Define the forward and backward difference as
\begin{align*}
	&\partial_1^+ v_{i,j}=\begin{cases}
		v_{i+1,j}-v_{i,j} & 1\leq i \leq M-1,\\
		v_{1,j}-v_{M,j} & i=M,
	\end{cases} 
	&& \partial_2^+ v_{i,j}=\begin{cases}
		v_{i,j+1}-v_{i,j} & 1\leq j \leq N-1,\\
		v_{i,1}-v_{i,N} & j=N,
	\end{cases} \\
	&\partial_1^- v_{i,j}=\begin{cases}
		v_{1,j}-v_{M,j} & i=1,\\
		v_{i,j}-v_{i-1,j} & 2\leq i \leq M,
	\end{cases} 
	&& \partial_2^- v_{i,j}=\begin{cases}
		v_{i,1}-v_{i,N} & j=1,\\
		v_{i,j}-v_{i,j-1} & 2\leq j \leq N.
	\end{cases} 
\end{align*}
We also define the central difference as 
$$
\partial^c_1 v_{i,j}= \frac{1}{2} \left(\partial_1^+ v_{i,j} + \partial_1^- v_{i,j}\right), \quad \partial^c_2 v_{i,j}= \frac{1}{2} \left(\partial_2^+ v_{i,j} + \partial_2^- v_{i,j}\right).
$$
The discrete gradient and divergence operation are defined as
\begin{align*}
	&\nabla^{\pm} v_{i,j}= (\partial_1^{\pm} v_{i,j}, \partial_2^{\pm} v_{i,j})^{\top},\
	\nabla^{\pm}\cdot \ub_{i,j}= \partial_1^{\pm} u^1_{i,j} + \partial_2^{\pm} u^2_{i,j}, \\
	&\nabla^{c} v_{i,j}= (\partial_1^{c} v_{i,j}, \partial_2^{c} v_{i,j})^{\top},\
	\nabla^{c}\cdot \ub_{i,j}= \partial_1^{c} u^1_{i,j} + \partial_2^{c} u^2_{i,j}.
\end{align*}

Define the shift and identity operator as
\begin{align*}
	\cS_1^{\pm}v(i,j)=v(i\pm 1,j), \quad \cS_2^{\pm}v(i,j)=v(i,j\pm 1), \quad \cI v(i,j)=v(i,j).
\end{align*}
We denote $\cF$ and $\cF^{-1}$ as the Fourier transform and the inverse transform, respectively, and define
$$
z_i=\frac{2\pi}{M}(i-1), \quad z_j=\frac{2\pi}{N}(j-1)
$$
for $i=1,...,M$ and $j=1,...,N$.
We have
\begin{align*}
	&\cF(\cS_1^{\pm}v)(i,j)= (\cos z_i\pm \sqrt{-1} \sin(z_i)) \cF(v)(i,j),\\
	&\cF(\cS_2^{\pm}v)(i,j)= (\cos z_j\pm \sqrt{-1} \sin(z_j)) \cF(v)(i,j).
\end{align*}

\subsection{Computing the discrete analogue of $\psi^{n+1/4}$ and $\ub^{n+1/4}$}
For $\psi^{n+1/4}$, we discretize (\ref{eq.psi1.3}) as
\begin{align}
	\psi^{n+1/4} -&\Delta t\beta_1\nabla^-\cdot(\nabla^+ \psi^{n+1/4}) \nonumber\\
	&= \psi^n-\Delta t\beta_1\nabla^-\cdot(\nabla^+ \psi^n) +\Delta t\eta_0\delta_{\varepsilon}(\psi^n)\left[\nabla^c \cdot \left( f^2\frac{\nabla^c \psi^n}{|\nabla^c \psi^n|}\right)\right].
	\label{eq.psi1.3.dis}
\end{align}

Denote
\begin{align*}
	b=\psi^n-\Delta t \beta_1 \nabla^-\cdot(\nabla^+\psi^n)+\Delta t\eta_0 \delta_{\varepsilon}(\psi^n)\left( \partial_1^c \left(f^2\frac{ \partial_1^c \psi}{|\nabla^c \psi^n|}\right) +\partial_2^c \left(f^2 \frac{\partial_2^c \psi}{|\nabla^c \psi^n|}\right) \right).
\end{align*}
Function $\psi^{n+1/4}$ satisfies
\begin{align*}
	(1-\Delta t\beta_1(\partial_1^-\partial_1^+ + \partial_2^-\partial_2^+))\psi^{n+1/4}=b,
\end{align*}
which is equivalent to
\begin{align}
	(\cI-\Delta t\beta_1 (\cI-\cS_1^-)(\cS_1^+-\cI)- \Delta t\beta_1 (\cI-\cS_2^-)(\cS_2^+-\cI))\psi^{n+1/4}=b.
	\label{eq.psi1.3.dis.1}
\end{align}
Computing the Fourier transform of both sides in (\ref{eq.psi1.3.dis.1}) gives
\begin{align*}
	w\cF(\psi^{n+1/4})=\cF(b),
\end{align*}
where
\begin{align*}
	w(i,j)=1-\Delta t\beta_1(1-e^{-\sqrt{-1}z_i})(e^{\sqrt{-1}z_i}-1)-\Delta t\beta_1(1-e^{-\sqrt{-1}z_j})(e^{\sqrt{-1}z_j}-1).
\end{align*}
We compute
\begin{align}
	\psi^{n+1/4}=\mathrm{Real}\left[\cF^{-1}\left(\frac{\cF(b)}{w}\right)\right].
\end{align}

For $\ub^{n+1/4}$, we discretize (\ref{eq.u.1}) as
\begin{align}
	\ub^{n+1/4}=\left(\gamma_1I-\Delta t\eta_2 \delta_{\varepsilon}(\psi^{n+1/4})|\nabla^c \psi^{n+1/4}|\pb_d\otimes \pb_d\right)^{-1} (\gamma_1 \ub^n).
	\label{eq.u.1.dis}
\end{align}

Denote $\pb_d=[p_{d,1},p_{d,2}]^{\top}$. We have 
\begin{align*}
	\pb_d\otimes \pb_d=\pb_d\pb_d^{\top}=\begin{bmatrix}
		p_{d,1}^2 & p_{d,1}p_{d,2}\\
		p_{d,1}p_{d,2} & p_{d,2}^2
	\end{bmatrix}.
\end{align*}
Denote $M=\{m_{k\ell}\}_{k,\ell=1,2}$ with
\begin{align*}
	&m_{11}=\gamma_1-\Delta t \eta_2\delta_{\varepsilon}(\psi^{n+1/4})|\nabla^c \psi^{n+1/4}|p_{d,1}^2,\\
	&m_{12}=m_{21}=-\Delta t \eta_2\delta_{\varepsilon}(\psi^{n+1/4})|\nabla^c \psi^{n+1/4}|p_{d,1}p_{d,2},\\
	&m_{22}=\gamma_1-\Delta t \eta_2\delta_{\varepsilon}(\psi^{n+1/4})|\nabla^c \psi^{n+1/4}|p_{d,2}^2,
\end{align*}
and $\ub^n=[u^n_1,u^n_2]^{\top}$. 
We compute 
\begin{align}
	\ub^{n+1/4}=\frac{1}{m_{11}m_{22}-m_{12}^2} \begin{pmatrix}
		m_{22} u^n_1- m_{12}u^n_2\\ -m_{12} u^n_1+m_{11}u^n_2
	\end{pmatrix}.
\end{align}

For $q^{n+1/4}$, we discretize (\ref{eq.q.1}) as
\begin{align}
	q^{n+1/4}=\frac{\gamma_2 q^n}{\gamma_2+\Delta t \eta_1\delta_{\varepsilon}(\psi^{n+1/4})|\nabla^c \psi^{n+1/4}|}.
\end{align}

\subsection{Computing the discrete analogue of $\ub^{n+2/4}$ and $q^{n+2/4}$}
Denote 
\begin{align*}
	\sbb=\begin{pmatrix}
		s_1\\ s_2
	\end{pmatrix}
	=\left(\gamma_1\ub^{n+1/4}+\alpha_1 \frac{\nabla^c \psi^{n+1/4}}{|\nabla^c \psi^{n+1/4}|}\right) -\nabla^c \left(\gamma_2q^{n+1/4}+ \alpha_2\nabla^c\cdot \frac{\nabla^c \psi^{n+1/4}}{|\nabla^c\psi^{n+1/4}|}\right)
\end{align*}
Equation (\ref{eq.step2.u}) is discretized as
\begin{align}
	\begin{pmatrix}
		(\gamma_1+\alpha_1) -(\gamma_2+\alpha_2)\partial_1^+\partial_1^- & -(\gamma_2+\alpha_2)\partial_1^+\partial_2^- \\
		-(\gamma_2+\alpha_2)\partial_2^+\partial_1^- & (\gamma_1+\alpha_1) -(\gamma_2+\alpha_2)\partial_1^+\partial_1^-
	\end{pmatrix} \ub^{n+2/4}= \sbb,
	\label{eq.step2.u.dis.1}
\end{align}
which is equivalent to
\begin{align}
	A \ub^{n+2/4}= \sbb
	\label{eq.step2.u.dis.2}
\end{align}
with
\begin{align*}
	A=\begin{pmatrix}
		(\gamma_1+\alpha_1) -(\gamma_2+\alpha_2)(\cS_1^+-\cI)(\cI-\cS_1^-) & -(\gamma_2+\alpha_2)(\cS_1^+-\cI)(\cI-\cS_2^-) \\
		-(\gamma_2+\alpha_2)(\cS_2^+-\cI)(\cI-\cS_1^-) & (\gamma_1+\alpha_1) -(\gamma_2+\alpha_2)(\cS_2^+-\cI)(\cI-\cS_2^-)
	\end{pmatrix}.
\end{align*}

Computing the Fourier transform of (\ref{eq.step2.u.dis.2}) on both sides gives rise to
\begin{align}
	\begin{pmatrix}
		a_{11} & a_{12} \\ a_{21} & a_{22}
	\end{pmatrix}\cF 
	\begin{pmatrix}
		u^{n+2/4}_1 \\ u^{n+2/4}_2
	\end{pmatrix}= \cF \begin{pmatrix}
		s_1\\ s_2
	\end{pmatrix},
\end{align}
where
\begin{align*}
	&a_{11}=(\gamma_1+\alpha_1)-(\gamma_2+\alpha_2) (e^{\sqrt{-1}z_i}-1)(1-e^{-\sqrt{-1}z_i}), \nonumber\\
	&a_{12}=-(\gamma_2+\alpha_2 ) (e^{\sqrt{-1}z_i}-1)(1-e^{-\sqrt{-1}z_j}), \nonumber\\
	&a_{21}=-(\gamma_2+\alpha_2 ) (e^{\sqrt{-1}z_j}-1)(1-e^{-\sqrt{-1}z_i}), \nonumber\\
	&a_{22}=(\gamma_1+\alpha_1 )-(\gamma_2+\alpha_2) (e^{\sqrt{-1}z_j}-1)(1-e^{-\sqrt{-1}z_j}).\nonumber
\end{align*}
We compute $\ub^{n+2/4}$ as
\begin{align}
	\begin{pmatrix}
		u^{n+2/4}_1 \\ u^{n+2/4}_2
	\end{pmatrix}=
	{\rm Real}\left(\cF^{-1}\left[ \frac{1}{a_{11}a_{22}-a_{12}a_{21}} 
	\begin{pmatrix}
		a_{22}\cF(s_1)-a_{12}\cF(s_2) \\ -a_{21}\cF(s_1)+a_{11}\cF(s_2)
	\end{pmatrix}\right]\right).
\end{align}
After $\ub^{n+2/4}$ is computed, we compute $q^{n+2/4}=\nabla^c\cdot \ub^{n+2/4}$.

\subsection{Computing the discrete analogue of $\psi^{n+1}$}
We discretize (\ref{eq.psi4.3}) as
\begin{align}
	\psi^{n+1} -&\Delta t\beta_2\nabla^-\cdot(\nabla^+ \psi^{n+1}) \nonumber\\
	&= \psi^{n+3/4}-\Delta t\beta_2\nabla^-\cdot(\nabla^+ \psi^{n+3/4}) +\Delta t\left[\nabla^c \cdot \left( G\frac{\nabla^c \psi^{n+3/4}}{|\nabla^c \psi^{n+3/4}|}\right)\right],
	\label{eq.psi4.3.dis}
\end{align}
where $G$ is defined in (\ref{eq.psi4.G}).
Define 
\begin{align*}
	\bar{b}=\psi^{n+3/4}-\Delta t\beta_2\nabla^-\cdot(\nabla^+ \psi^{n+3/4}) +\Delta t\left[\nabla^c \cdot \left( G\frac{\nabla^c \psi^{n+3/4}}{|\nabla^c \psi^{n+3/4}|}\right)\right].
\end{align*} 
We rewrite (\ref{eq.psi4.3.dis}) as
\begin{align*}
	(1-\Delta t\beta_2(\partial_1^-\partial_1^+ + \partial_2^-\partial_2^+))\psi^{n+1/4}=\bar{b},
\end{align*}
which is equivalent to
\begin{align}
	(\cI-\Delta t\beta_2 (\cI-\cS_1^-)(\cS_1^+-\cI)- \Delta t\beta_2 (\cI-\cS_2^-)(\cS_2^+-\cI))\psi^{n+1/4}=\bar{b}.
\end{align}
Taking Fourier transform on both sides gives rise to
\begin{align}
	\bar{w}\cF(\psi^{n+1})=\cF(\bar{b})
\end{align}
with
\begin{align*}
	\bar{w}(i,j)=1-\Delta t\beta_2(1-e^{-\sqrt{-1}z_i})(e^{\sqrt{-1}z_i}-1)-\Delta t\beta_2(1-e^{-\sqrt{-1}z_j})(e^{\sqrt{-1}z_j}-1).
\end{align*}
We compute $\psi^{n+1}$ as
\begin{align}
	\psi^{n+1}=\mathrm{Real}\left[\cF^{-1}\left(\frac{\cF(\bar{b})}{\bar{w}}\right)\right].
\end{align}

\subsection{Extension of the discrete analogue computation to three-dimensional space}
In this subsection, we discuss the computation of the discrete analogs of $\mathbf{u}^{n+1/4}$ and $\mathbf{u}^{n+2/4}$ in three-dimensional space. The computation of the discrete analogs for other quantities can be directly extended to three-dimensional space.
\subsubsection{Computing the discrete analogue of $\ub^{n+1/4}$ in three-dimensional space}
In three dimensional space, we define 
\begin{align*}
	W=\pb_d\otimes \pb_d=\pb_d\pb_d^{\top}=\begin{bmatrix}
		p_{d,1}^2 & p_{d,1}p_{d,2} & p_{d,1}p_{d,3}\\
		p_{d,1}p_{d,2} & p_{d,2}^2 & p_{d,1}p_{d,3}\\
		p_{d,1}p_{d,3} & p_{d,2}p_{d,3} & p_{d,3}^2\\
	\end{bmatrix}.
\end{align*}
Denote $M=\{m_{ij}\}_{i,j=1,2,3}$ with
$$
m_{ij}=\begin{cases}
	\gamma_1-\Delta t\eta_2 \delta(\psi^{n+1/4})|\nabla^c \psi^{n+1/4}|w_{ij} & \mbox{ if } i=j,\\
	-\Delta t\eta_2 \delta(\psi^{n+1/4})|\nabla^c \psi^{n+1/4}|w_{ij} & \mbox{ otherwise},
\end{cases}
$$
where we used the notation $W=\{w_{ij}\}_{i,j=1,2,3}$.
We compute $\ub^{n+1/4}$ by solving (\ref{eq.u.1}) as
\begin{align}
	\ub^{n+1/4}=\frac{1}{\det(M)}\mathrm{Adj}(M)(\gamma_1\ub^n),
\end{align}
where $\mathrm{Adj}(M)$ is the adjoint matrix of $M$:
\begin{align*}
	\mathrm{Adj}(M)=\begin{pmatrix}
		m_{22}m_{33}-m_{23}m_{32} & m_{23}m_{31}-m_{21}m_{33} & m_{21}m_{32}-m_{22}m_{31}\\
		m_{13}m_{32}-m_{12}m_{33} & m_{11}m_{33}-m_{13}m_{31} & m_{12}m_{31}-m_{11}m_{32}\\
		m_{12}m_{23}-m_{13}m_{22} & m_{13}m_{21}-m_{11}m_{23} & m_{11}m_{22}-m_{12}m_{21}
	\end{pmatrix}^{\top}.
\end{align*}

\subsubsection{Computing the discrete analogue of $\ub^{n+2/4}$ in three-dimensional space}
We discretize (\ref{eq.step2.u}) as
\begin{align}
	A\ub^{n+2/4}=\sbb,
\end{align}
where $A=\{a_{ij}\}$ for $i,j=1,2,3$ with 
\begin{align*}
	a_{ij}=\begin{cases}
		(\gamma_1+\Delta t)-(\gamma_2+\Delta t) (e^{\sqrt{-1}z_{i}}-1)(1-e^{\sqrt{-1}z_{i}}) & \mbox{ if } i=j,\\
		-(\gamma_2+\Delta t) (e^{\sqrt{-1}z_{i}}-1)(1-e^{\sqrt{-1}z_{j}}) & \mbox{ if } i\neq j,
	\end{cases}
\end{align*}
and 
\begin{align*}
	\sbb=\begin{pmatrix}
		p_1\\ p_2\\ p_3
	\end{pmatrix}
	=\left(\gamma_1\ub^{n+1/4}+\Delta t \frac{\nabla^c \psi^{n+1/4}}{|\nabla^c \psi^{n+1/4}|}\right) -\nabla^c \left(\gamma_2q^{n+1/4}+ \Delta t \nabla^c\cdot \frac{\nabla^c \psi^{n+1/4}}{|\nabla^c \psi^{n+1/4}|}\right).
\end{align*}
The updating formula for $\ub^{n+2/4}$ reads as
\begin{align}
	\begin{pmatrix}
		u^{n+2/4}_1 \\ u^{n+2/4}_2 \\ u^{n+2/4}_3
	\end{pmatrix}=
	{\rm Real}\left(\cF^{-1}\left[ \frac{1}{\det(A)} 
	\mathrm{Adj}(A) \begin{pmatrix}
		p_1\\ p_2 \\ p_3
	\end{pmatrix}\right]\right).
\end{align}

\section{Implementation details}
We discuss the details of implementing our proposed method, which include the implementation of PCA, the computation of the distance function $f(\mathbf{x})$, the selection of the weight function $r(\mathbf{x})$, the choice of initial conditions, and the process of reinitialization.
\label{sec.implementation}
\subsection{Computation of principal components}
We utilize PCA to estimate the normal direction from $\cS$. Let $\Omega$ be our computational domain.  For any $\xb\in \Omega$, we denote the set of principal directions computed from $\cS$ by $\{\pb_i(\xb)\}_{i=1}^d$. Let $\cW(\xb)$ represent a local window centered at $\xb$. The directions $\widehat{\pb}_i(\xb)$ are eigenvectors of the following covariance matrix:
\begin{align}
	\sum_{\zb\in \cS\cap \cW(\xb)} (\zb-\bar{\zb})(\zb-\bar{\zb})^{\top} \quad \mbox{ with } \quad \bar{\zb}= \frac{1}{\#\{\cS\cap \cW(\xb)\}}\sum_{\zb\in \cS\cap \cW(\xb)} \zb,
\end{align}
where $\#\{\cS\cap \cW(\xb)\}$ denotes the number of elements in $\cS\cap \cW(\xb)$.
The directions $\{\mathbf{p}_i(\mathbf{x})\}_{i=1}^d$ are ordered in non-increasing order according to their corresponding eigenvalues. When $\mathbf{x} \in \Gamma^*$, $\mathbf{p}_d$ serves as an estimate of $\mathbf{n}^*$. Conversely, when $\mathbf{x} \notin \Gamma$, $\mathbf{p}_d$ represents a natural extension of the normal direction at $\mathbf{x}$.

The procedure outlined above may encounter two issues if $\#\{\mathcal{S} \cap \mathcal{W}(\mathbf{x})\}$ is too small. Firstly, if $\#\{\mathcal{S} \cap \mathcal{W}(\mathbf{x})\} = 0$, PCA cannot be performed, and $\mathbf{p}_d(\mathbf{x})$ cannot be calculated. This situation arises when $\mathbf{x}$ is located far from the point cloud $\mathcal{S}$. However, to solve problem (\ref{eq.energy.1}), it is essential for $\mathbf{p}_d(\mathbf{x})$ to be defined across the entire computational domain. Secondly, if $\#\{\mathcal{S} \cap \mathcal{W}(\mathbf{x})\}$ is too small, meaning there are insufficient data points within $\mathcal{W}(\mathbf{x})$, the estimated normal direction becomes highly inaccurate.

To address these difficulties, we define $\mathbf{p}_d(\mathbf{x})$ in the following manner. Let $c_p$ be a positive integer, and denote $\bar{\mathbf{x}}$ as the center of $\Omega$. We define
\begin{align}
	\pb_d(\xb)=
	\begin{cases}
		\frac{\xb-\bar{\xb}}{|\xb-\bar{\xb}|} & \mbox{ if } \#\{\cS\cap \cW(\xb)\}< c_p,\\
		\widehat{\pb}_d(\xb) & \mbox{ if } \#\{\cS\cap \cW(\xb)\}\geq c_p.
	\end{cases}
	\label{eq.pbd}
\end{align}
With $\mathbf{p}_d(\mathbf{x})$ defined as above, during the evolution of $\psi$ by solving (\ref{eq.ivp}) and when $\Gamma$ is distant from $\mathcal{S}$, the normal information term in (\ref{eq.energy.1}) compels $\Gamma$ to maintain a spherical shape. The evolution is primarily influenced by the distance term, which drives $\Gamma$ toward $\mathcal{S}$. As $\Gamma$ approaches $\mathcal{S}$, $\mathbf{p}_d(\mathbf{x})$ serves as the estimated normal direction, regularizing $\Gamma$ to be normal to it.

In this paper, we choose $\mathcal{W}(\mathbf{x})$ to be a square in the two-dimensional space and a cube in the three-dimensional space. We denote the window size (half of the edge length of the square or cube) by $\lambda$.

\subsection{Computing the distance function $f(\xb)$}
Given the point cloud $\cS$, we compute the distance function $f(\xb)$ in the computation domain $\Omega$ by solving the eikonal equation
\begin{align}
	\begin{cases}
		|\nabla f|=1 & \mbox{ in } \Omega,\\
		f(\xb)=0 & \mbox{ for } \xb\in \cS.
	\end{cases}
	\label{eq.eikonal}
\end{align}
Problem can be efficiently solved by the fast sweeping method developed based on the Lax-Friedrichs	Hamiltonian \cite{kao2004lax} or the Godunov Hamiltonians \cite{zhao2005fast}. In our experiments, the scheme from \cite{kao2004lax} is used.

\subsection{Choice of $r(\xb)$ and window size}
\label{sec.r}
In (\ref{eq.energy.1}), $r(\xb)$ controls the location dependent effect of the normal information term. We discuss the choice of $r(\xb)$ for three situations:
\begin{itemize}
	\item {\bf Complete clean data.} 
	When the given data is complete, i.e., every region of the manifold has corresponding samples in $\mathcal{S}$ and the data is free of noise, the distance term (the first term) effectively locates the surface, while the curvature term (the second term) aids in regularizing the reconstructed surface. In this scenario, the normal information term does not need to exert a strong influence. We select $r(\mathbf{x}) = 1$ with a small value for $\eta_2$. Since the data is free from noise, we can utilize a small window size to obtain accurate estimations of the normal direction.
	
	\item {\bf Complete noisy data.} When the data is complete but perturbed by noise, relying solely on the distance term may lead to oscillatory reconstructed surfaces. The normal information term can be employed to regularize the reconstructed surface. Therefore, we set $r(\mathbf{x}) = 1$ and choose a large value for $\eta_2$. Given the presence of noise, it is necessary to utilize a larger window to obtain a reliable estimate of the normal direction.
	
	\item {\bf Incomplete clean data.} When the data is incomplete but clean, the distance term is only reliable in the data-available regions. In contrast, in the data-missing regions, the reconstruction should primarily be guided by the normal information term. Therefore, we should select $r(\mathbf{x})$ such that it takes a small value in the data-available regions and a large value in the data-missing regions. In this paper, we propose choosing $r(\mathbf{x}) = \sqrt{f(\mathbf{x})}$. Regarding the window size, it is preferable to use a larger window so that the point cloud information can be propagated to a broader region via PCA.
\end{itemize}

\subsection{Initial condition}
\label{sec.initial}
A straightforward choice for the initial condition is to use a hypercube or hypersphere that encloses all the data. A more sophisticated initial condition can be established by thresholding the distance function $f(\mathbf{x})$. Consider the set $\{\mathbf{x} : f(\mathbf{x}) = c\}$ for some $c > 0$. This set consists of two layers: the outer layer encloses all the data, while the inner layer resides within the point cloud. Both layers are close to the point cloud when $c$ is small, and we can utilize the outer layer as the initial condition. This strategy has been employed in \cite{zhao2001fast}. However, when the data is incomplete, this method may be ineffective because, in the data-missing regions, $f(\mathbf{x})$ is typically large. If $c$ is too small, the outer layer may connect with the inner layer.

In our experiments, we utilize the hypercube as the initial condition, which proves effective for all cases. We also require that the initial level set function $\psi^0$ be a signed distance function. Once $\psi^0$ is initialized, we initialize
$$
\ub_0=\frac{\nabla^c \psi^0 }{|\nabla^c \psi^0|}, \quad q_0=\nabla^c\cdot \ub_0.
$$

\subsection{Reinitialization}
\label{sec.reinitial}
During the evolution, $\psi$ may exhibit very steep or flat gradients near its zero level set, which can render our algorithm unstable or inaccurate. One approach to address this issue is to maintain $\psi$ as a signed distance function throughout the evolution. This can be achieved through a reinitialization process \cite{osher2004level}. By introducing an artificial time $\tau$, we reinitialize $\psi^n$ at the $n$-th iteration by solving
\begin{align}
	\begin{cases}
		\phi_{\tau}+\sign(\psi^n)(|\nabla \phi|-1)=0 & \mbox{ in } \Omega,\\
		\phi(0)=\psi^n
	\end{cases}
	\label{eq.reinitial}
\end{align}
until steady state, where $\sign(\psi^n)$ denotes the sign of $\psi^n$. Subsequently, we update $\psi^n$ with the steady state solution. In practice, since we are primarily concerned with the zero level set of $\psi^n$, solving (\ref{eq.reinitial}) for just a few iterations is sufficient.
\section{Numerical experiments}
\label{sec.experiments}
We demonstrate the effectiveness of the proposed method through experiments conducted in two-dimensional and three-dimensional spaces. In all experiments, unless otherwise specified, we set $\varepsilon = 1$ in the approximation (\ref{eq.delta.approx}) for the Delta function. Additionally, throughout the experiments, we use $\beta_1 = \beta_2 = 0.1$ as the frozen coefficients in (\ref{eq.psi1.3}) and (\ref{eq.psi4.3}).

\begin{figure}[t!]
	\centering
	\begin{tabular}{ccc}
		(a) Data & (b) Our results & (c) Energy\\
		\includegraphics[width=0.2\textwidth]{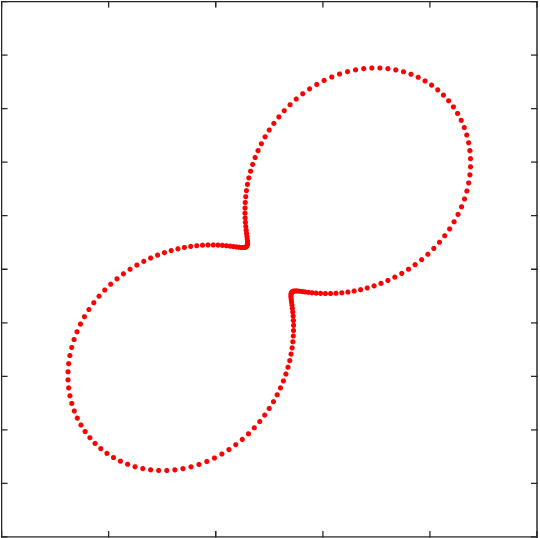} & 
		\includegraphics[width=0.2\textwidth]{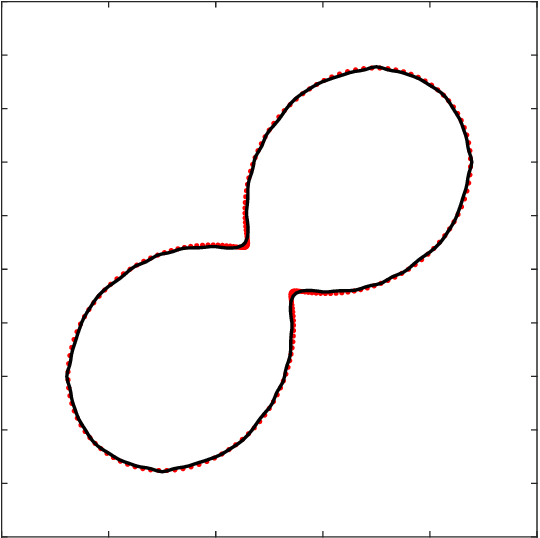}&
		\includegraphics[height=0.2\textwidth]{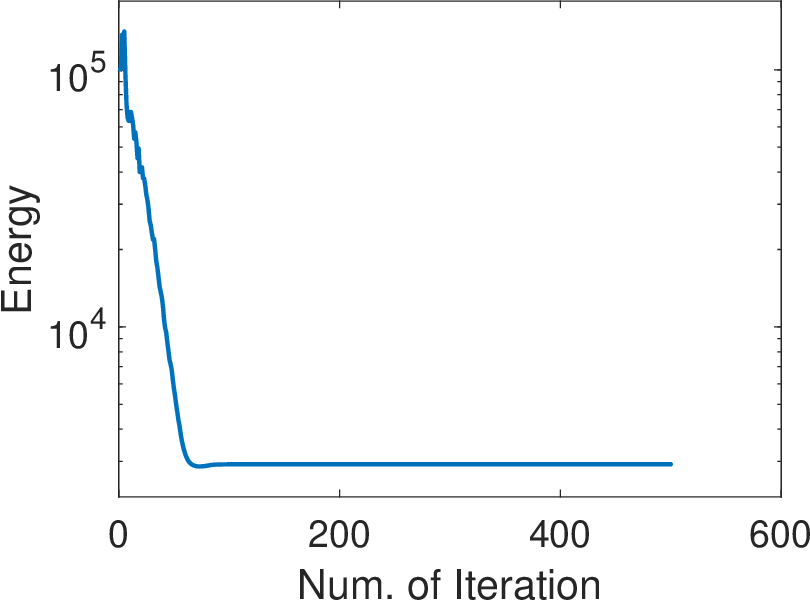} \\
		\includegraphics[width=0.2\textwidth]{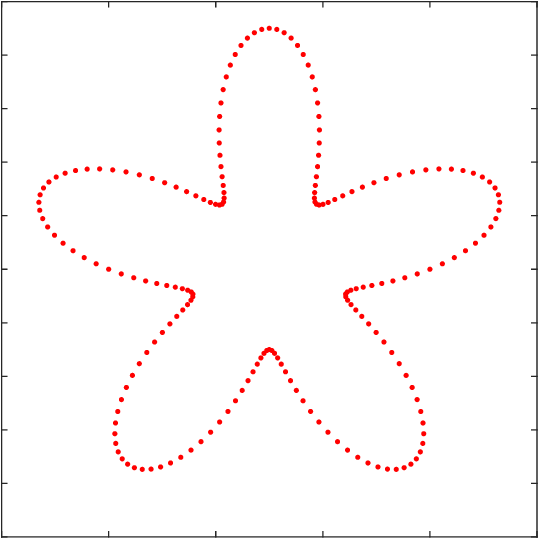} & \includegraphics[width=0.2\textwidth]{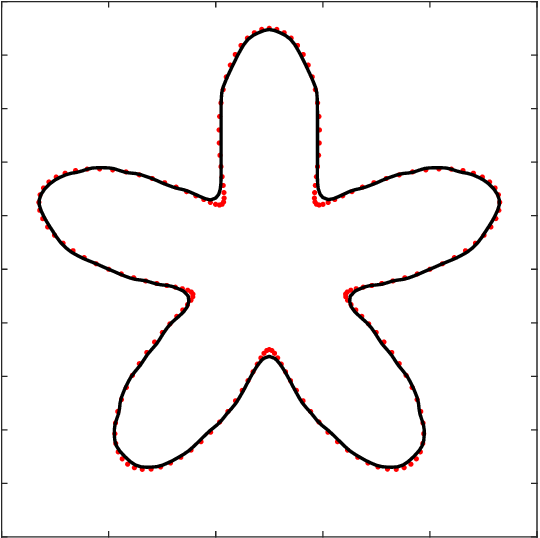} & 
		\includegraphics[height=0.2\textwidth]{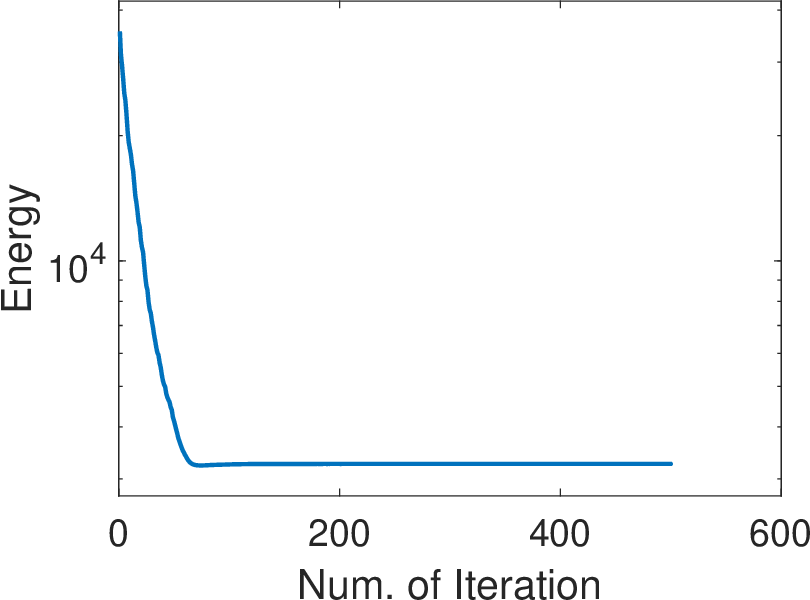} \\
		\includegraphics[width=0.2\textwidth]{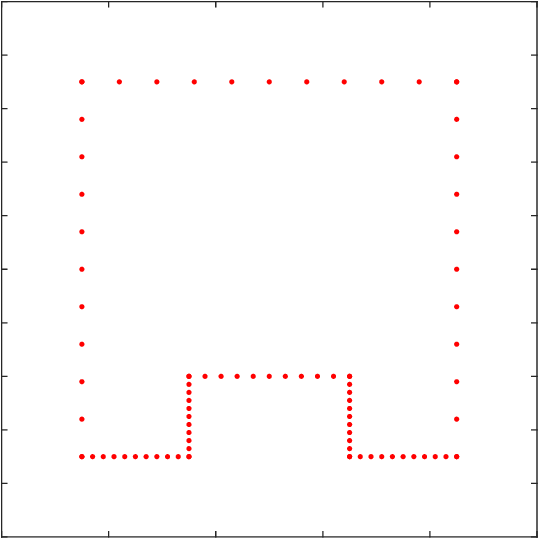} &
		\includegraphics[width=0.2\textwidth]{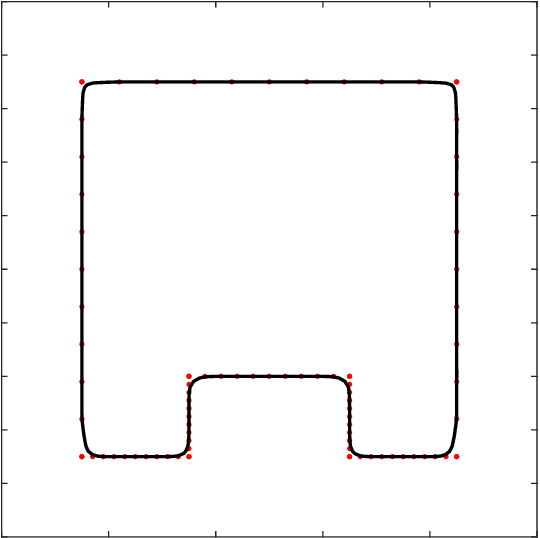}  &
		\includegraphics[height=0.2\textwidth]{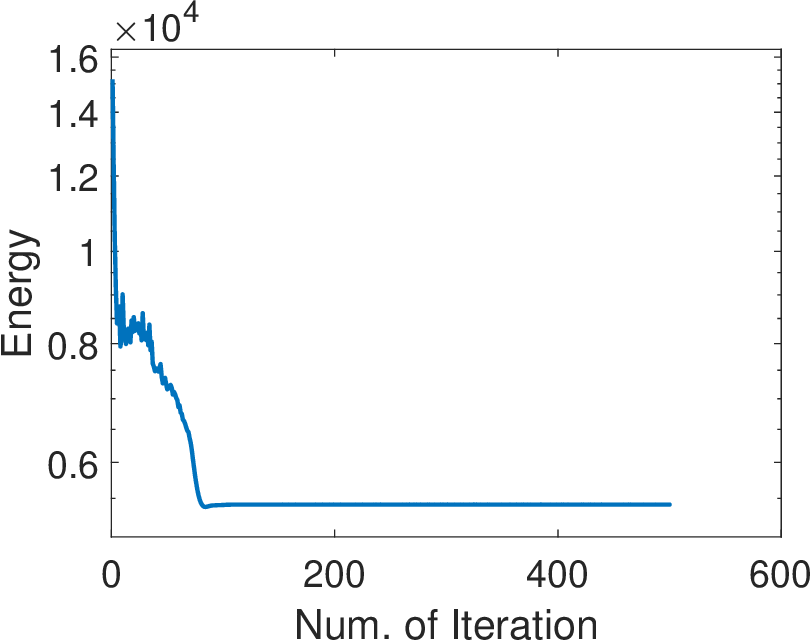} 
	\end{tabular}
	\caption{Two-dimensional examples. (a) Shows the point cloud data. (b) Shows our reconstructed curve. (c) Shows the evolution of the energy in (\ref{eq.energy.2}).}
	\label{fig.2d.clean}
\end{figure}

\subsection{Two-dimensional experiments}
In all two-dimensional experiments, we use the computational domain $\Omega = [0, 100] \times [0, 100]$ with grid space $\Delta x_1=\Delta x_2=1$. After each iteration, we reinitialize $\psi^n$ by solving (\ref{eq.reinitial}) for three iterations. In all experiment, we set $\gamma_1 = \gamma_2 = 100$ and $\alpha_1 = \alpha_2 = 4\gamma_1/\Delta t$.
\subsubsection{General performance}
We first evaluate the proposed method on complete and clean data. We set the parameters as follows: $\eta_0 = 1$, $\eta_1 = 2$, $\eta_2 = 1$, with a window size of 4 for PCA and $\Delta t = 0.5$. Given that the data is complete, we use $r(\mathbf{x}) = 1$. We consider three examples, as illustrated in Figure \ref{fig.2d.clean}(a). The point cloud size for the three examples is 200, 200 and 88 from top to bottom. Our reconstructed curves are displayed in Figure \ref{fig.2d.clean}(b). The proposed method effectively reconstructs the underlying curve of each point cloud, capturing detailed structures. Additionally, we present the evolution of the energy defined in (\ref{eq.energy.2}) in Figure \ref{fig.2d.clean}(c). For all three examples, our algorithm successfully minimizes the energy and converges within 100 iterations.

\begin{figure}[t!]
	\centering
	\begin{tabular}{cccccc}
		& (a) Square & & (b) Hexagon & \\
		&\includegraphics[height=0.17\textwidth]{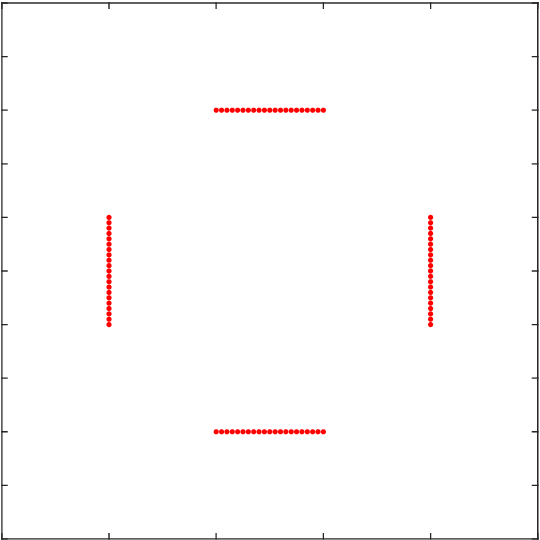} & &
		\includegraphics[height=0.17\textwidth]{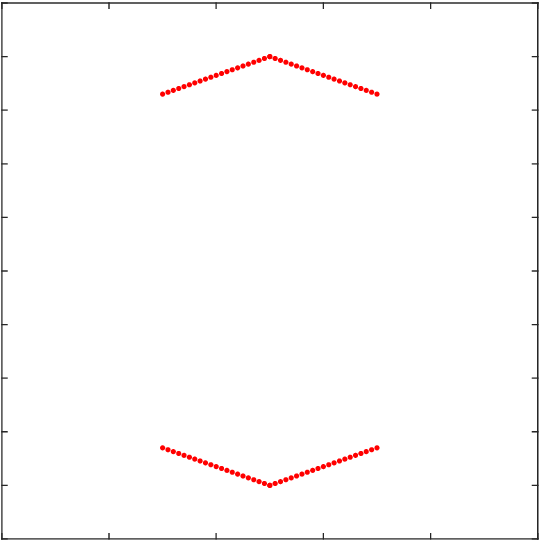} \\
		(c) Our result &(d) DS &(e) CR & (f)  DSP & (g) TVG\\
		\includegraphics[height=0.17\textwidth]{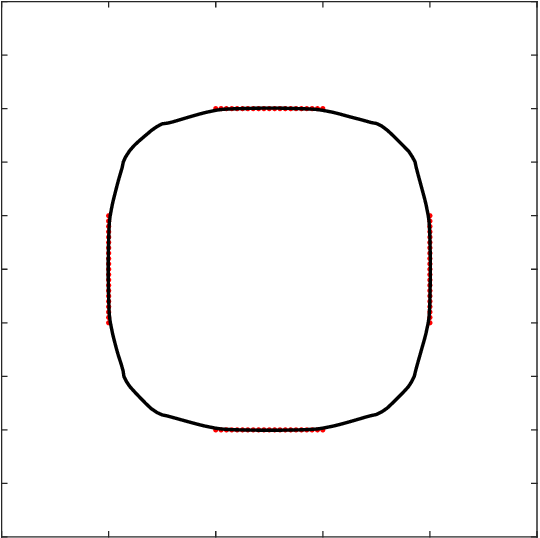} &
		\includegraphics[height=0.17\textwidth]{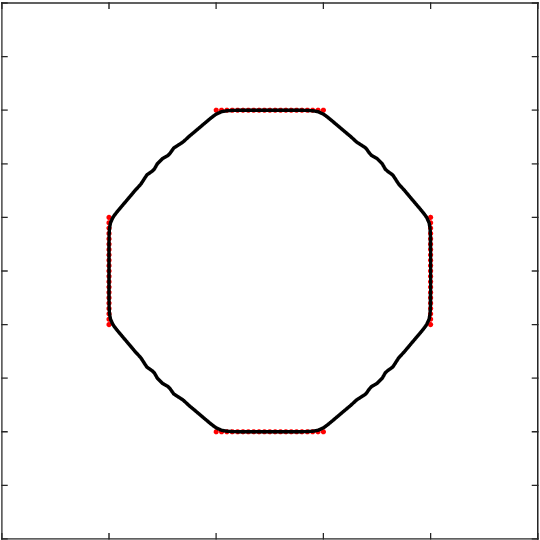} &
		\includegraphics[height=0.17\textwidth]{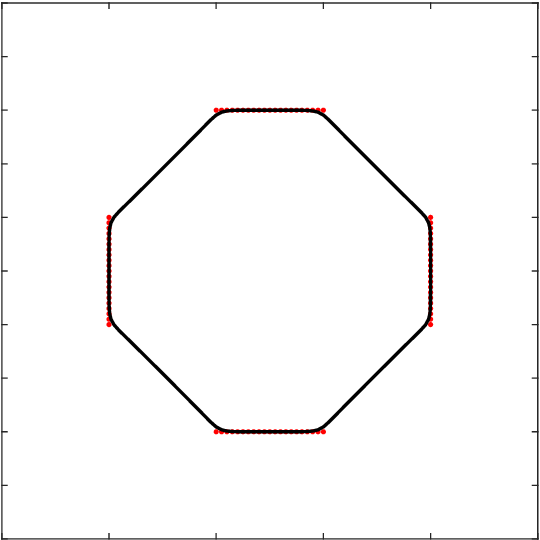}&
		\includegraphics[height=0.17\textwidth]{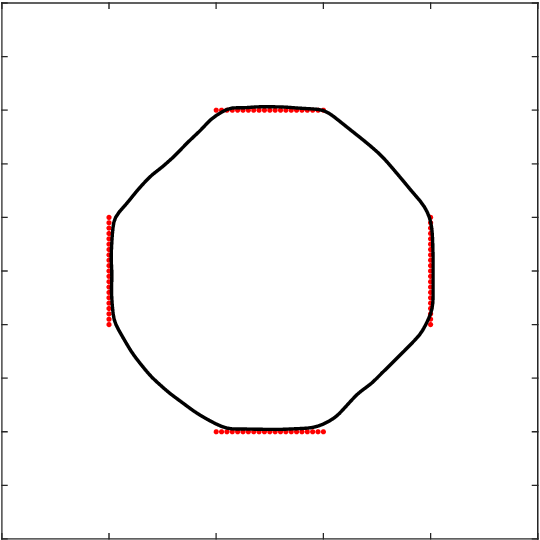} &
		\includegraphics[height=0.17\textwidth]{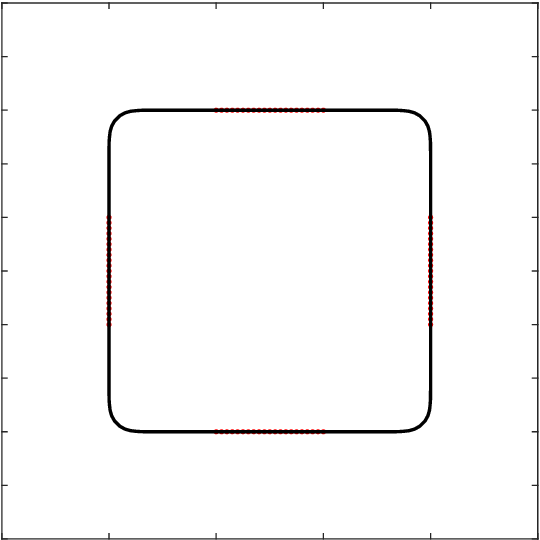}\\
		
		\includegraphics[height=0.17\textwidth]{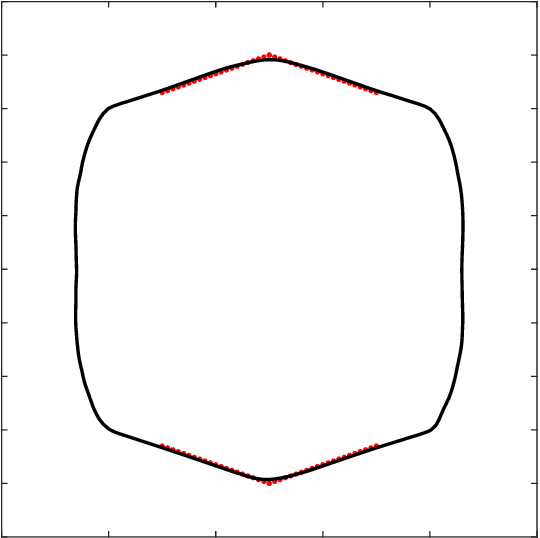} &
		\includegraphics[height=0.17\textwidth]{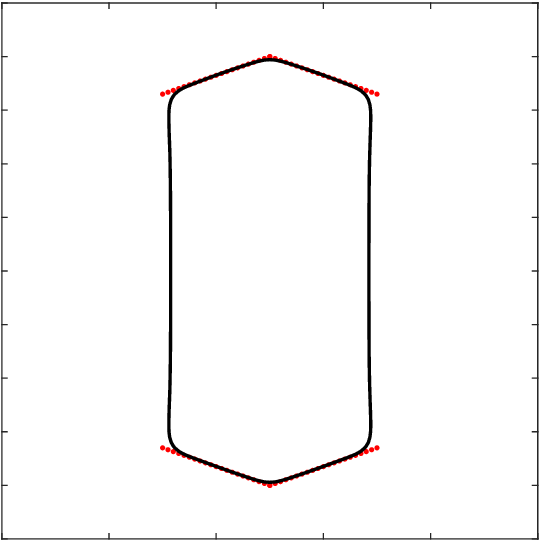}&
		\includegraphics[height=0.17\textwidth]{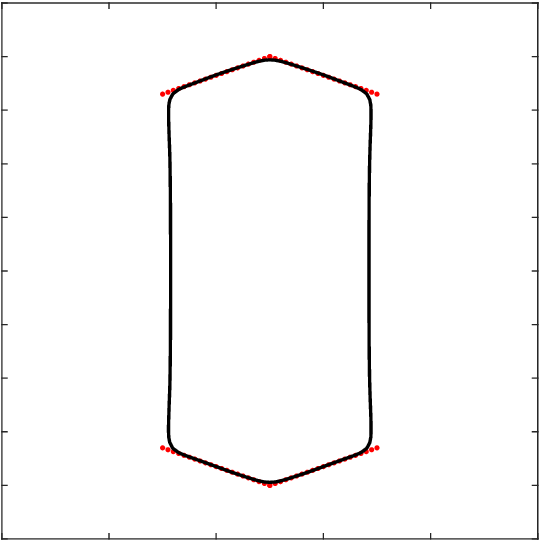}&
		\includegraphics[height=0.17\textwidth]{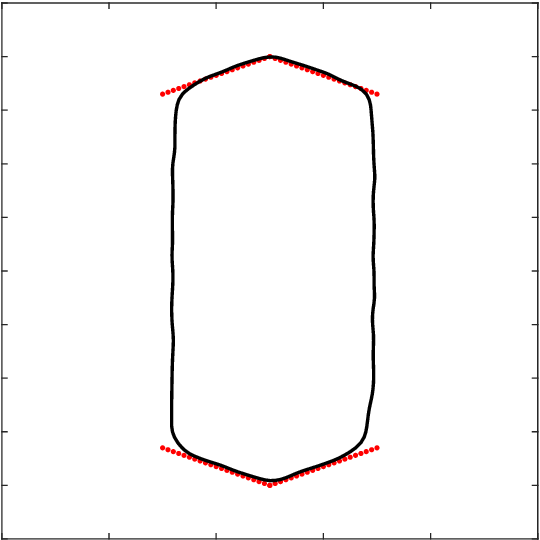} &
		\includegraphics[height=0.17\textwidth]{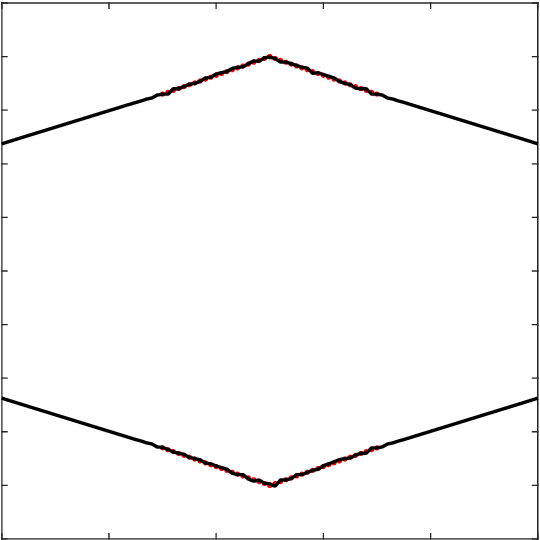}\\
		
	\end{tabular}
	\caption{Comparison on two-dimensional incomplete data. The first row shows the given data: (a) Shows data sampled from a square, while data around the four corners are missing. (b) shows data sampled from a hexagon. But only data near the two opposite corners are available. Column (c)-(g) show results by the proposed method, and the methods from DS \cite{zhao2000implicit}, CR \cite{he2020curvature}, DSP \cite{estellers2012efficient} and TVG \cite{liang2013robust}, respectively.}
	\label{fig.incomplete}
\end{figure}

\begin{table}
	\centering
	\begin{tabular}{c|c|c|c|c|c}
		\hline
		& Ours & DS & CR & DSP & TVG\\
		\hline
		Incomplete Square& 7.98s & 3.26s & 3.58s & 2.18s & 1.05s\\
		\hline
		Incomplete Hexagon & 15.18s & 6.50s & 7.15s & 2.18s & 0.98\\
		\hline
	\end{tabular}
	\caption{Comparison on two-dimensional incomplete data. CPU time used to compute results in Figure \ref{fig.incomplete}.}
	\label{tab.incomplete.cpu}
\end{table}

\begin{figure}[t!]
	\centering
	\begin{tabular}{cc}
		\includegraphics[width=0.3\textwidth]{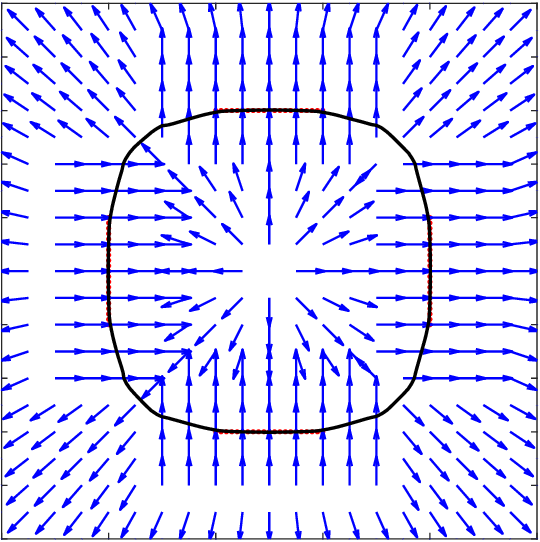} &
		\includegraphics[width=0.3\textwidth]{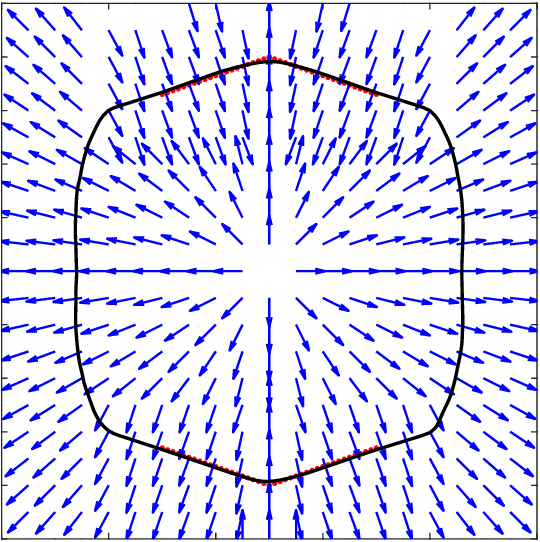} 
	\end{tabular}
	\caption{Two-dimensional incomplete data. The two plots show the vector field $\pb_d$ for the results by the proposed method in Figure \ref{fig.incomplete}(b).}
	\label{fig.incomplete.field}
\end{figure}

\subsubsection{Incomplete data}
We evaluate and compare our results on incomplete data. In the proposed algorithm, due to the sparsity of the data, it is essential to guide our solution primarily through the curvature and normal information regularizers in the data-missing regions, necessitating large values for $\eta_1$ and $\eta_2$. In our experiments, we set $r(\mathbf{x}) = \sqrt{f(\mathbf{x})}$, with parameters $\eta_0=10$, $\eta_1 = 2 \times 10^4$, $\eta_2 = 8 \times 10^4$, $\Delta t = 2 \times 10^{-4}$, and $\beta_1 = \beta_2 = 0.1$. 
We compare our method with the distance model (DS) \cite{zhao2000implicit}, a distance-preserving algorithm (DSP) \cite{estellers2012efficient}, a curvature-regularized model (CR) \cite{he2020curvature}, and a weighted total variation model (TVG) \cite{liang2013robust}. Among these methods, DS only uses the distance function $f(\xb)$ to reconstruct the surface. Based on DS, CR introduces a surface curvature terms as regularizer, which helps reconstruct surface corners. With level set formulation, in order to  stabilize the algorithm, DS and CR need reinitialization steps to make the surface level set function stay a signed distance function. DSP incorporates this property as a constraint of the energy. Its algorithm does not need reinitialization. TVG extends total variation based image segmentation models to surface reconstruction. The method uses anisotropic Gaussian and K-nearest neighbors to compute the edge deterctor function, which serves as the weight for the total variation term.  In our experiments, we test various choices of hyperparameters of these methods and present those yielding the best visual results.

We assess all methods using two examples. In the first example, the data are sampled from the four edges of a square, while data around the four corners are missing. In the second example, the data are sampled from a hexagon, with only the points near two opposite corners available. The given data for both examples are visualized in Figure \ref{fig.incomplete} (a) and (b). Our objective is to reconstruct the square and hexagon from this incomplete dataset. The results from these four methods are presented in Figure \ref{fig.incomplete} Column (c)-(g). In this comparison, the proposed method successfully reconstructs the square and hexagon, demonstrating the advantages of the normal information term. TVG reconstructs a better square. But it reconstructs two open curves for the hexagon example. This is because the performance of TVG highly depends on the initial guess of the curve. TVG uses data outward normal information to construct a curve that separate the computational domain into a region inside the curve and another region outside of the curve. This curve is constructed by linearly extending curves in data-available region to data-missing region. When two extended curves meet, a corner will be produced. Thus TVG is suitable for the square example in Figure \ref{fig.incomplete}(a). But for the hexagon, the extended curves meet at some points outside of the computational domain, making to the initial guess and result of TVG being two open curves. Compared to TVG, the proposed method only extends curves from data-available region to data-missing region to certain level instead of to infinity. All other methods yield results that merely connect the available data points.

We compare in Table \ref{tab.incomplete.cpu} the CPU time used for all methods to achieve results in Figure \ref{fig.incomplete}. To achieve convergence, for the incomplete square example, we use 500 iterations for the proposed method, DS and CR, 1000 iterations for DSP, and 50 iterations for TVG. For the incomplete hexagon example, we use 1000 iterations for the propose method, DS, CR and DSP, and 50 iterations for TVG. The TVG method is the fastest one. Due to the increased model complexity, the CPU time used by the proposed method doubles that of DS and CR.

To visualize the effects of the normal information term, we plot the normal vector field $\mathbf{p}_d$ for both examples in Figure \ref{fig.incomplete.field}. The reconstructed curves are nearly normal to $\mathbf{p}_d$.

\begin{figure}[t!]
	\centering
	\begin{tabular}{ccccc}
		& (a) Ellipse & & (b) Flower\\
		&\includegraphics[width=0.17\textwidth]{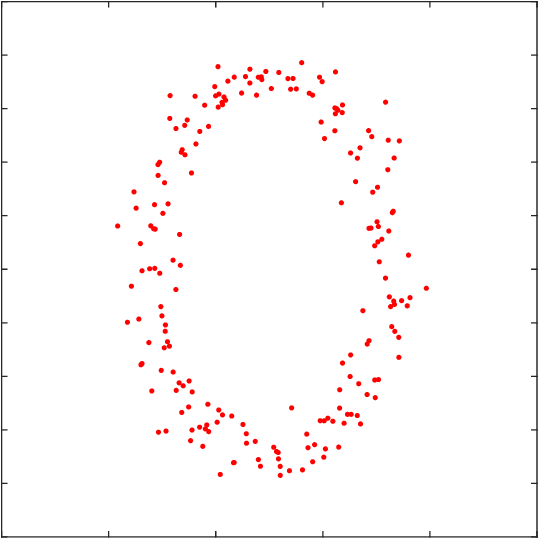} & &
		\includegraphics[width=0.17\textwidth]{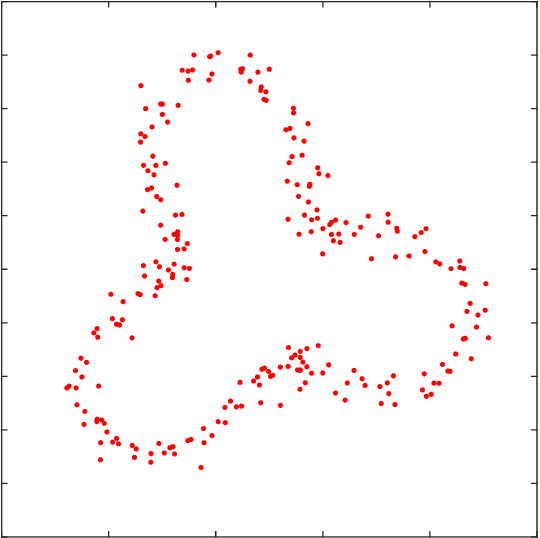} \\
		(c) Our result & (d) DS & (e) CR & (f)  DSP & (g) TVG\\
		\includegraphics[width=0.17\textwidth]{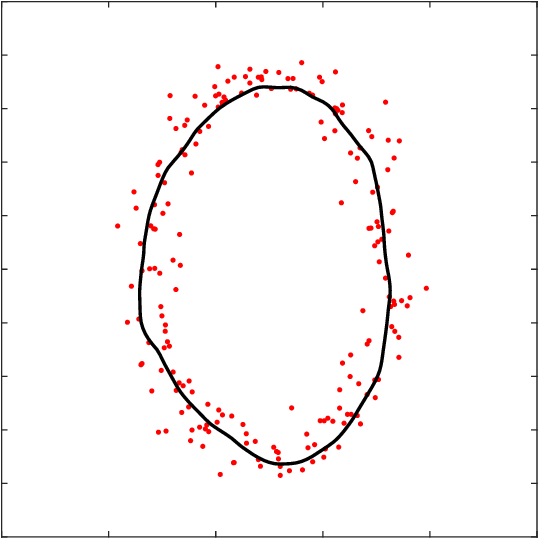} &
		\includegraphics[width=0.17\textwidth]{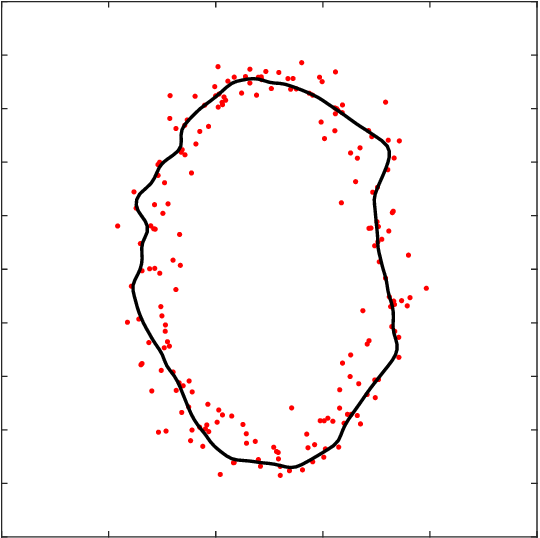} &
		\includegraphics[width=0.17\textwidth]{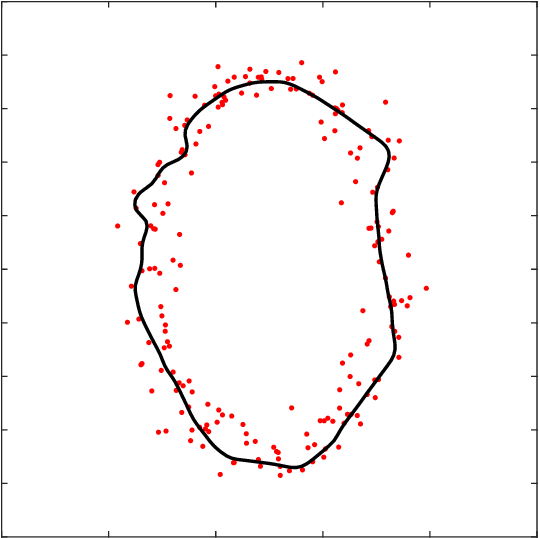}&
		\includegraphics[width=0.17\textwidth]{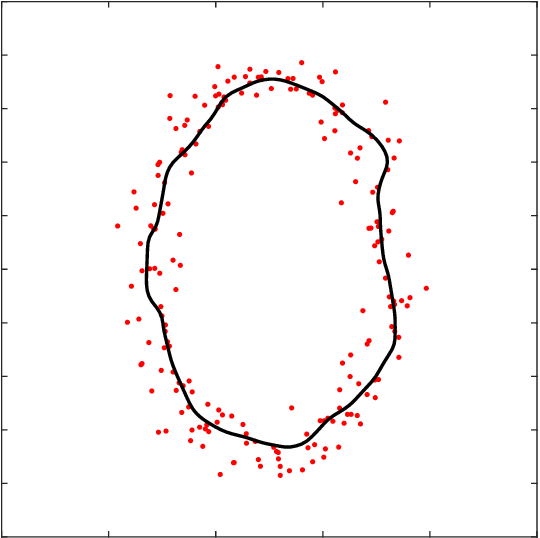}&
		\includegraphics[width=0.17\textwidth]{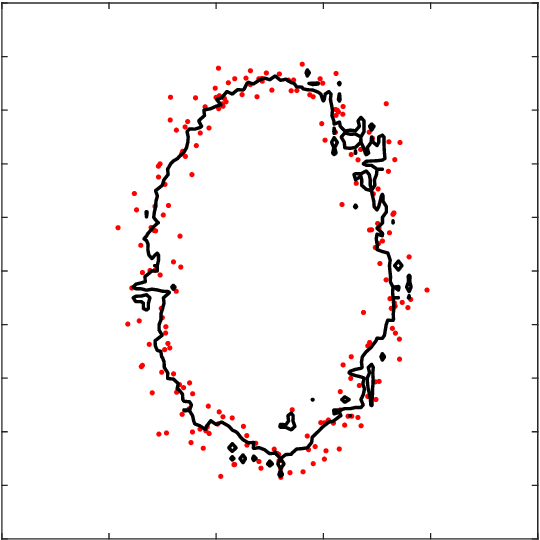}\\
		
		\includegraphics[width=0.17\textwidth]{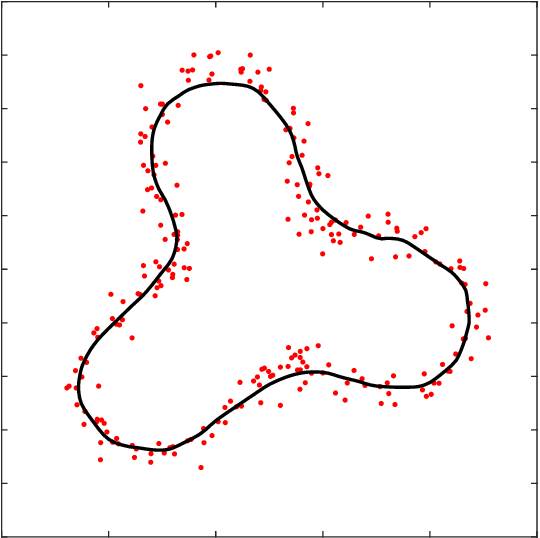} &
		\includegraphics[width=0.17\textwidth]{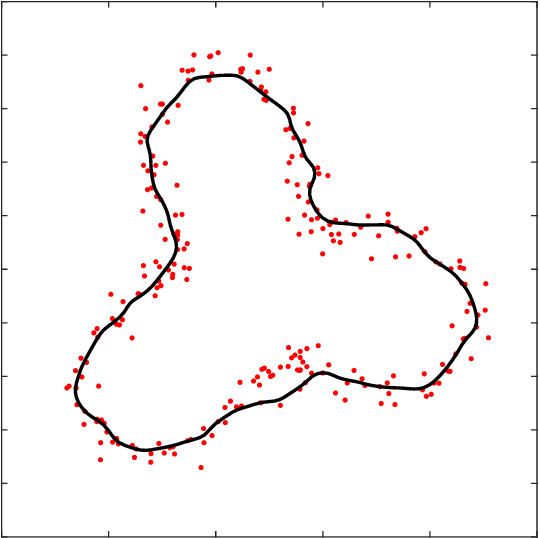} &
		\includegraphics[width=0.17\textwidth]{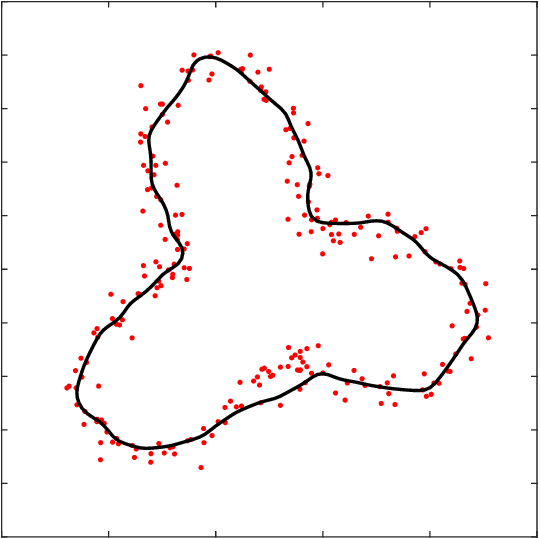}&
		\includegraphics[width=0.17\textwidth]{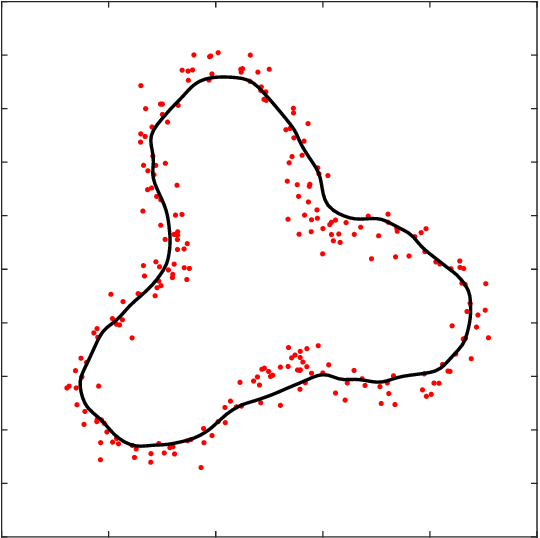}&
		\includegraphics[width=0.17\textwidth]{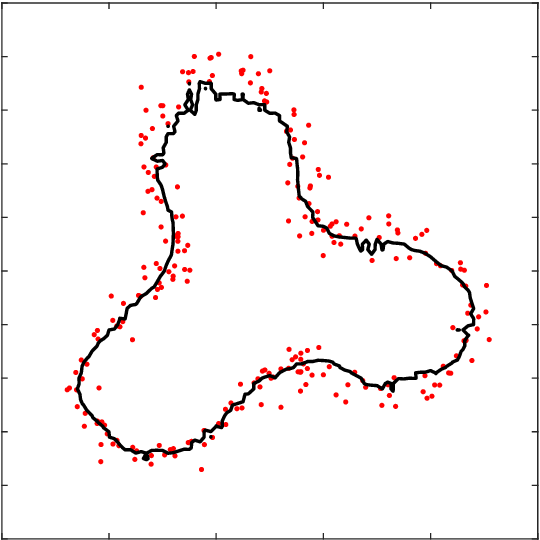}
	\end{tabular}
	\caption{Comparison on two-dimensional noisy data. The first row shows the given data: (a) Shows data sampled from an ellipse with noise. (b) shows data sampled from a a three-petal flower with noise. Column (c)-(g) show results by the proposed method, and the methods from DS \cite{zhao2000implicit}, CR \cite{he2020curvature}, DSP \cite{estellers2012efficient} and TVG \cite{liang2013robust}, respectively.
	}
	\label{fig.noisy}
\end{figure}

\begin{figure}[t!]
	\centering
	\begin{tabular}{cc}
		\includegraphics[width=0.3\textwidth]{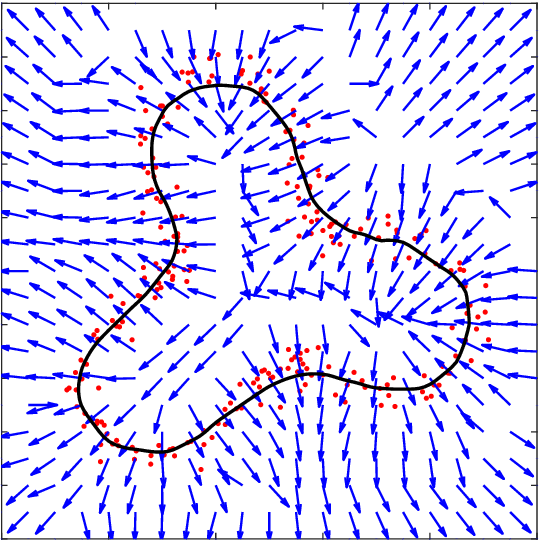} &
		\includegraphics[width=0.3\textwidth]{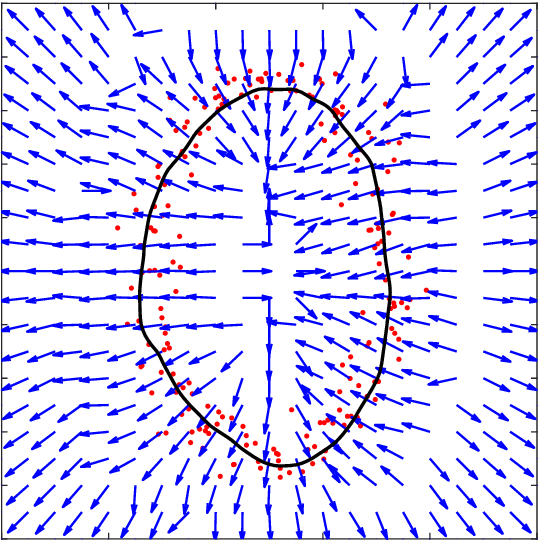} 
	\end{tabular}
	\caption{Two-dimensional noisy data. The two plots show the vector field $\pb_d$ for the results by the proposed method in Figure \ref{fig.noisy}(b).}
	\label{fig.noisy.field}
\end{figure}

\subsubsection{Noisy data}
\label{sec.2d.noisy}
We evaluate and compare the proposed method with other methods from \cite{zhao2000implicit,he2020curvature,estellers2012efficient} on a noisy dataset. We set $r(\mathbf{x}) = 1$. With a large value of $\eta_2$, our results may become detached from the point cloud. To address this issue, we adopt a two-stage strategy. In the first stage, we use a small $\eta_2$ to evolve the solution close to the point cloud. In the second stage, we increase $\eta_2$ to regularize the solution. Specifically, for both stages, we set $\eta_0=50$, $\eta_1 = 10^3$. In the first stage, we set $\eta_2 = 10^4$. In the second stage, we set $\eta_2 = 3 \times 10^4$.

We test all algorithms on two examples. The first example consists of a point cloud (200 points) sampled from an ellipse, while the second example consists of 250 points sampled from a three-petal flower. We introduce random Gaussian noise into both point clouds. The data are visualized in Figure \ref{fig.noisy} (a). For the ellipse example, we set $\Delta t = 2 \times 10^{-3}$ in the first stage and $\Delta t = 1 \times 10^{-3}$ in the second stage. For the three-petal flower example, due to the complexity of the data structure, a smaller time step is required to ensure the stability of the algorithm with a large $\eta_2$. In this case, we use $\Delta t = 1 \times 10^{-3}$ in the first stage and $\Delta t = 8 \times 10^{-4}$ in the second stage.

The results produced by the five methods are displayed in Figure \ref{fig.noisy} (c)-(g). The results in Column (c)-(g) yield oscillating reconstructed curves due to the presence of noise. In contrast, the proposed method, which incorporates the normal information term that enforces smooth variation in the normal direction of the reconstructed curve, successfully reconstructs smooth curves that reflect the underlying structures of the point clouds. Although the TVG method \cite{liang2013robust} also utilizes normal directions of point cloud, its results relies on initialized inside and outside regions precomputed from the point cloud. This region separation is sensitive to noise, especially when noise is large that makes the points spread out over a large region. Additionally, we plot the vector field $\mathbf{p}_d$ for both examples in Figure \ref{fig.noisy.field}. As observed in Figure \ref{fig.incomplete.field}, the reconstructed curves are nearly normal to the field. This observation substantiates that the normal information term serves as an effective regularizer for noisy data.

\begin{figure}[t!]
	\centering
	\begin{tabular}{ccccc}
		(a) Data & (b) $\eta_2=0$ & (c) $\eta_2=4\times 10^3$ & (d)$\eta_2=8\times 10^3$ & (e)  $\eta_2=8\times 10^4$\\
		\includegraphics[height=0.17\textwidth]{figures/PartialHexagon} &
		\includegraphics[height=0.17\textwidth]{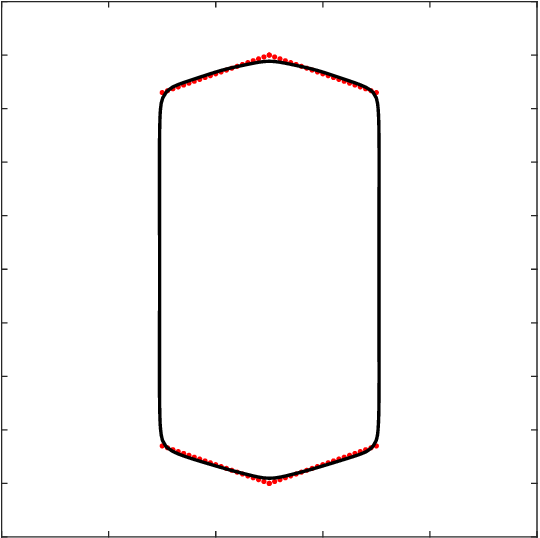} &
		\includegraphics[height=0.17\textwidth]{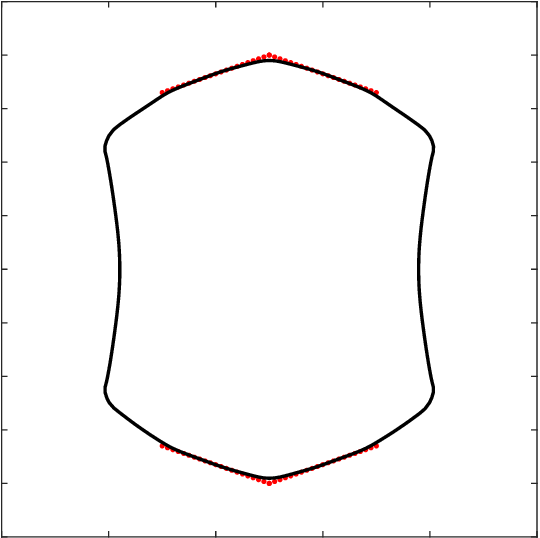} &
		\includegraphics[height=0.17\textwidth]{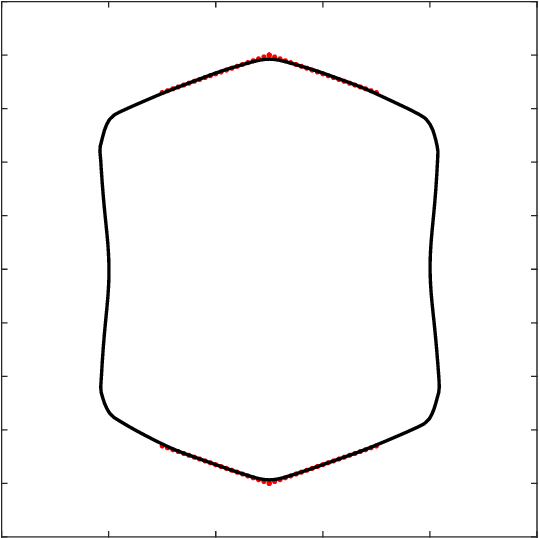}&
		\includegraphics[height=0.17\textwidth]{figures/PartialHexagon_PCA_new}\\
		
	\end{tabular}
	\caption{Impacts of $\eta_2$ on two-dimensional incomplete data. (a) shows the data set sampled from a hexagon. But only data near the two opposite corners are available. (b)-(e) show results by the proposed method with (b) $\eta_2=0$, (c) $\eta_2=4\times 10^3$, (d)$\eta_2=8\times 10^3$ and (e) $\eta_2=8\times 10^4$.}
	\label{fig.normal.eta}
\end{figure}

\begin{figure}[t!]
	\centering
	\begin{tabular}{ccccc}
		(a) Data & (b) win. size 2 & (c) win. size 6 & (d)win. size 10 & (e)  win. size 14\\
		\includegraphics[height=0.17\textwidth]{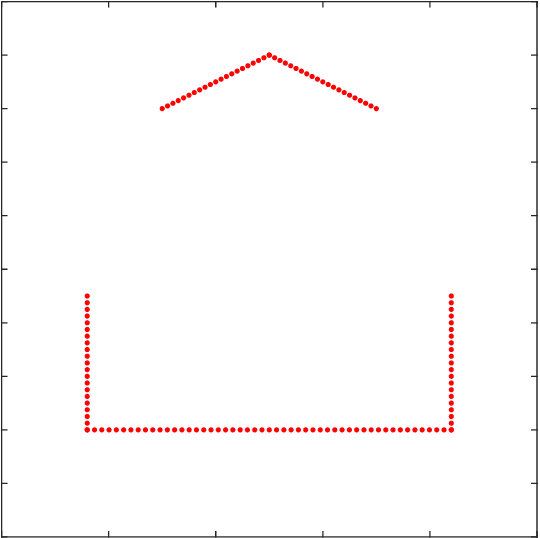} &
		\includegraphics[height=0.17\textwidth]{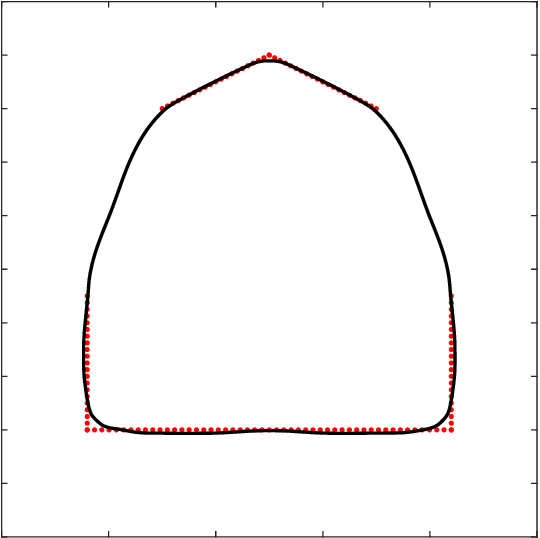} &
		\includegraphics[height=0.17\textwidth]{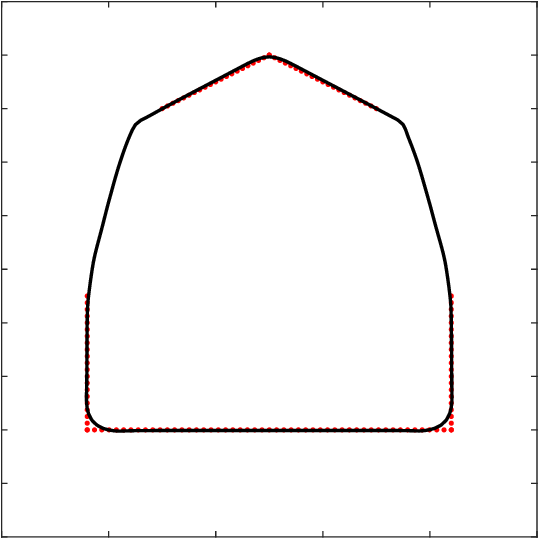} &
		\includegraphics[height=0.17\textwidth]{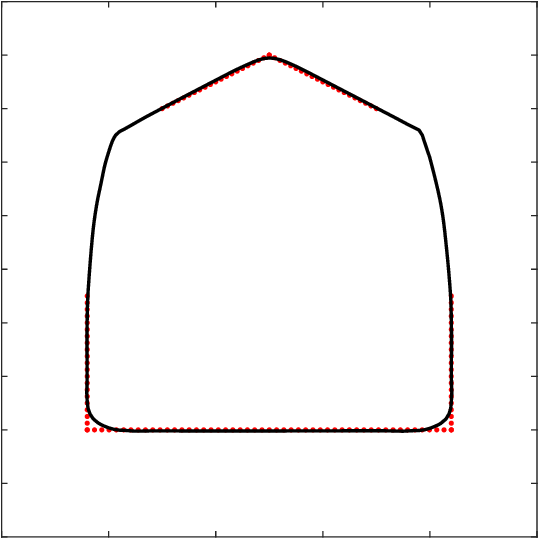}&
		\includegraphics[height=0.17\textwidth]{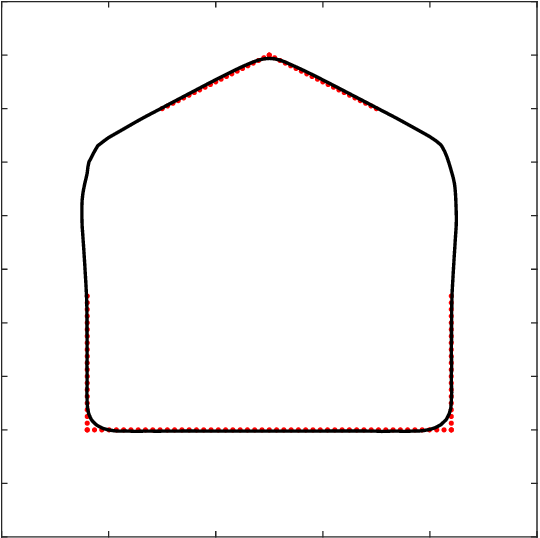}\\
		
	\end{tabular}
	\caption{Impacts of windows size in PCA on two-dimensional incomplete data. (a) shows the data set sampled from a pentagon. But data around the shoulder are missing. (b)-(e) show results by the proposed method with (b) window size 2, (c) window size 6, (d) window size 10 and (e) window size 14.}
	\label{fig.normal.win}
\end{figure}

\subsection{Impacts of the normal information term}
We examine the effects of the normal information term on the algorithm's performance by adjusting its weight $\eta_2$ and the window size used for PCA. We first consider the incomplete point cloud data from a hexagon with 84 points, as shown in Figure \ref{fig.normal.eta}(a), to assess the impacts of $\eta_2$. We anticipate that a larger $\eta_2$ will lead to a stronger regularizing effect of the normal information term. In this set of experiments, we set $r(\mathbf{x}) = \sqrt{f}(\xb)$ and fix the parameters as follows: $\eta_0 = 10$, $\eta_1 = 2 \times 10^4$, with a window size of 12 and $\Delta t = 2 \times 10^{-4}$. We apply our method with $\eta_2$ varying among $0$, $4 \times 10^3$, $8 \times 10^3$, and $8 \times 10^4$. The results are displayed in Figure \ref{fig.normal.eta} (b)-(e). 

When the normal information is not utilized, i.e., for $\eta_2 = 0$, the result merely connects the points. For $\eta_2 > 0$, the upper and lower parts of the reconstructed curve extend the trend of the point cloud. A larger $\eta_2$ allows the extended curve to align more closely with this trend.

To assess the impacts of the window size, we consider the incomplete point cloud consisting of 135 points sampled from a pentagon, with data around the shoulder missing, as illustrated in Figure \ref{fig.normal.win}(a). Intuitively, a larger window size extends the estimated normal information over a larger region, thereby propagating the pattern of the point cloud into a more extensive data-missing area. In this example, we set $r(\mathbf{x}) = \sqrt{f(\xb)}$ and fix the parameters as follows: $\eta_0 = 30$, $\eta_1 = 10^4$, $\eta_2 = 4 \times 10^4$, and $\Delta t = 2 \times 10^{-4}$. We vary the window size among $2$, $6$, $10$, and $14$. The results are presented in Figure \ref{fig.normal.win} (b)-(e). As the window size increases, the reconstructed curve exhibits a wider shoulder, which aligns with our expectations.

\subsection{Three-dimensional experiments}
We examine performances of the proposed model on three-dimensional problems. In all experiments, $\Delta x_1=\Delta x_2=\Delta x_3=1$ are used. We set $\alpha_1 = \alpha_2 = 500$ and $\gamma_1 = \gamma_2 = 10$. For reinitialization, we perform three iterations. 

\begin{figure}[t!]
	\centering
	\begin{tabular}{ccc}
		\includegraphics[trim={3cm 2cm 3cm 2cm},clip,width=0.3\textwidth]{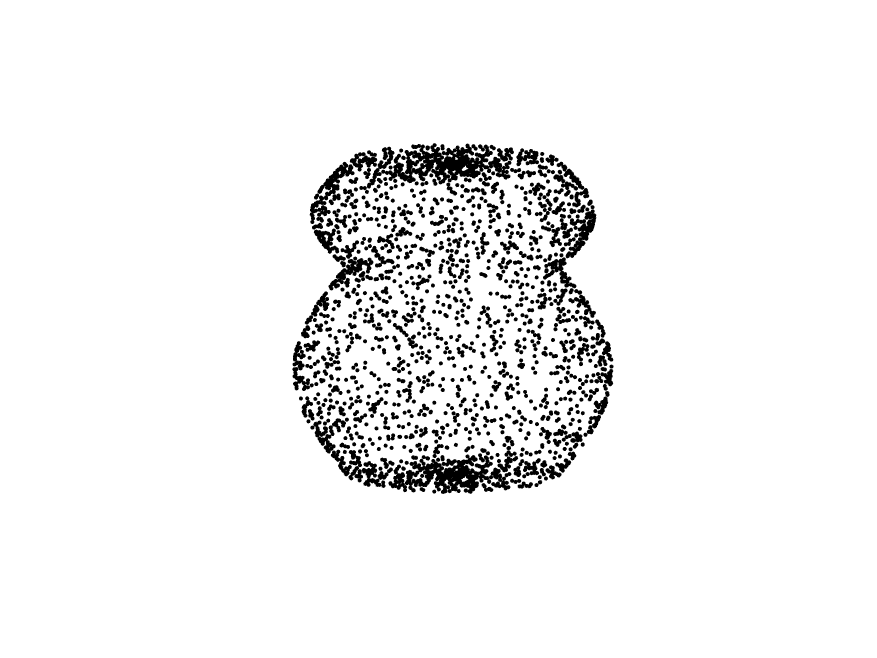} &
		\includegraphics[trim={4cm 3cm 4cm 3cm},clip,width=0.3\textwidth]{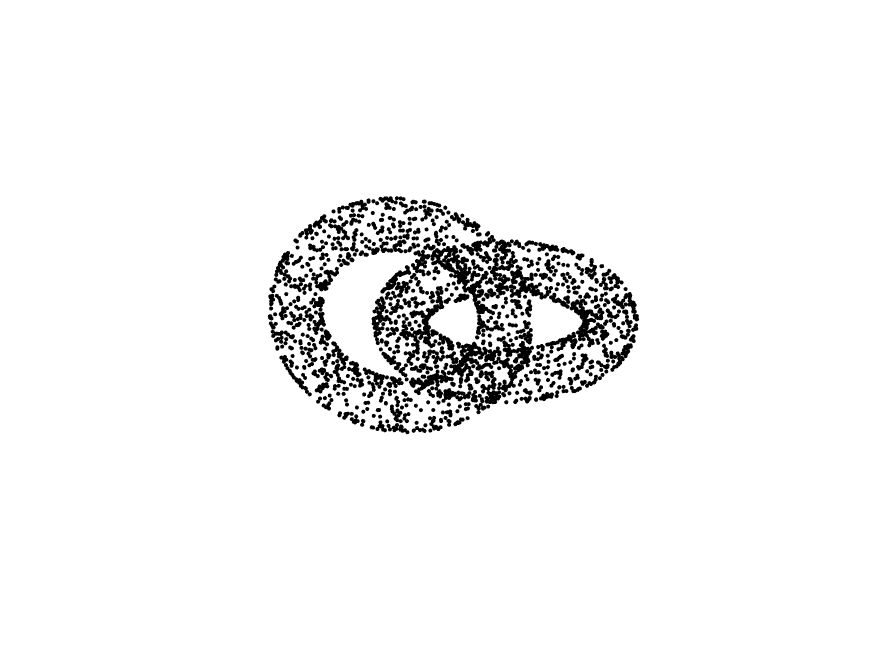} &
		\includegraphics[trim={3.5cm 2cm 3.5cm 1.5cm},clip,width=0.3\textwidth]{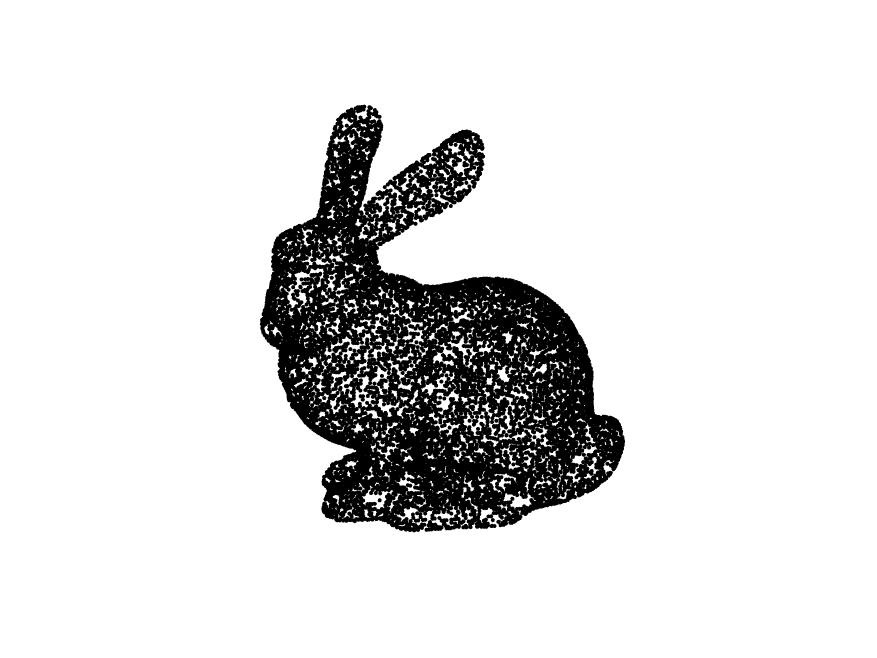} \\
		\includegraphics[trim={3cm 2cm 3cm 2cm},clip,width=0.3\textwidth]{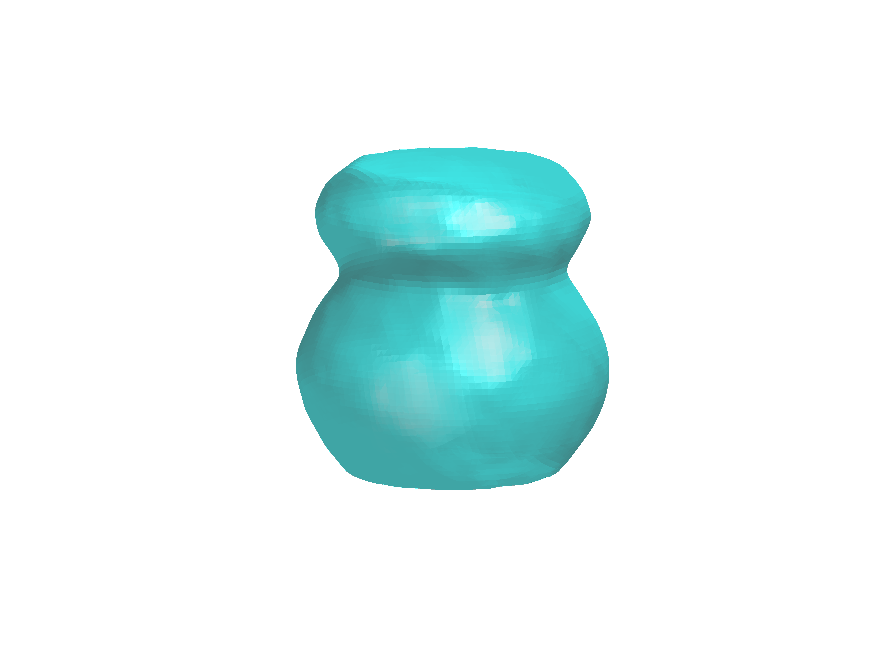} &
		\includegraphics[trim={4cm 3cm 4cm 3cm},clip,width=0.3\textwidth]{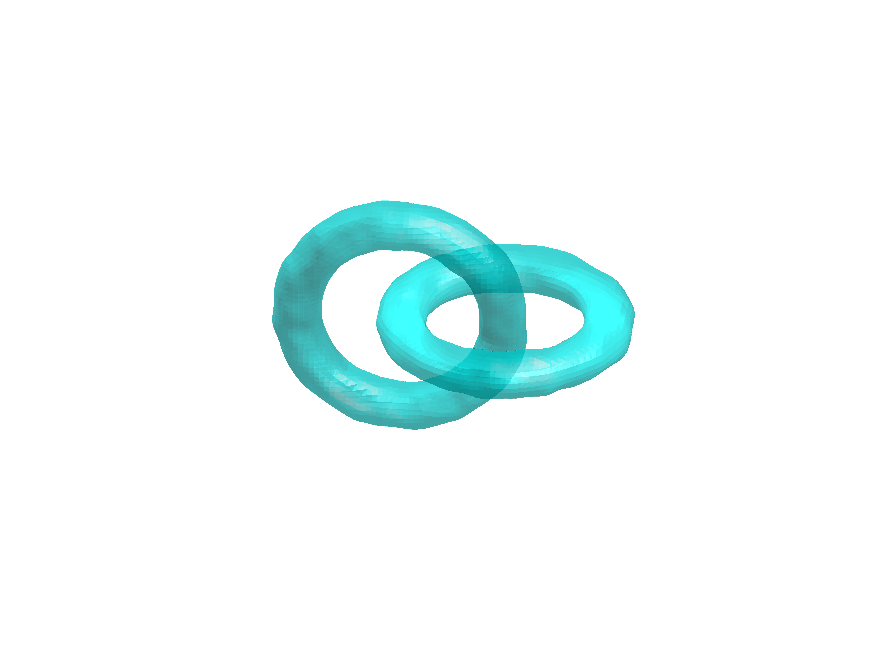} &
		\includegraphics[trim={3.5cm 2cm 3.5cm 1.5cm},clip,width=0.3\textwidth]{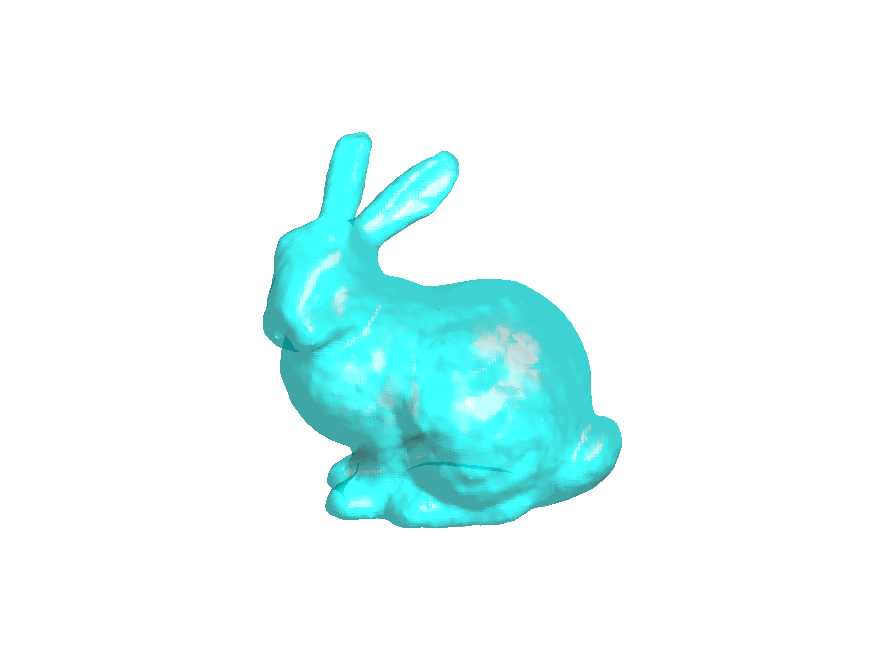} 
	\end{tabular}
	\caption{Three-dimensional examples. The first row shows the point cloud data. The second row shows the reconstructed surface.}
	\label{fig.3d.clean}
\end{figure}

\subsubsection{General performance}
We first apply our method to complete point clouds, considering three examples: the pot, the double tori, and the Stanford bunny. We use 3500 points, 3000 points and 15000 points in the point cloud, and computational domain $[0,70]^3$, $[0,90]\times [0,70]\times [0,70]$ and $[0,150]^3$ with $\Delta x=\Delta y=\Delta z=1$  for the pot, double tori and Stanford bunny, respectively. The point cloud data are visualized in the first row of Figure \ref{fig.3d.clean}. In this set of experiments, since the data is clean and complete, the normal information term does not require a strong influence. We use $r(\mathbf{x}) = 1$, a window size of 8 for PCA, and a time step of $\Delta t = 2$. We set $\eta_0 = \eta_1 = 0.1$, $\eta_2 = 0.2$ and $\varepsilon = 1$ for the pot and double tori. For the Stanford bunny, due to its more complicated structures, we use $\eta_0 = 0.1$, $\eta_1 = \eta_2 = 0.05$ and $\varepsilon = 0.01$, . The results are presented in the second row of Figure \ref{fig.3d.clean}. Our method effectively reconstructs the underlying surfaces with various details. It is noteworthy that our initial condition is a hypercube that encloses the entire point cloud. For the double tori example, the reconstructed surface automatically adapts to the data geometry, thanks to the level set formulation.

\begin{figure}[t!]
	\centering
	\begin{tabular}{ccc}
		(a) Data & (b) Our result & (c) DS\\
		\includegraphics[trim={1cm 0.5cm 1cm 0.5cm},clip,width=0.3\textwidth]{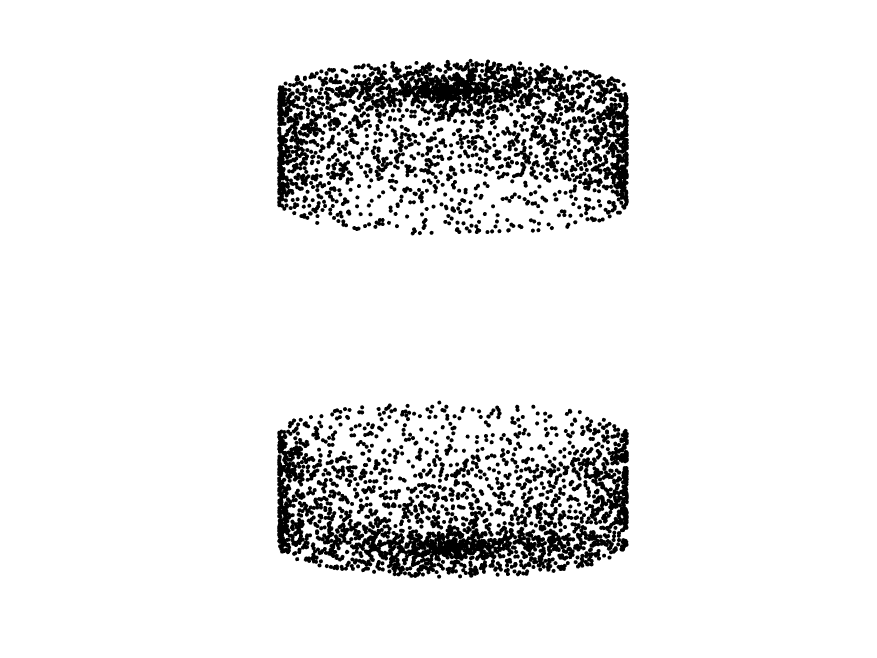} &
		\includegraphics[trim={1cm 0.5cm 1cm 0.5cm},clip,width=0.3\textwidth]{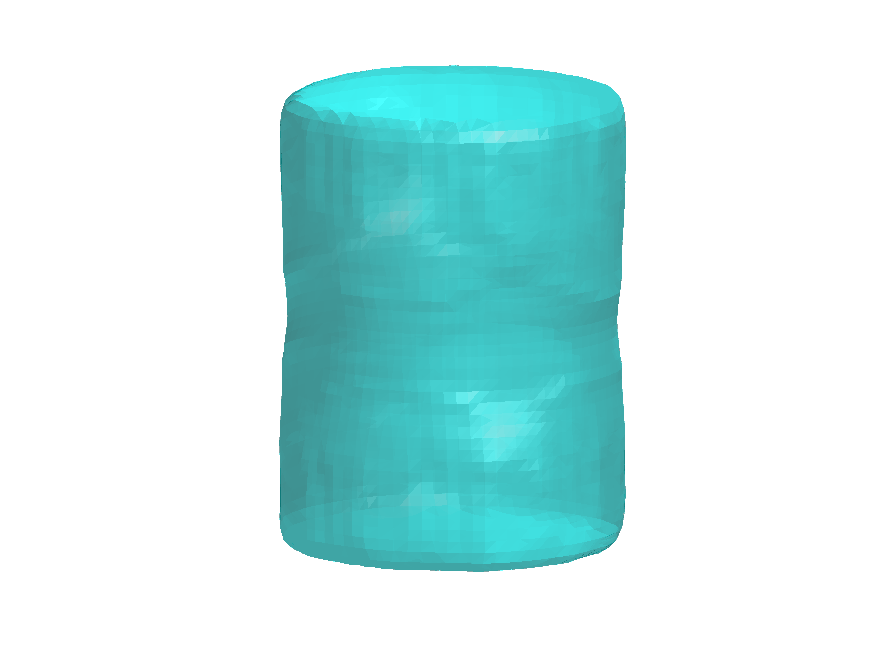} &
		\includegraphics[trim={1cm 0.5cm 1cm 0.5cm},clip,width=0.3\textwidth]{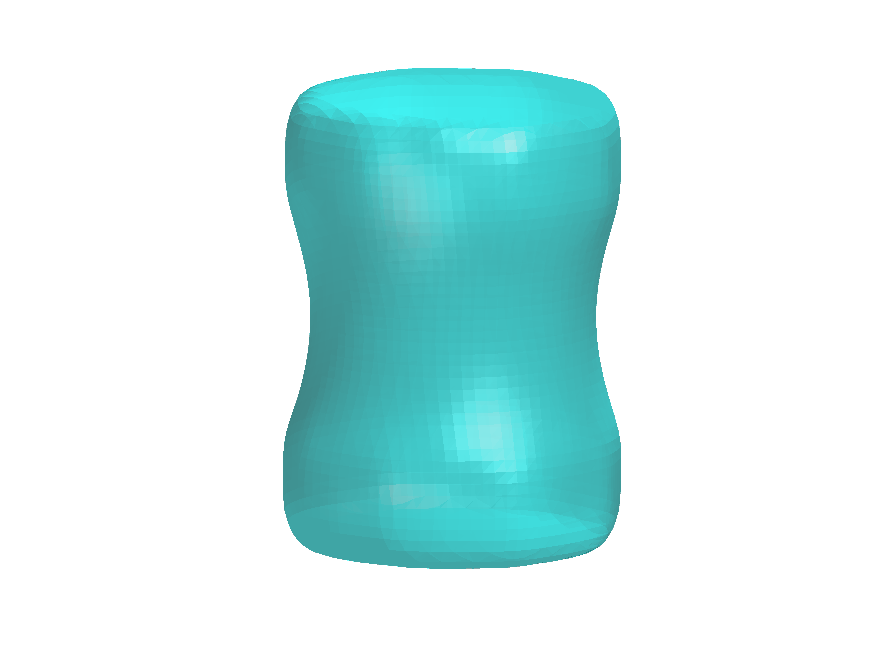} \\
		(d) CR & (e) DSP & (f) TVG\\
		\includegraphics[trim={1cm 0.5cm 1cm 0.5cm},clip,width=0.3\textwidth]{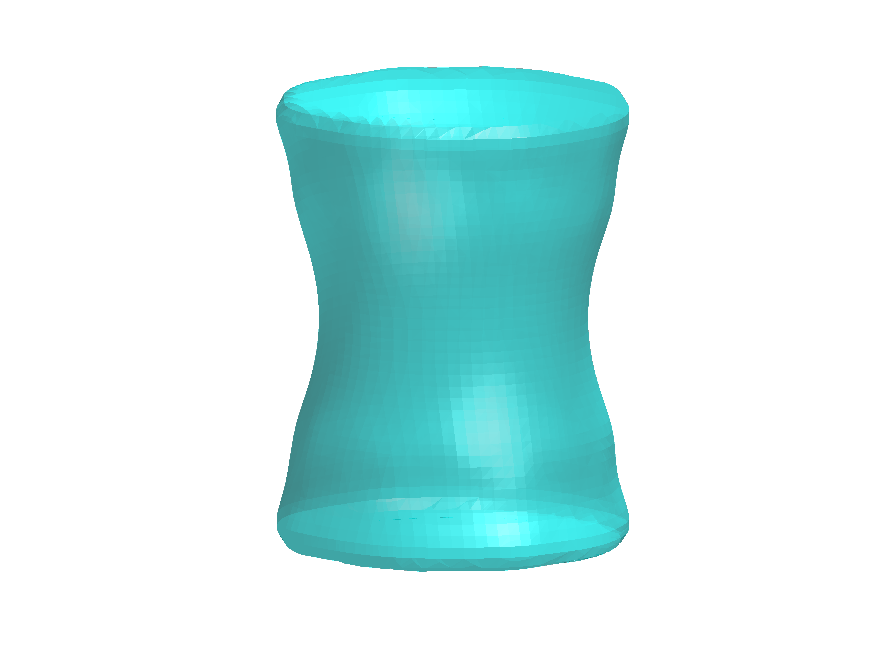} &
		\includegraphics[trim={1cm 0.5cm 1cm 0.5cm},clip,width=0.3\textwidth]{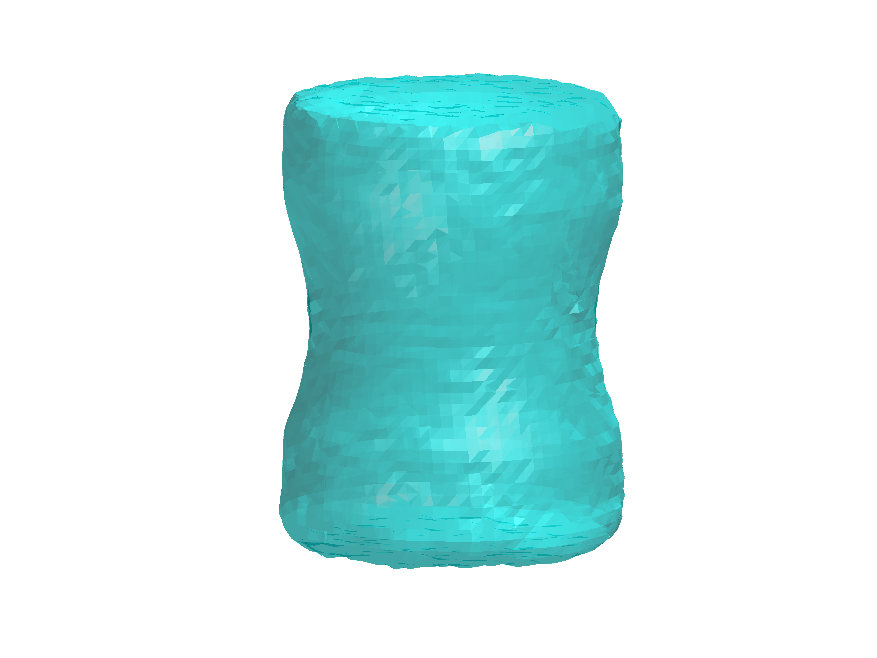} &
		\includegraphics[trim={1cm 0.5cm 1cm 0.5cm},clip,width=0.3\textwidth]{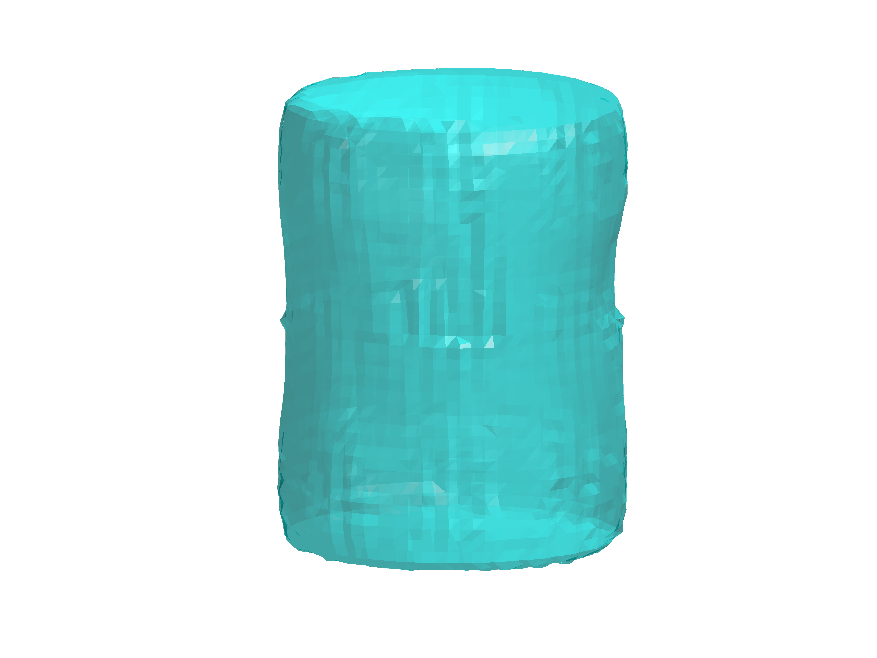} 
	\end{tabular}
	\caption{Comparison on three-dimensional incomplete data. (a) shows the data set sampled from a cylinder. Data in the middle region are missing. (b)-(e) show results by the proposed method, DS, CR, DSP and TVG, respectively. }
	\label{fig.3d.incomplete.cylinder}
\end{figure}

\begin{figure}[t!]
	\centering
	\begin{tabular}{ccc}
		(a) Data & (b) Our result & DS\\
		\includegraphics[trim={1.5cm 2cm 2cm 2cm},clip,width=0.3\textwidth]{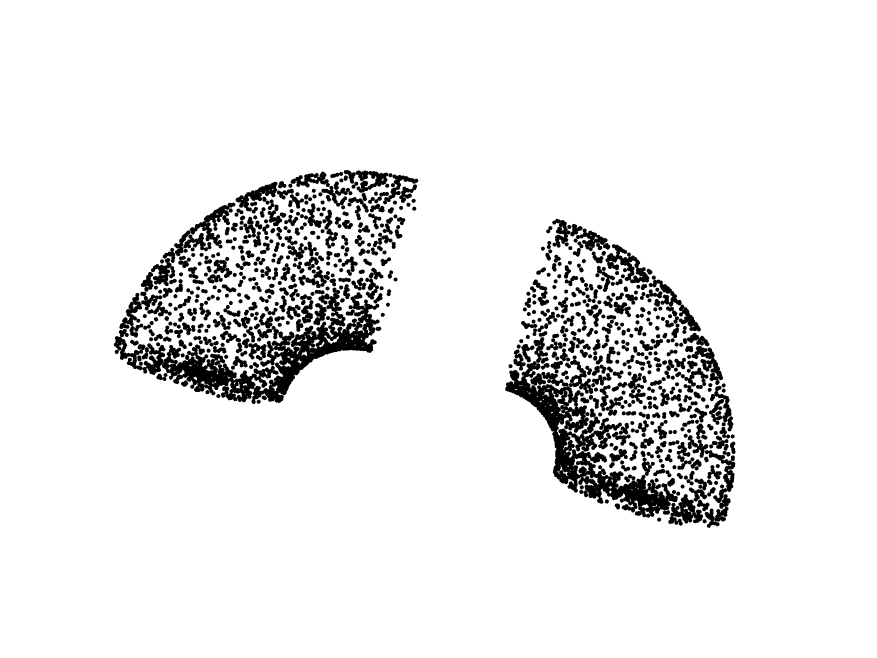} &
		\includegraphics[trim={1.5cm 2cm 2cm 2cm},clip,width=0.3\textwidth]{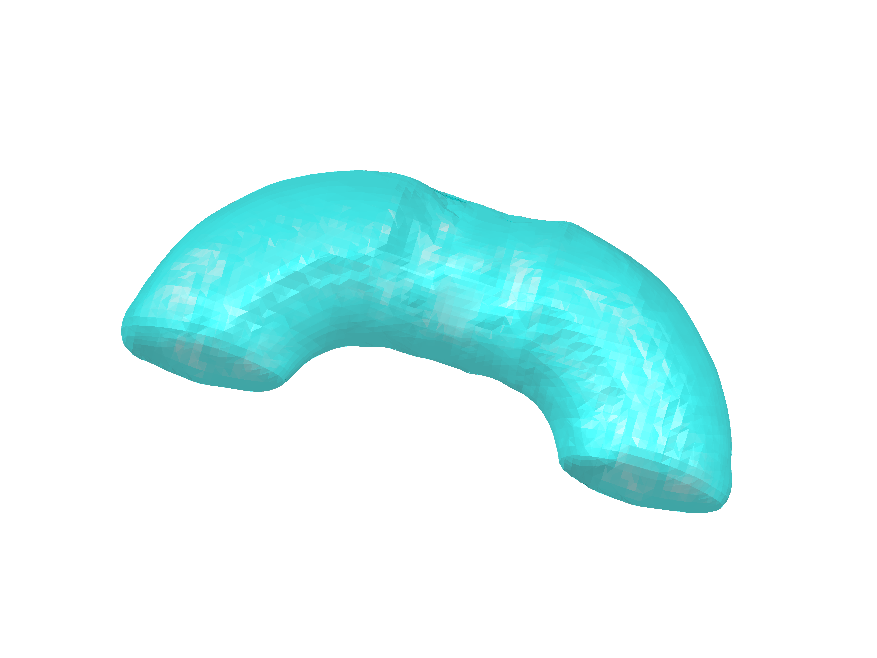} &
		\includegraphics[trim={1.5cm 2cm 2cm 2cm},clip,width=0.3\textwidth]{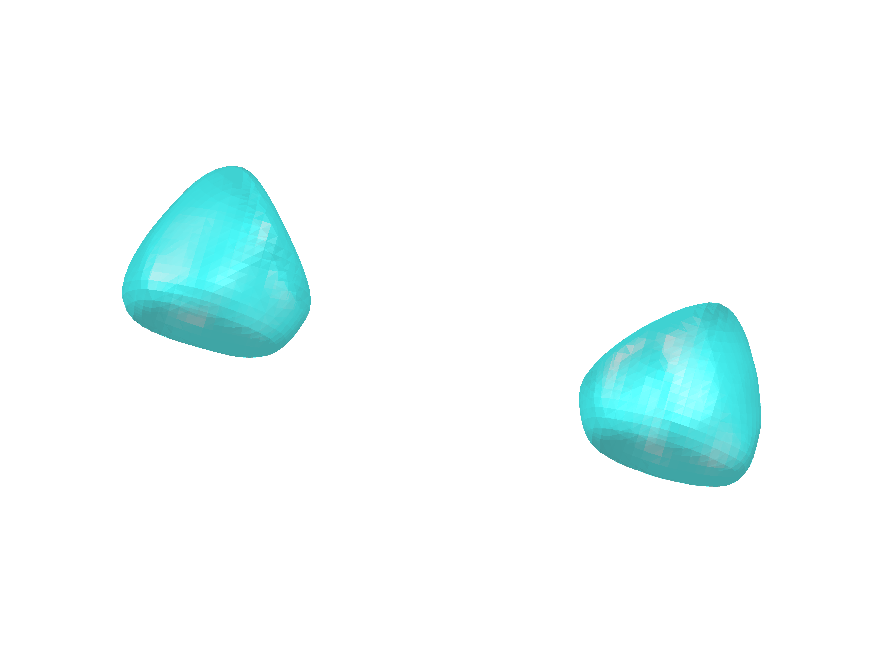}\\
		(d) CR & (e) DSP & (f) TVG\\
		\includegraphics[trim={1.5cm 2cm 2cm 2cm},clip,width=0.3\textwidth]{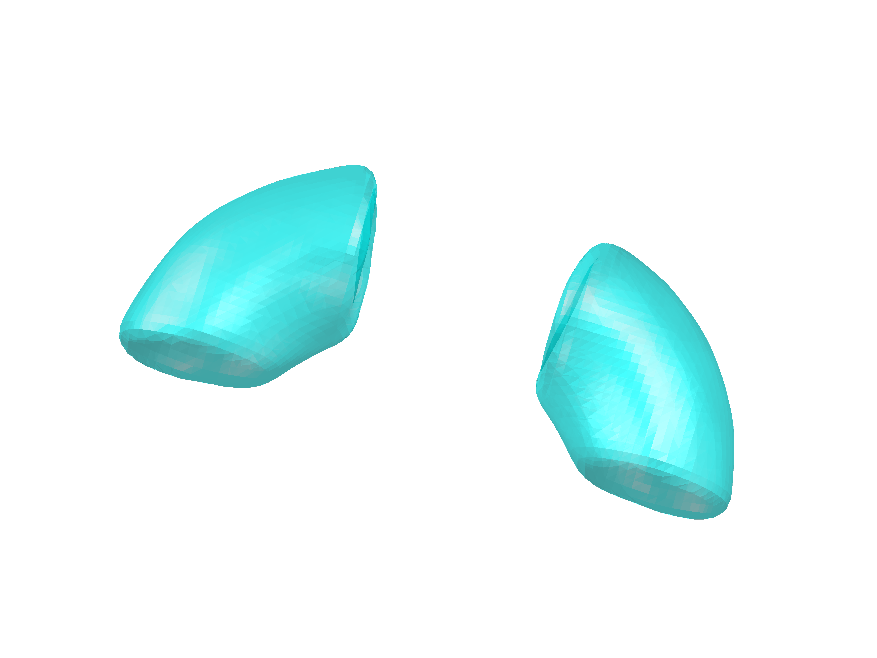} &
		\includegraphics[trim={1.5cm 2cm 2cm 2cm},clip,width=0.3\textwidth]{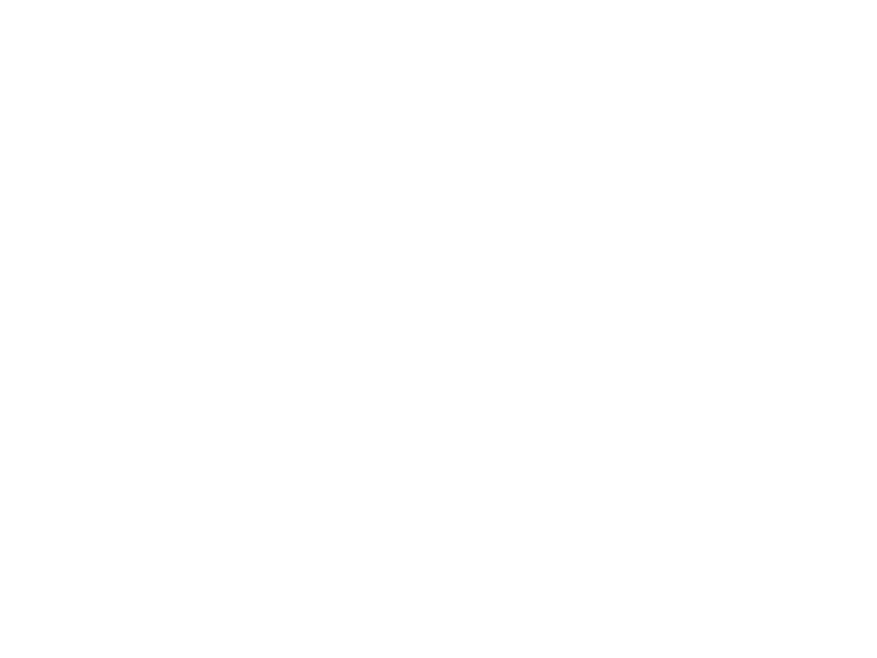} &
		\includegraphics[trim={1.5cm 2cm 2cm 2cm},clip,width=0.3\textwidth]{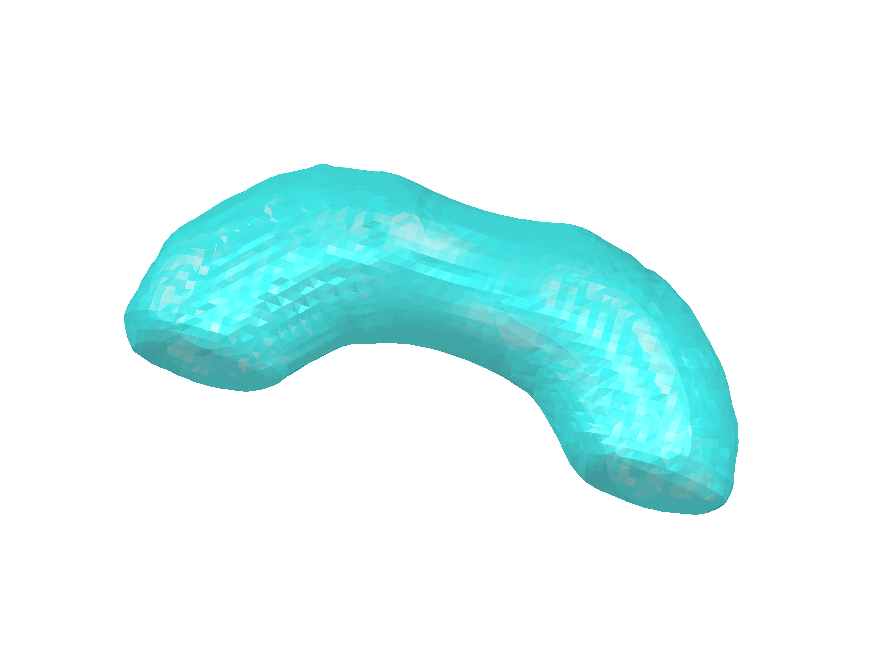} 
	\end{tabular}
	\caption{Comparison on three-dimensional incomplete data. (a) shows the data set sampled from a hand rail. Data in the middle region are missing. (b)-(e) show results by the proposed method, DS, CR, DSP and TVG, respectively. The method from DSP failed to reconstruct a surface. }
	\label{fig.3d.incomplete.hand}
\end{figure}

\begin{figure}[t!]
	\centering
	\begin{tabular}{ccc}
		(a) Data & (b) Our result & (c) DS\\
		\includegraphics[trim={0.3cm 1.5cm 1.2cm 1.5cm},clip,width=0.3\textwidth]{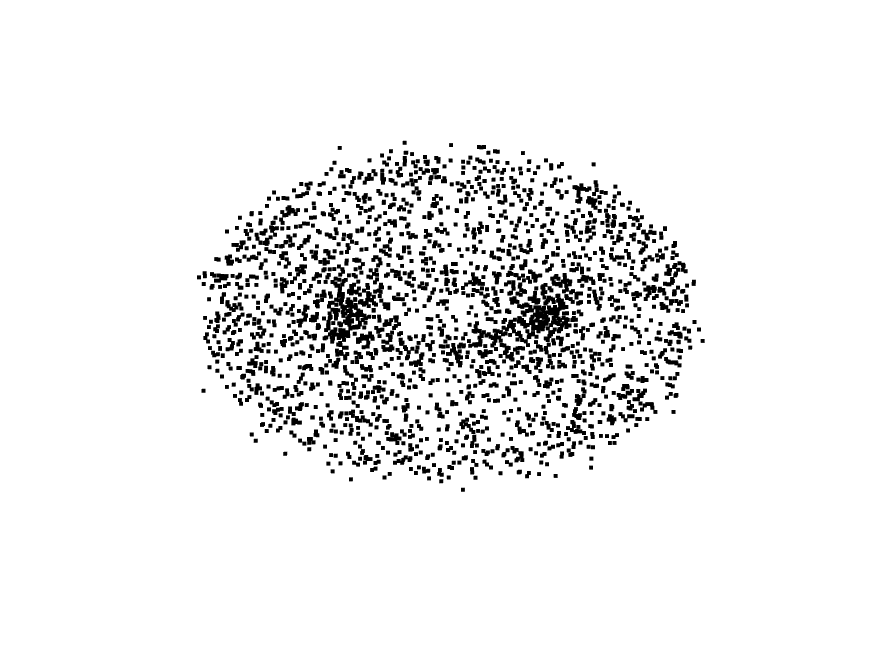} &
		\includegraphics[trim={0.3cm 1.5cm 1.2cm 1.5cm},clip,width=0.3\textwidth]{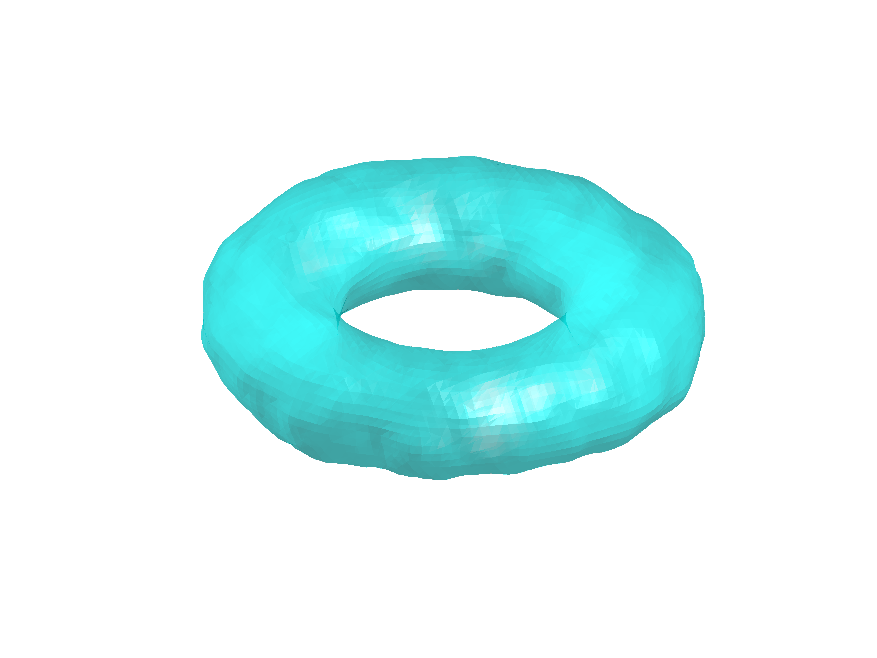} &
		\includegraphics[trim={0.3cm 1.5cm 1.2cm 1.5cm},clip,width=0.3\textwidth]{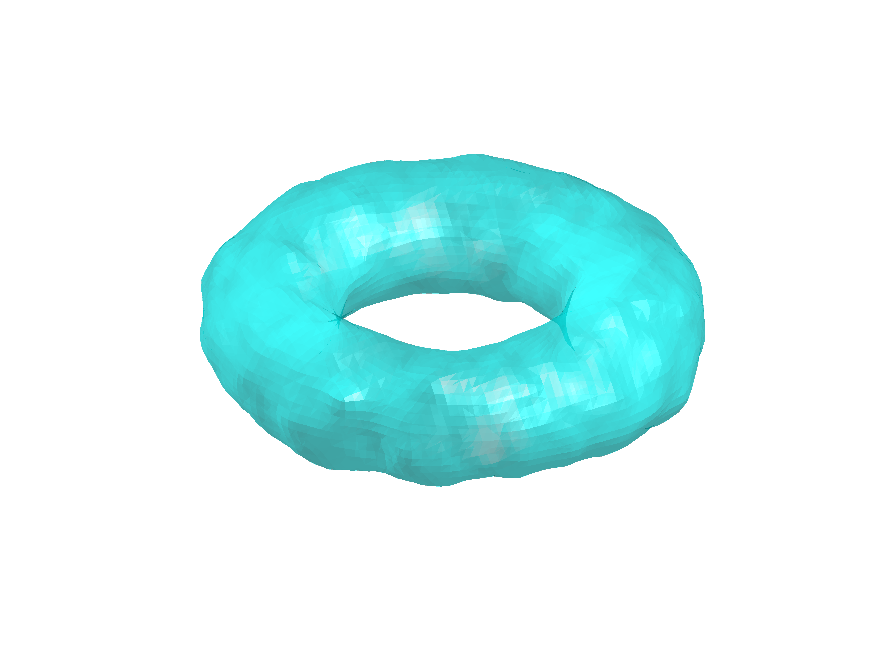}\\
		(d) CR & (e) DSP & (f) TVG\\
		\includegraphics[trim={0.3cm 1.5cm 1.2cm 1.5cm},clip,width=0.3\textwidth]{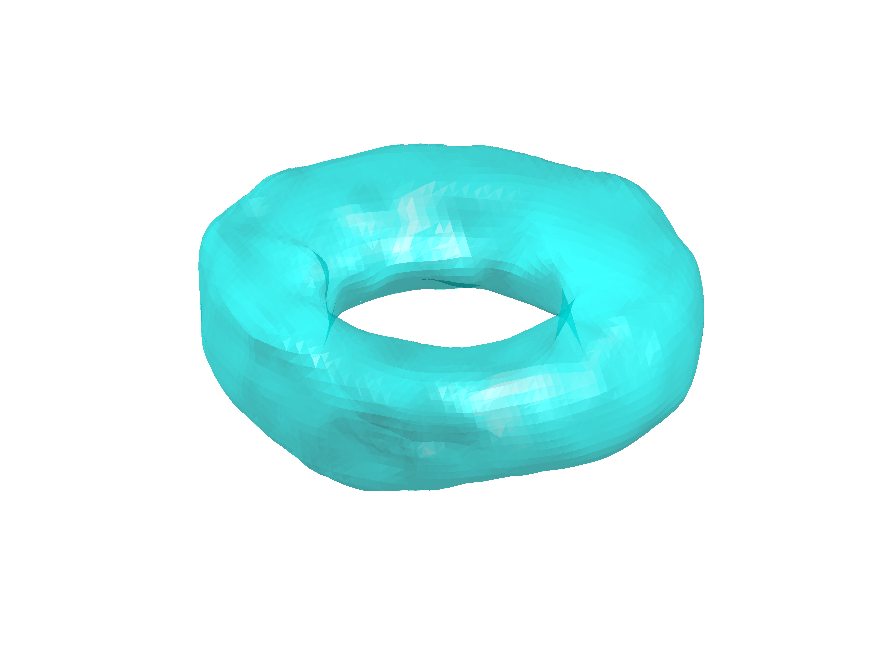} &
		\includegraphics[trim={0.3cm 1.5cm 1.2cm 1.5cm},clip,width=0.3\textwidth]{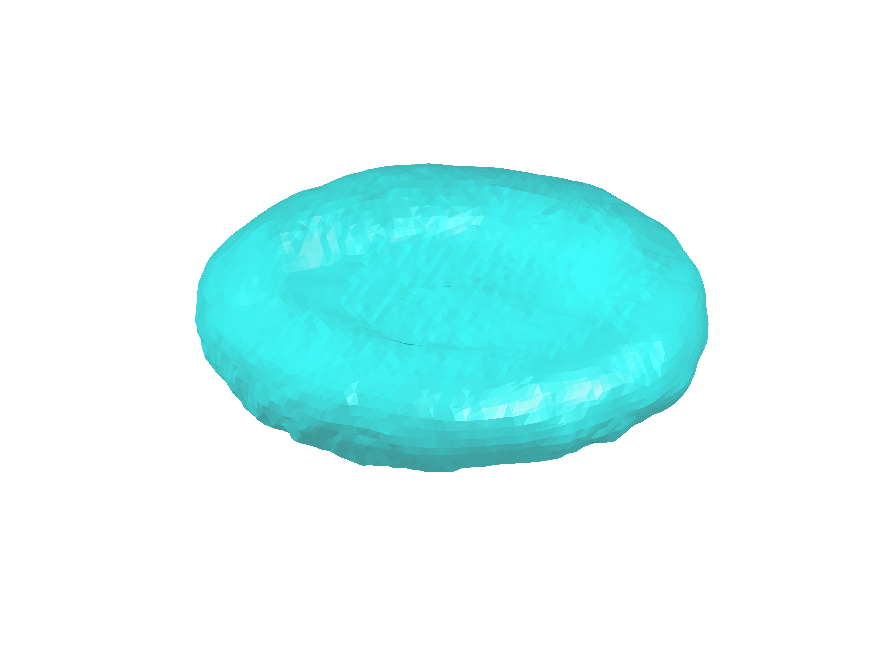} &
		\includegraphics[trim={1.5cm 2cm 2cm 2cm},clip,width=0.3\textwidth]{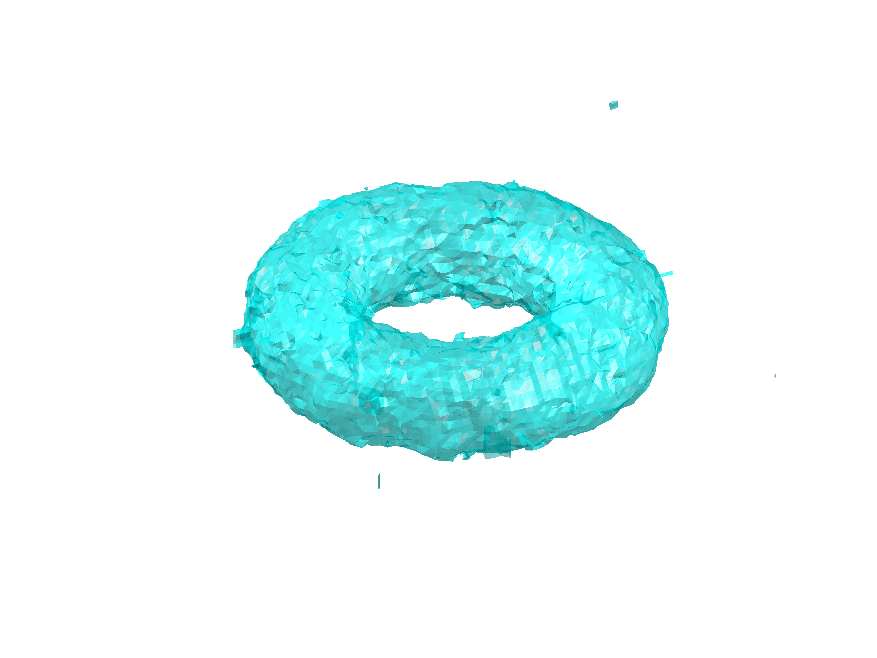} 
	\end{tabular}
	\caption{Comparison on three-dimensional noisy data. (a) shows the data set sampled from a hand rail. Data in the middle region are missing. (b)-(e) show results by the proposed method, DS, CR, DSP and TVG, respectively. }
	\label{fig.3d.noisy.torus}
\end{figure}

\begin{table}
	\centering
	\begin{tabular}{c|c|c|c|c|c}
		\hline
		& Ours & DS & CR & DSP & TVG\\
		\hline
		Incomplete Cylinder (Figure \ref{fig.3d.incomplete.cylinder})& 480.99s & 200.50s & 356.58s & 658.93s & 80.05s\\
		\hline
		Noisy Torus (Figure \ref{fig.3d.noisy.torus})& 128.50s & 58.43s & 74.77s & 243.91s & 32.42s\\
		\hline
	\end{tabular}
	\caption{Comparison on three-dimensional data. CPU time used to compute results in Figure \ref{fig.3d.incomplete.cylinder} and \ref{fig.3d.noisy.torus}.}
	\label{tab.3d.cpu}
\end{table}
\subsubsection{Incomplete and noisy data}
We then apply our method to incomplete and noisy data and compare it with the methods from \cite{zhao2000implicit,he2020curvature,estellers2012efficient}. 

In this set of experiments for incomplete data (the first two examples), we require the normal information term to guide the reconstruction of the surface over the data-missing regions. We use $r(\mathbf{x}) = \sqrt{f(\mathbf{x})}$, $\eta_0 = 0.01$, $\eta_1 = 0$, and $\Delta t = 5$. The first example uses point cloud data sampled from a cylinder, with samples from the middle part missing. The point cloud contains 6000 points. We use computation domain $[0,50]^3$. The data is visualized in Figure \ref{fig.3d.incomplete.cylinder}(a). In this example, we set $\eta_2 = 1$ and a window size of 12 for PCA. The results by different  methods are shown in Figure \ref{fig.3d.incomplete.cylinder}(b)-(f). In this comparison, DS, CR and DSP can reconstruct the surface, but the results exhibit hyperbolic characteristics: the waist has a noticeably smaller radius than the top and bottom sections. In contrast, our proposed method and TVG utilize the normal direction information and effectively reconstruct the cylinder, maintaining an almost uniform radius from the top to the bottom.

Our second example uses data sampled from a handrail, with samples from the middle part of the rail missing. The point cloud contains 6000 points. We use computation domain $[0,90]\times [0,40]\times [0,30]$. The data is visualized in Figure \ref{fig.3d.incomplete.hand}(a). In this case, since the underlying surface is thin and there is a significant data-missing region, we employ a stronger weight of $\eta_2 = 3$ for the normal information term, along with a slightly smaller window size of 10 for PCA. The results from the four methods are displayed in Figure \ref{fig.3d.incomplete.hand}(b)-(e). Due to the incompleteness of the point cloud, the methods from \cite{zhao2000implicit,he2020curvature,estellers2012efficient} fail to reconstruct the handrail. In particular, the methods from \cite{zhao2000implicit,he2020curvature} yield two separate surfaces, failing to connect the two parts of the point cloud. Furthermore, the method from \cite{estellers2012efficient} fails to reconstruct a surface; during the iterations, the reconstructed surface consistently shrinks and ultimately disappears. In contrast, our proposed method successfully reconstructs the handrail as a connected single surface, effectively capturing the structure of the underlying surface of the point cloud.

Our third example considers a noisy point cloud, which is sampled from a torus and purturbed by Gaussian noise. The point cloud contains 3000 points. We use computation domain $[0,65]\times [0,65]\times [0,30]$. The data is visualized in Figure \ref{fig.3d.noisy.torus}(a). In the proposed method, we use $\eta_0=\eta_1=0.1$. Since we need the normal information term to regularize the reconstructed surface, we use $\eta_2=1,r(\xb)=1$, together with a window size of 8. The results by all of the five methods are shown in Figure \ref{fig.3d.noisy.torus}(b)-(f). In this example, with the same initial condition, DSP fails to resolve the topology of the torus. TVG successively reconstructs a torus. But the surface is very noisy and contains outliers, which is similar to our observations for two-dimensional noisy point cloud in Section \ref{sec.2d.noisy}. All of the proposed method, DS and CR reconstruct smooth torus, as shown in (b)-(d). While the result by the proposed method is smoother and more uniform along the axis of revolution.

We then compare in Table \ref{tab.3d.cpu} the CPU time used by all methods to achieve results in Figure \ref{fig.3d.incomplete.cylinder} and \ref{fig.3d.noisy.torus}. For the incomplete cylinder example, it requires 1000 iterations for the proposed method, DS, CR and DSP, and 200 iterations for TVG to converge. For the noisy torus example, we use 500 iterations for the proposed method, DS, CR and DSP, and 200 iterations for TVG. Similar to our observation in Table \ref{tab.incomplete.cpu}, TVG is the fastest method. The propose method is slower than other methods due to the increased model complexity. For three dimensional examples, DSP is much slower than its performance on two dimensional examples. This is because it estimates the surface normal during iterations. And this estimation scales with the point cloud size.

\section{Conclusion}
\label{sec.conclusion}

We propose a PCA based model for surface reconstruction from incomplete point cloud data. Our model employs the level set method to represent the surface and consists of three components: a fidelity term that utilizes the distance function to the point cloud, and two regularization terms that incorporate surface curvature and estimated surface normal information. The normal information is estimated using PCA and is employed to guide the reconstructed surface in data-missing regions. An operator-splitting approach is proposed to solve the model. 
Compared to existing models, the proposed model demonstrates improved performance in reconstructing surfaces from incomplete point cloud data. For incomplete data, the reconstructed surface transitions smoothly from the data-available regions to the data-missing regions while maintaining surface connectivity. Additionally, an ablation study is conducted to highlight the effects of the normal information term.

\appendix
\section*{Appendix}
\section{Derivation of (\ref{eq.opti.fourth-order})}
\label{sec.variation.curvature}
Problem (\ref{eq.dis2.1.penal.1}) is equivalent to
\begin{align}
	\begin{cases}
		(\psi^{n+2/4},\ub^{n+2/4})=\argmin_{\phi,\vb}  \bigg[\displaystyle\frac{1}{2} \int_{\Omega} |\phi-\psi^{n+1/4}|^2d\xb + \frac{\gamma_1}{2} \displaystyle\int_{\Omega} |\vb-\ub^{n+1/4}|^2d\xb \\
		\hspace{0.5cm} +\displaystyle\frac{\gamma_2}{2} \int_{\Omega} |p-q^{n+1/4}|^2d\xb 
		+ \frac{\Delta t \alpha_1}{2}\displaystyle\int_{\Omega} \left| \vb-\frac{\nabla \phi}{|\nabla \phi|}\right|^2d\xb + \frac{\Delta t\alpha_2}{2} \int_{\Omega} \left|\nabla\cdot \vb-\nabla \cdot \frac{\nabla \phi}{|\nabla \phi|}\right|^2d\xb \bigg],\\
		q^{n+2/4}=\nabla\cdot \ub^{n+2/4}.
	\end{cases}
	\label{eq.appen.dis2.1.penal.1}
\end{align}
In (\ref{eq.opti.fourth-order}), the derivation of equation for $\ub^{n+2/4}$ is standard. Here we state detailed derivation for the equation of $\psi^{n+2/4}$.

Note that the equation for $\psi^{n+2/4}$ can be written as
\begin{align}
	\psi^{n+2/4}+ \partial_{\phi} J_4(\psi^{n+2/4},\ub^{n+2/4}) + \partial_{\phi} J_5(\psi^{n+2/4},\nabla\cdot \ub^{n+2/4})\ni 0.
	\label{eq.appen.phi}
\end{align}
It remains to derive the expression for $\partial_{\phi} J_4(\phi,\vb)$ and $\partial_{\phi} J_5(\phi,p)$. 

Denote $\nbb_{\Omega}$ as the outward normal direction of $\Omega$. For $\partial_{\phi} J_4(\phi,\vb)$, suppose $\nabla \phi \cdot \nbb_{\Omega}=\vb\cdot\nbb_{\Omega}=0$ on $\partial \Omega$. For any function $h\in H^1(\Omega)$, the variation of $J_4(\phi,\vb)$ with respect to $\phi$ in direction $h$ is
\begin{align}
	\delta_{\phi}[J_4(\psi,\vb),h]=&\lim_{\varepsilon\rightarrow 0}\frac{\alpha_1}{2\varepsilon}\int_{\Omega} \left| \vb-\frac{\nabla (\phi+\varepsilon h)}{|\nabla (\phi+\varepsilon h)|}\right|^2- \left| \vb-\frac{\nabla \phi}{|\nabla \phi|}\right|^2d\xb \nonumber\\
	=& \lim_{\varepsilon\rightarrow 0}\frac{\alpha_1}{2\varepsilon}\int_{\Omega}  2\vb\cdot \left(\frac{\nabla \phi}{|\nabla \phi|}-\frac{\nabla (\phi+\varepsilon h)}{|\nabla (\phi+\varepsilon h)|} \right)d\xb  \nonumber\\
	=& \lim_{\varepsilon\rightarrow 0}\frac{\alpha_1}{\varepsilon}\int_{\Omega}  \vb\cdot \left(\frac{\nabla \phi}{|\nabla \phi|}-\frac{\nabla (\phi+\varepsilon h)}{|\nabla \phi|+ \varepsilon \frac{\nabla\phi\cdot \nabla h}{|\nabla \phi|}+O(\varepsilon^2) } \right)d\xb \nonumber\\
	=& \lim_{\varepsilon\rightarrow 0}\frac{\alpha_1}{\varepsilon}\int_{\Omega}  \vb\cdot \frac{\varepsilon \frac{\nabla\phi\cdot \nabla h}{|\nabla \phi|}\nabla \phi- \varepsilon |\nabla \phi|\nabla h +O(\varepsilon^2)}{|\nabla \phi|(|\nabla \phi|+\varepsilon \frac{\nabla\phi\cdot \nabla h}{|\nabla \phi|}+ O(\varepsilon^2))} d\xb \nonumber\\
	=& \lim_{\varepsilon\rightarrow 0}\alpha_1\int_{\Omega}  \vb\cdot \frac{ \frac{\nabla\phi\cdot \nabla h}{|\nabla \phi|}\nabla \phi-  |\nabla \phi|\nabla h +O(\varepsilon)}{|\nabla \phi|(|\nabla \phi|+\varepsilon \frac{\nabla\phi\cdot \nabla h}{|\nabla \phi|}+ O(\varepsilon^2))} d\xb \nonumber\\
	=& \alpha_1\int_{\Omega} \frac{\nabla\phi\cdot \nabla h}{|\nabla \phi|^3}\nabla \phi \cdot \vb -\frac{1}{|\nabla \phi|}\nabla h \cdot \vb d\xb \nonumber\\
	=&-\alpha_1 \int_{\Omega} \nabla\cdot \left(\frac{\nabla \phi \cdot \vb}{|\nabla \phi|^3}\nabla\phi-\frac{\vb}{|\nabla \phi|}\right) h d\xb.
\end{align}
It implies that
\begin{align}
	\partial_{\phi} J_4(\phi,\vb)=-\alpha_1\nabla\cdot \left(\frac{\nabla \phi \cdot \vb}{|\nabla \phi|^3}\nabla\phi-\frac{\vb}{|\nabla \phi|}\right).
	\label{eq.J4}
\end{align}

For $\partial_{\phi} J_5(\phi,p)$, suppose $\nabla \phi\cdot \nbb_{\Omega}=0, p-\nabla\cdot\frac{\nabla \phi}{|\nabla \phi|}=0$ and $\nabla \left(p-\nabla\cdot\frac{\nabla \phi}{|\nabla \phi|}\right)\cdot \nbb_{\Omega}=0$ on $\partial\Omega$. For any $h\in \{h\in H^2(\Omega): \nabla h \cdot \nbb_{\Omega}=0\}$, the variation of $J_5(\phi,p)$ with respect to $\phi$ in direction $h$ is
\begin{align}
	&\delta_{\phi}[J_5(\psi,p),h] \nonumber\\
	= &\lim_{\varepsilon\rightarrow 0}\frac{\alpha_2}{2\varepsilon}\int_{\Omega} \left| p-\nabla\cdot\frac{\nabla (\phi+\varepsilon h)}{|\nabla (\phi+\varepsilon h)|}\right|^2- \left| p- \nabla\cdot\frac{\nabla \phi}{|\nabla \phi|}\right|^2d\xb \nonumber\\
	=& \lim_{\varepsilon\rightarrow 0}\frac{\alpha_2}{2\varepsilon}\int_{\Omega}  2p\nabla\cdot \left(\frac{\nabla \phi}{|\nabla \phi|}-\frac{\nabla (\phi+\varepsilon h)}{|\nabla (\phi+\varepsilon h)|} \right) + \left(\nabla\cdot\frac{\nabla (\phi+\varepsilon h)}{|\nabla (\phi+\varepsilon h)|}\right)^2 - \left( \nabla\cdot\frac{\nabla \phi}{|\nabla \phi|} \right)^2d\xb  \nonumber\\
	=& \lim_{\varepsilon\rightarrow 0}\frac{\alpha_2}{2\varepsilon}\int_{\Omega}  2p\nabla\cdot \left(\frac{\nabla \phi}{|\nabla \phi|}-\frac{\nabla (\phi+\varepsilon h)}{|\nabla (\phi+\varepsilon h)|} \right) d\xb \nonumber\\
	&\hspace{3cm}+ \left(\nabla\cdot\frac{\nabla (\phi+\varepsilon h)}{|\nabla (\phi+\varepsilon h)|}+\nabla\cdot\frac{\nabla \phi}{|\nabla \phi|}\right) \left(\nabla\cdot\frac{\nabla (\phi+\varepsilon h)}{|\nabla (\phi+\varepsilon h)|}- \nabla\cdot\frac{\nabla \phi}{|\nabla \phi|} \right)d\xb \nonumber \\
	=& \lim_{\varepsilon\rightarrow 0}\frac{\alpha_2}{2\varepsilon}\int_{\Omega}  \left(2p-\nabla\cdot\frac{\nabla (\phi+\varepsilon h)}{|\nabla (\phi+\varepsilon h)|}-\nabla\cdot\frac{\nabla \phi}{|\nabla \phi|}\right)\nabla\cdot \left(\frac{\nabla \phi}{|\nabla \phi|}-\frac{\nabla (\phi+\varepsilon h)}{|\nabla (\phi+\varepsilon h)|} \right) d\xb\nonumber\\
	=& \lim_{\varepsilon\rightarrow 0}\frac{\alpha_2}{2\varepsilon}\int_{\Omega}  \left(2p-\nabla\cdot\frac{\nabla (\phi+\varepsilon h)}{|\nabla (\phi+\varepsilon h)|}-\nabla\cdot\frac{\nabla \phi}{|\nabla \phi|}\right)\nabla\cdot \left(\frac{\nabla \phi}{|\nabla \phi|}-\frac{\nabla (\phi+\varepsilon h)}{|\nabla \phi|+ \varepsilon \frac{\nabla\phi\cdot \nabla h}{|\nabla \phi|}+O(\varepsilon^2)} \right) d\xb\nonumber\\
	=&\lim_{\varepsilon\rightarrow 0}\frac{\alpha_2}{2\varepsilon}\int_{\Omega}  \left(2p-\nabla\cdot\frac{\nabla (\phi+\varepsilon h)}{|\nabla (\phi+\varepsilon h)|}-\nabla\cdot\frac{\nabla \phi}{|\nabla \phi|}\right)\nabla\cdot \left(\frac{\varepsilon \frac{\nabla\phi\cdot \nabla h}{|\nabla \phi|}\nabla \phi- \varepsilon |\nabla \phi|\nabla h +O(\varepsilon^2)}{|\nabla \phi|(|\nabla \phi|+\varepsilon \frac{\nabla\phi\cdot \nabla h}{|\nabla \phi|}+ O(\varepsilon^2))} \right) d\xb\nonumber\\
	=&\lim_{\varepsilon\rightarrow 0}\frac{\alpha_2}{2}\int_{\Omega}  \left(2p-\nabla\cdot\frac{\nabla (\phi+\varepsilon h)}{|\nabla (\phi+\varepsilon h)|}-\nabla\cdot\frac{\nabla \phi}{|\nabla \phi|}\right)\nabla\cdot \left(\frac{ \frac{\nabla\phi\cdot \nabla h}{|\nabla \phi|}\nabla \phi-  |\nabla \phi|\nabla h +O(\varepsilon)}{|\nabla \phi|(|\nabla \phi|+\varepsilon \frac{\nabla\phi\cdot \nabla h}{|\nabla \phi|}+ O(\varepsilon^2))} \right) d\xb\nonumber\\
	=&\alpha_2\int_{\Omega}  \left(p-\nabla\cdot\frac{\nabla \phi}{|\nabla \phi|}\right)\nabla\cdot \left(\frac{ \frac{\nabla\phi\cdot \nabla h}{|\nabla \phi|}\nabla \phi-  |\nabla \phi|\nabla h }{|\nabla \phi|^2} \right)d\xb \nonumber\\
	=&-\alpha_2 \int_{\Omega} \nabla \left(p-\nabla\cdot\frac{\nabla \phi}{|\nabla \phi|}\right) \cdot \left( \frac{\nabla\phi\cdot \nabla h}{|\nabla \phi|^3}\nabla \phi -  \frac{1}{|\nabla \phi|}\nabla h \right)d\xb \nonumber\\
	=& -\alpha_2 \left( \int_{\Omega}\left(\frac{\nabla\phi}{|\nabla \phi|^3} \cdot \nabla \left(p-\nabla\cdot\frac{\nabla \phi}{|\nabla \phi|}\right)\right) \nabla\phi \cdot \nabla h d\xb - \int_{\Omega} \frac{1}{|\nabla \phi|} \nabla \left(p-\nabla\cdot\frac{\nabla \phi}{|\nabla \phi|}\right) \cdot \nabla h d\xb\right) \nonumber\\
	=&\alpha_2 \int_{\Omega} \nabla\cdot \left( \left(\frac{\nabla\phi}{|\nabla \phi|^3} \cdot \nabla \left(p-\nabla\cdot\frac{\nabla \phi}{|\nabla \phi|}\right)\right) \nabla\phi- \frac{1}{|\nabla \phi|} \nabla \left(p-\nabla\cdot\frac{\nabla \phi}{|\nabla \phi|}\right)  \right) h d\xb.
\end{align}
It implies that
\begin{align}
	\partial_{\phi} J_5(\phi,p)=\alpha_2\left[\nabla\cdot \left( \left(\frac{\nabla\phi}{|\nabla \phi|^3} \cdot \nabla \left(p-\nabla\cdot\frac{\nabla \phi}{|\nabla \phi|}\right)\right) \nabla\phi- \frac{1}{|\nabla \phi|} \nabla \left(p-\nabla\cdot\frac{\nabla \phi}{|\nabla \phi|}\right)  \right)\right].
	\label{eq.J5}
\end{align}
Substituting (\ref{eq.J4}) and (\ref{eq.J5}) into (\ref{eq.appen.phi}) gives rise to the equation of $\psi^{n+2/4}$.

\bibliographystyle{abbrv}
\bibliography{ref}
\end{document}